\documentclass{article}

    \PassOptionsToPackage{numbers, sort&compress}{natbib}

\usepackage[preprint]{neurips_2022}

\usepackage[utf8]{inputenc} %
\usepackage[T1]{fontenc}    %
\usepackage[pagebackref=true]{hyperref}       %
\usepackage{url}            %
\usepackage{booktabs}       %
\usepackage{amsfonts}       %
\usepackage{nicefrac}       %
\usepackage{microtype}      %

\usepackage{amsmath,amsfonts,bm}

\def\eqref#1{equation~\ref{#1}}

\def\1{\bm{1}}

\def\vm{{\bm{m}}}

\def\vp{{\bm{p}}}
\def\vq{{\bm{q}}}

\def\vw{{\bm{w}}}
\def\vx{{\bm{x}}}

\DeclareMathAlphabet{\mathsfit}{\encodingdefault}{\sfdefault}{m}{sl}
\SetMathAlphabet{\mathsfit}{bold}{\encodingdefault}{\sfdefault}{bx}{n}

\usepackage[font=footnotesize]{caption}
\usepackage{graphicx}
\usepackage[position=top]{subfig}
\usepackage{bigstrut}
\usepackage{paralist}
\usepackage{multicol}
\usepackage{multirow}
\usepackage{tabularx}
\usepackage{float}
\usepackage[table]{xcolor}

\usepackage{cuted}
\usepackage{flushend}
\usepackage{balance}
\usepackage{enumitem}
\usepackage{wrapfig}
\usepackage{pifont}

\newcommand{\cmark}{\textcolor{blue}{\ding{51}}}%
\newcommand{\xmark}{\textcolor{red}{\ding{55}}}%
\newcommand{\namark}{\textendash}%

\usepackage{hyperref}

\makeatletter
\def\@fnsymbol#1{\ensuremath{\ifcase#1\or * \or w \or c \or mpc \or \dagger\or \mathsection\or
   \ddager\or \mathparagraph\or \|\or **\or \dagger\dagger
   \or \ddagger\ddagger \else\@ctrerr\fi}}
\makeatother

\usepackage{amsmath}
\usepackage{amssymb}
\usepackage{mathtools}
\usepackage{amsthm}

\usepackage[linesnumbered,ruled,vlined]{algorithm2e}

\usepackage[capitalize,noabbrev]{cleveref}

\usepackage{dsfont}

\newcommand{\eg}{{\it e.g.}, }
\newcommand{\ie}{{\it i.e.}, }
\newcommand\op[1]{\operatorname{#1}}
\newcommand{\algo}{\textsc{Gem-Miner}}

\newcommand{\ep}{Edge-Popup}
\newcommand{\taskcifar}{\textbf{(Task 1)}}
\newcommand{\tasktimg}{\textbf{(Task 2)}}
\newcommand{\taskcal}{\textbf{(Task 3)}}

\newcommand{\gm}{\textsc{Gem-Miner}}

\theoremstyle{plain}

\theoremstyle{definition}

\usepackage[textsize=tiny]{todonotes}

\usepackage[page,toc,titletoc,title]{appendix}
\usepackage[subfigure]{tocloft}
\definecolor{mydarkblue}{rgb}{0,0.08,0.45}
\definecolor{mydarkgreen}{rgb}{0,0.45,0.08}
  \hypersetup{ %
    colorlinks=true,
    linkcolor=mydarkblue,
    citecolor=mydarkblue,
    filecolor=mydarkblue,
    urlcolor=mydarkblue}

\title{Rare Gems: Finding Lottery Tickets at Initialization}

\author{
Kartik Sreenivasan\thanks{Authors contributed equally to this paper.}\hspace{2mm}\footnotemark[2]\hspace{2mm},\ \ Jy-yong Sohn\footnotemark[1]\hspace{2mm}\footnotemark[2]\hspace{2mm},\ \  Liu Yang\footnotemark[2]\hspace{2mm},\ \ Matthew Grinde\footnotemark[2]\\\normalsize \textbf{Alliot Nagle}\footnotemark[2],\ \ \textbf{
Hongyi Wang\footnotemark[3],\ \ Eric Xing\footnotemark[4], \hspace{2mm} \ \ Kangwook Lee\footnotemark[2],\ \ Dimitris Papailiopoulos\footnotemark[2]} \\ \\
\normalsize$^c$ Carnegie Mellon University \ \ $^m$MBZUAI\ \ \ $^p$Petuum, Inc\\ $^w$ University of Wisconsin-Madison
}

\begin{document}

\maketitle

\begin{abstract}
Large neural networks can be pruned to a small fraction of their original size, with little loss in accuracy, by following a time-consuming ``train, prune, re-train'' approach.
\citet{frankle2018lottery} conjecture that we can
avoid this by training {\it lottery tickets}, \ie special sparse subnetworks found {\it at initialization}, that can be trained to high accuracy.
However, a subsequent line of work~\citep{frankle2020pruning, su2020sanity} presents concrete evidence that current algorithms for finding trainable networks at initialization, fail simple baseline comparisons, \eg against training random sparse subnetworks. 
Finding lottery tickets that train to better accuracy compared to simple baselines remains an open problem. 
In this work, we resolve this open problem by proposing \gm{} which finds lottery tickets {\it at initialization} that beat current baselines. \gm{} finds lottery tickets trainable to accuracy competitive or better than Iterative Magnitude Pruning (IMP), and does so up to $19\times$ faster. 

\end{abstract}

\section{Introduction}

A large body of research since the 1980s empirically observed that large neural networks can be compressed or sparsified to a small fraction of their original size while maintaining their predictive accuracy
\citep{HassibiStork93,MozerSmolensky89,LeCunEtAl90,han2015learning, HanEtAl16,hubara2017quantized,zhu2017prune}. Although several pruning methods have been proposed during the past few decades, many of them follow the ``\textit{train, prune, re-train}'' paradigm. Although the above methods result in very sparse, accurate models, they typically require several rounds of re-training, which is computationally intensive.

\citet{frankle2018lottery} suggest that this computational burden may be avoidable. 
They conjecture that given a randomly initialized network, one can find {\it a sparse subnetwork that can be trained to accuracy comparable to that of its fully trained dense counterpart}.  This trainable subnetwork found {\it at initialization} is referred to as a \emph{lottery ticket}.
The study above introduced iterative magnitude pruning (IMP) as a means of finding these lottery tickets. Their experimental findings laid the groundwork for what is now known as the {\it Lottery Ticket Hypothesis} (LTH).

Although \citet{frankle2018lottery} establish that the LTH is true for tasks like image classification on MNIST, they were not able to get satisfactory results for more complex datasets like CIFAR-10 and ImageNet when using deeper networks, such as VGG and ResNets~\citep{frankle2020linear}. In fact, subsequent work brought the effectiveness of IMP into question. \citet{su2020sanity} showed that even randomly sampled sparse subnetworks at initialization can beat lottery tickets found by IMP as long as the layerwise sparsities are chosen carefully. \citet{gale2019state} showed that methods like IMP which train tickets from initialization cannot compete with the accuracy of a model trained with pruning as part of the optimization process. 

\citet{frankle2020linear} explain the failures of IMP using the concept of \textit{linear mode connectivity} which measures the stability of these subnetworks to SGD noise. Extensive follow-up studies propose several heuristics for finding trainable sparse subnetworks at initialization~\citep{lee2018snip, wang2019picking, tanaka2020pruning}. However, subsequent work by \citet{frankle2020pruning, su2020sanity} show experimentally that all of these methods fail simple sanity checks. Most methods seem to merely identify good sparsities at each layer, but given those, random sparse subnetworks can be trained to similar or better accuracy.

\citet{frankle2020linear} show that with a small modification, IMP can beat these sanity checks; the caveat is that it no longer finds these subnetworks at initialization, but after a few epochs of \textit{warm-up} training. Since these  subnetworks are found {\it after initialization}, {\bf IMP with warmup does not find  lottery tickets}. 

As noted in the original work by Frankle \& Carbin~\cite{frankle2018lottery}, the importance of finding trainable subnetworks at initialization is computational efficiency. It is far preferable to train a sparse model from scratch, rather than having to deal with a large dense model, even if that is for a few epochs (which is what IMP with warmup does).
To the best of our knowledge, the empirical validity of the \textit{Lottery Ticket Hypothesis}, \ie the hunt for subnetworks at initialization trainable to SOTA accuracy, remains an open problem.

\begin{figure}[t] 
\centering
\includegraphics[width=0.9\textwidth]{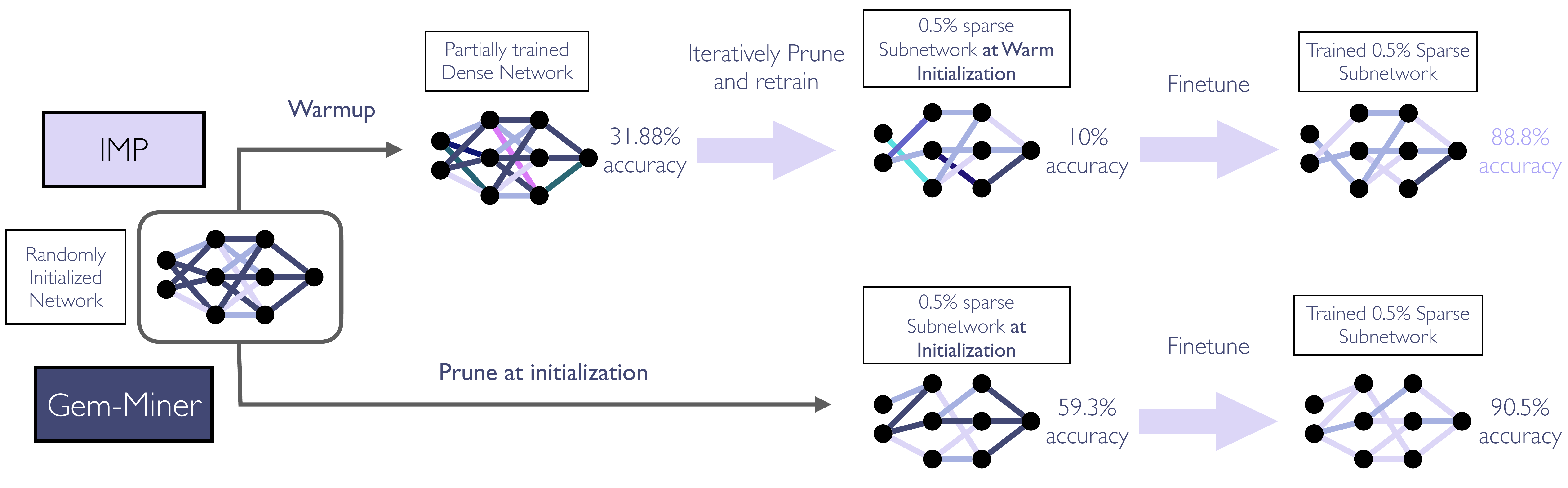}\\
\caption{Conceptual visualization of \gm{} vs IMP with warmup. The accuracies listed are on a $0.5\%$ sparse VGG-16 trained on CIFAR-10.
Given a randomly initialized network, both methods output a subnetwork which is then finetuned. IMP requires warmup \ie few epochs of training before it can find a sparse subnetwork. \gm{} finds a \emph{rare gem}, a subnetwork \emph{at initialization} that achieves high accuracy both before and after weight training.
}
\label{fig:GemMiner}
\end{figure}

\vspace{-2mm}
\paragraph{Our Contributions.} We resolve this open problem by developing \gm{}, an algorithm that finds sparse subnetworks \emph{at initialization}, trainable to accuracy comparable or better than IMP \emph{with warm-up}. 
\gm{} does so by first discovering {\it rare gems}. Rare gems are subnetworks at initialization that attain  accuracy far above random guessing, even before training. Rare gems can then be \textit{refined} to achieve near state-of-the-art accuracy. Simply put, rare gems are lottery tickets that also have high accuracy at initialization.

High accuracy at initialization is not a requirement for a network to be defined as a lottery ticket. However, if our end goal is high accuracy after training, then having high accuracy at initialization likely helps.

Rare gems found by \gm{} are the first lottery tickets to beat all baselines in \cite{frankle2020pruning, su2020sanity}. In Fig.~\ref{fig:GemMiner} we give a sketch of how our proposed algorithm {\it \gm{}} compares with IMP with warm start.
\gm{} finds these subnetworks in exactly the same number of epochs that it takes to train  them, and is up to 19$\times$ faster than IMP {\it with warmup}.

\section{Related Work}
\label{sec:related_work}
\paragraph{Lottery ticket hypothesis.}

Following the pioneering work of \citet{frankle2018lottery}, the search for lottery tickets has grown across several applications, such as language tasks, graph neural networks and federated learning~\citep{chen2021unified, li2020lotteryfl, girish2021lottery, chen2020lottery}. 
While the LTH itself has yet to be proven mathematically, the so-called strong LTH has been derived which shows that any target network can be approximated by pruning a randomly initialized network with minimal overparameterization~\citep{malach2020proving, pensia2020optimal, orseau2020logarithmic}. 
Recently, it has been shown that for such approximation results it suffices to prune a random binary network with slightly larger overparameterization~\citep{sreenivasan2021finding, diffenderfer2020multi}.

\vspace{-0.3cm}
\paragraph{Pruning at initialization.}
While network pruning has been studied since the 1980s,
finding sparse subnetworks at initialization is a more recently explored approach. \citet{lee2018snip} propose SNIP, which prunes based on a heuristic that approximates the importance of a connection.
\citet{tanaka2020pruning} propose SynFlow which prunes the network to a target sparsity without ever looking at the data. 
\citet{wang2019picking} propose GraSP which computes the importance of a weight based on the Hessian gradient product.
The goal of these algorithms is to find a subnetwork that can be trained to high accuracy. \citet{ramanujan2020s} propose Edge-Popup (EP) which finds a subnetwork at initialization that has high accuracy to begin with. 
Unfortunately, they also note that these subnetworks are not conducive to further finetuning.

The above algorithms are all based on the idea that one can assign a ``score'' to each weight to measure its importance. Once such a score is assigned, one simply keeps the top fraction of these scores based on the desired target sparsity. This may be done by sorting the scores layer-wise or globally across the network. Additionally, this can be done in \emph{one-shot} (SNIP, GraSP) or \emph{iteratively} (SynFlow). Note that IMP can also be fit into the above framework by defining the ``score'' to be the \textit{magnitude} of the weights and then pruning globally across the network iteratively.

More recently, \citet{alizadeh2021prospect} propose ProsPr which utilizes the idea of \textit{meta-gradients} through the first few steps of optimization to determine which weights to prune. Their intuition is that this will lead to masks at initialization that are more amenable to training to high accuracy within a few steps. While it finds high accuracy subnetworks, we show in Section~\ref{subsec:prospr} that it fails to pass the sanity checks of \citet{frankle2020pruning} and \citet{su2020sanity}.

\vspace{-3mm}
\paragraph{Sanity checks for lottery tickets.}
A natural question that arises with pruning at initialization is whether these algorithms are truly finding interesting and nontrivial subnetworks, or if their performance after finetuning can be matched by simply training equally sparse, yet random subnetworks. \citet{ma2021sanity} propose more rigorous definitions of winning tickets and study IMP under several settings with careful tuning of hyperparameters. 
\citet{frankle2020pruning} and \citet{su2020sanity} introduce several sanity checks (i) Random shuffling (ii) Weight reinitialization (iii) Score inversion and (iv) Random Tickets. Even at their best performance, they show that SNIP, GraSP and SynFlow merely find a good sparsity ratio in each layer and fail to surpass, in term of accuracy, fully trained randomly selected subnetworks, whose sparsity per layer is similarly tuned. \citet{frankle2020pruning} show through extensive experiments that none of these methods show accuracy deterioration after random reshuffling.
We explain the sanity checks in detail in Section~\ref{sec:experiments} and use them as baselines to test our own algorithm.

\begin{table}[t]
    \caption{
	We compare the different popular pruning methods in the literature on whether they prune at initialization, are finetunable and pass sanity checks. We also list the amount of computation they need to find a $1.4\%$ sparse subnetwork on ResNet-20, CIFAR-10. For consistency, we do not include the time required to finetune this subnetwork to full accuracy as it would be equal for all methods. For single-shot pruning method we list it as $1$ epoch but this depends on the choice of batch-size. Learning Rate Rewinding which we label \citet{renda2020comparing} is a pruning after training algorithm and just outputs a high accuracy subnetwork and hence the sanity checks do not apply to it.
	}
    \vspace{1mm}
    \centering
    \scriptsize
    \setlength{\tabcolsep}{4pt} %
    \renewcommand{\arraystretch}{0.5}
		 {
			\begin{tabular}{ccccc}
				\toprule Pruning Method
				& Prunes at initialization & Finetunable  & Passes sanity checks & \shortstack{Computation to reach $1.4\%$ sparsity}
				\bigstrut\\
				\midrule
				IMP~\cite{frankle2018lottery}
				& \xmark
				& \cmark
				& \cmark
				& 2850 epochs
				\bigstrut\\
				SNIP~\cite{lee2018snip}
				& \cmark
				& \cmark
				& \xmark
				& 1 epoch
				\bigstrut\\
				GraSP~\cite{wang2019picking}
				& \cmark
				& \cmark
				& \xmark
				& 1 epoch
				\bigstrut\\
				SynFlow~\cite{tanaka2020pruning}
				& \cmark
				& \cmark
				& \xmark
				& 1 epoch
				\bigstrut\\
				Edge-popup~\cite{ramanujan2020s}
				& \cmark
				& \xmark
				& \xmark
				& 150 epochs
				\bigstrut\\
				Smart Ratio~\cite{su2020sanity}
				& \cmark
				& \cmark
				& \namark
				& $\mathcal{O}(1)$
				\bigstrut\\
				Learning Rate Rewinding~\cite{renda2020comparing}
				& \xmark
				& \namark
				& \namark
				& 3000 epochs
				\bigstrut\\
				\textbf{\algo{}}
				& \textbf{\cmark}
				& \textbf{\cmark}
				& \textbf{\cmark}
				& \textbf{150 epochs}
				\bigstrut\\
				\bottomrule
			\end{tabular}}%
    \vspace{2mm}	
	\label{table:lits_of_pruning_algos}
\end{table}

\vspace{-3mm}
\paragraph{Pruning during/after training.}
While the above algorithms prune at/near initialization, there exists a rich literature on algorithms which prune during/after training. Unlike IMP, algorithms in this category do not rewind the weights. They continue training and pruning iteratively. \citet{frankle2020pruning} and \citet{gale2019state} show that pruning at initialization cannot hope to compete with these algorithms. While they do not find lottery tickets, they do find high accuracy sparse networks. \citet{zhu2017prune} propose a gradual pruning schedule where the smallest fraction of weights are pruned at a predefined frequency. They show that this results in models up to 95\% sparsity with negligible loss in performance on language as well as image processing tasks. \citet{gale2019state} and \citet{frankle2020pruning} also study this as a baseline under the name \textit{magnitude pruning after training}. \citet{renda2020comparing} show that rewinding the learning rate as opposed to weights(like in IMP) leads to the best performing sparse networks. However, it is important to remark that these are not Lottery Tickets, merely high accuracy sparse networks. We contrast these different methods in Table~\ref{table:lits_of_pruning_algos} in terms of whether they prune at initialization, their finetunability, whether they pass sanity checks as well as their computational costs.

Finally, we note that identifying a good pruning mask can be thought of as training a binary network where the loss is computed over the element-wise product of the original network with the mask. This has been explored in the quantization during training literature~\citep{courbariaux2015binaryconnect, simons2019review, hubara2016binarized}.

\section{\algo{}: Discovering Rare Gems}

\paragraph{Setting and notation.}

Let ${S} = \{ (\vx_i, y_i) \}_{i=1}^n$ be a given training dataset for a $k$-classification problem, where $\vx_i \in \mathbb{R}^{d_0}$ denotes a feature vector and label $y_i \in \{1,\ldots,k\}$ denotes its label.
Typically, we wish to train a neural network classifier 
$f(\vw; \vx):\mathbb{R}^{d_0}\rightarrow \{1,\ldots,k\}$, where $\vw \in \mathbb{R}^d$ denotes the set of weight parameters of this neural network. 
The goal of a pruning algorithm is to extract a mask $\vm =\{0,1\}^{d}$, so that the pruned network is denoted by $f(\vw \odot \vm ; \vx)$, where $\odot$ denotes the element-wise product.
We define the \emph{sparsity} of this network to be the fraction of non-zero weights $s = \Vert \vw \odot \vm \Vert_{0}/d$. The loss of a classifier on a single sample $(\vx, y)$ is denoted by $\ell(f(\vw \odot \vm; \vx ), y)$, which captures a measure of discrepancy between prediction and reality.
In what follows, we will denote by $\vw_0 \in \mathbb{R}^d$ to be the set of random initial weights. The type of randomness will be explicitly mentioned when necessary.
\vspace{-0.3cm}

\paragraph{On the path to rare gems; first stop: Maximize pre-training accuracy.}

A rare gem needs to satisfy three conditions: (i) sparsity, (ii) non-trivial pre-training accuracy, and (iii) that it can be finetuned to achieve  accuracy close to that of the fully trained dense network. 
This is not an easy task as we have two different objectives in terms of accuracy (pre-training and post-training), and it is unclear if a good subnetwork for one objective is also good for the other.
However, since pre-training accuracy serves as a lower bound on the final performace, we focus on maximizing that first, and then attempt to further improve it by finetuning.

\begin{wrapfigure}{hr}{0.5\columnwidth}
\vspace{-0.8cm}
	\centering
	\includegraphics[width=0.5\textwidth]{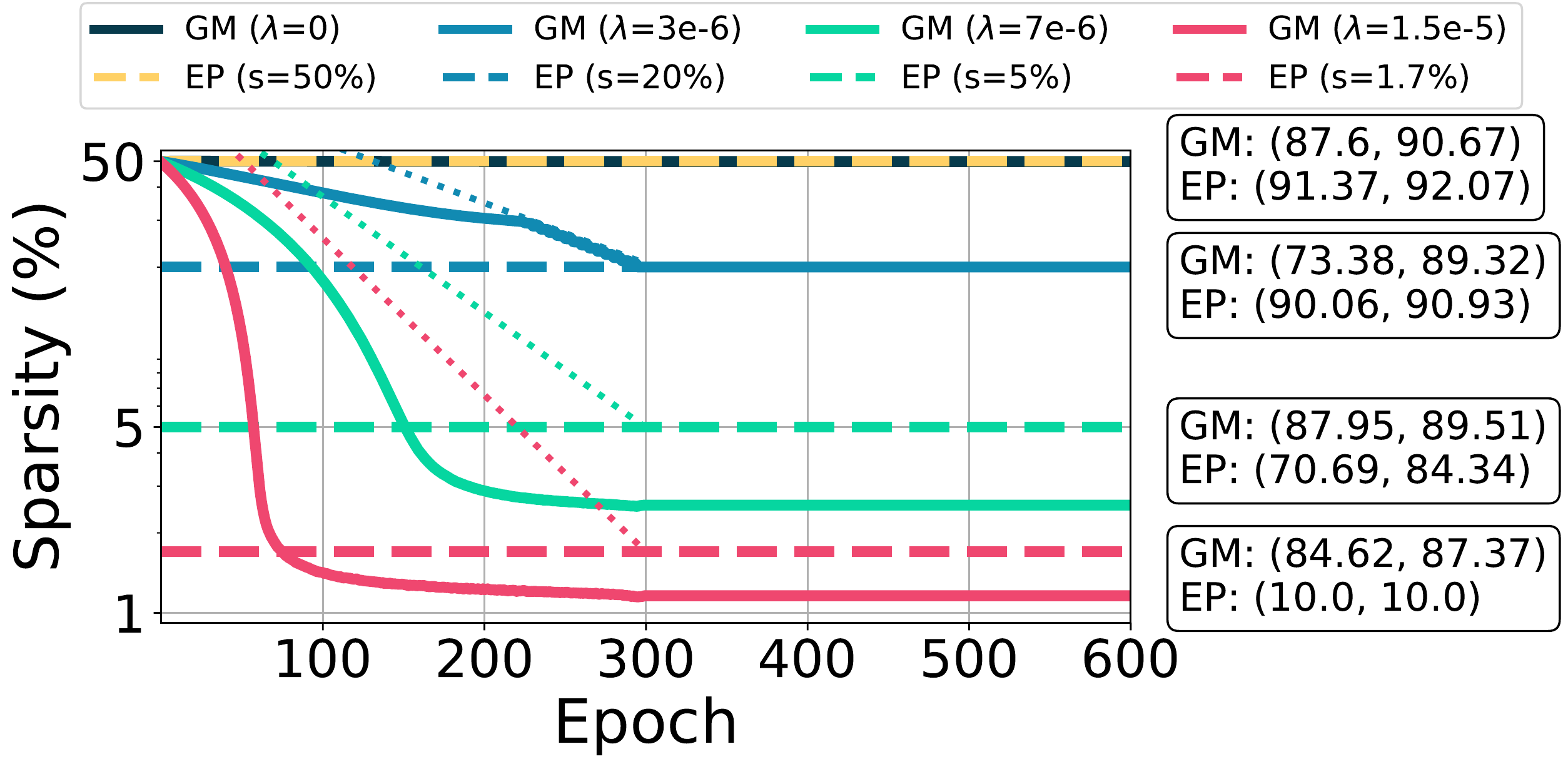}
	\caption{The sparsity of intermediate results, the accuracy of the final output, and the accuracy after finetuning on MobileNet-V2, CIFAR-10. For \gm{} (GM), we also visualize the sparsity upper bounds as dotted lines. As $\lambda$ increases, note that the sparsity of \gm{}'s output decreases. For$\lambda = 3\cdot 10^{-6}$, the iterative freezing algorithm kicks in around epoch 220, regularizing the sparsity thereafter. The gem found by \gm{}($\lambda = 1.5\cdot 10^{-5}$) achieves accuracy of $84.62\%$ before finetuning and $87.37\%$ after finetuning, while EP is unable to achieve non-trivial accuracy before or after finetuning.}\label{fig:EP_GM}
\vspace{-0.9cm}
\end{wrapfigure}

Our algorithm is inspired by  Edge-Popup (EP)~\citep{ramanujan2020s}. EP successfully finds subnetworks with high pre-training accuracy but it has two major limitations: (i) it does not work well in the high sparsity regime (\eg $<5\%$), and (ii) most importantly, the subnetworks it finds are not conducive to further finetuning. 

In the following, we take \gm{} apart and describe the components that allow it to surpass these issues.

\vspace{-0.3cm}
\paragraph{\gm{} without sparsity control.}
Much like EP, \gm{} employs a form of backpropagation, and works as follows. Each of the random weights $[{\bf w}_0]_i$ in the original network is associated with a normalized score $p_i\in[0,1]$. These normalized scores become our optimization variables and are responsible for computing the {\it supermask} ${\bf m}$, \ie the pruning pattern of the network at initialization.

For a given set of weights $\vw$ and scores $\vp$, \gm{} sets the effective weights as $\vw_{\op{eff}} = \vw \odot r(\vp)$, where $\op{r}(\cdot)$ is an element-wise rounding function, and ${\bf m} = r(\vp)$ is the resulting supermask. 
The rounding function can be changed, \eg $r$ can perform randomized rounding, in which case $p_i$ would be the probability of keeping weight $w_i$ in ${\bf m}$. In our case, we found that simple deterministic rounding, \ie $r(p_i) = {\bf 1}_{p_i\ge 0.5}$ works well.

At every iteration \gm{} samples a batch of training data and performs backpropagation on the loss of the effective weights, with respect to the scores $\vp$, while projecting back to $[0,1]$ when needed. During the forward pass, due to the rounding function, the effective network used is indeed a subnetwork of the given network.
Here, since $r(\vp)$ is a non-differentiable operation we use the Straight Through Estimator (STE)~\cite{bengio2013estimating} which backpropagates through the indicator function as though it were the identity function.

Note that this vanilla version of \gm{} is unable to exercise control over the final sparsity of the model.
For reasons that will become evident in below, we will call this version of our algorithm \gm($0$).
There is already a stark difference from EP: \gm($0$) will automatically find the optimal sparsity, while EP requires the target sparsity $s$ as an input parameter.

However, at the same time, this also significantly limits the applicability of \gm($0$) as one cannot obtain a highly sparse gem. 
Shown as a dark blue curve in Fig.~\ref{fig:EP_GM} is the sparsity of \gm{}($0$).
Here, we run \gm{} with a randomly initialized MobileNet-V2 network on CIFAR-10.
Note that the sparsity stays around $50\%$ throughout the run, which is consistent with the observation by \citet{ramanujan2020s} that accuracy of subnetworks at initialization is maximized at around $50\%$ sparsity.

\SetKwComment{Comment}{/*}{*/}

\begin{wrapfigure}{L}{0.5\textwidth}
\begin{minipage}{0.5\textwidth}
\begin{algorithm}[H]
\small
\DontPrintSemicolon
\SetNoFillComment
\caption{\algo{}}\label{Algo:GM}
\KwIn{Dataset $D = \{(\vx_i, y_i)\}$, learning rate $\eta$, rounding function $r(\cdot)$, number of epochs $E$, freezing period $T$, target sparsity $s \in [0,1]$}
\KwOut{Mask $\vm = r(\vp) \odot \vq \in \{0,1\}^d$}
$c \gets \frac{\ln{(1/s)}}{E}$, $\vq \leftarrow \mathbf{1}_d$\;
$\vw, \vp \gets$ random vector in $\mathbb{R}^d$, \\
\hspace{10mm} random vector in $[0,1]^d$\;
\For{$j$ in $1,2,\dots, E$}{
    \For{$(\vx_i, y_i) \in D$}{
      $\vw_{\op{eff}} \gets (\vw \odot \vq) \odot r(\vp)$\;
      $\vp \gets \vp - \eta \nabla_{\vp}\  \ell( f(\vw_{\op{eff}}; \vx_i), y_i)$\; ~{\tcc{STE}}
      $\vp \gets \op{proj}_{[0,1]^d} \vp$\;
    }
    \uIf{$\mathrm{mod} (j, T) = 0$}{
        $I_{1} \gets \{i: q_i = 1\}$ \\ $\vp_{sorted} \gets \op{sort}(\vp_{i \in I_{1}})$\;
        $\vp_{bottom} \gets$ Bottom-($1-e^{cT}$) fraction \\ \hspace{14mm} of $\vp_{sorted}$\;
        $q \gets q \odot \mathds{1}_{p_i \notin \vp_{bottom}}$\;
      }
    }
\end{algorithm}
\vspace{-0.5cm}
\end{minipage}
\end{wrapfigure}

\vspace{-0.6cm}
\paragraph{Regularization and Iterative freezing.}
\gm{}($0$) is a good baseline algorithm for finding accurate subnetworks at initialization, but it cannot be used to find \emph{rare gems}, which need to be sparse and trainable. 
To overcome this limitation, we apply a standard trick -- we add a regularization term to encourage sparsity. 
Thus, in addition to the task loss computed with the effective weights, we also compute the $L_2$ or $L_1$ norm of the score vector $\vp$ and optimize over the total regularized loss.
More formally, we minimize $\ell := \ell_{\op{task}} + \lambda \ell_{\op{reg}}$, where $\lambda$ is the hyperparameter and $\ell_{\op{reg}}$ is either $L_2$ or $L_1$ norm of the score vector $\vp$. 

We call this variant \gm{}($\lambda$), where $\lambda$ denotes the regularization weight.
This naming convention should explain why we called the initial version \gm{}($0$).

The experimental results in Fig.~\ref{fig:EP_GM} show that this simple modification indeed allows us to control the sparsity of the solution.
We chose to use the $L_2$ regularizer, however preliminary experiments showed that $L_1$ performs almost identically.
By varying $\lambda$ from $\lambda = 0$ to $\lambda = 7 \cdot 10^{-6}$ and $\lambda = 1.5 \cdot 10^{-5}$, the final sparsity of the gem found by \gm($\lambda$) becomes $2.5\%$ and $1.4\%$, respectively.

One drawback of this regularization approach is that it only indirectly controls the sparsity. 
If we have a target sparsity $s$, then there is no easy way of finding the appropriate value of $\lambda$ such that the resulting subnetwork is $s$-sparse.
If we choose $\lambda$ to be too large, then it will give us a gem that is way too sparse; too small a $\lambda$ and we will end up with a denser gem than what is needed. 
As a simple heuristic, we employ \emph{iterative freezing}, which is widely used in several existing pruning algorithms, including IMP~\citep{frankle2018lottery,zhu2017prune,gale2019state}.
More specifically, we can design an exponential function $\overline{s}(j) = e^{-cj}$ for some $c > 0$, which will serve as the upper bound on the sparsity.
If the total number of epochs is $E$ and the target sparsity is $s$, we have $\overline{s}(E) = e^{-cE} = s$. 
Thus, we have $c = \ln{(1/s)}/E$.

Once this sparsity upper bound is designed, the iterative freezing mechanism regularly checks the current sparsity to see if the upper bound is violated or not. 
If the sparsity bound is violated, it finds the smallest scores, zeros them out, and freezes their values thereafter.
By doing so, we can guarantee the final sparsity even when $\lambda$ was not sufficiently large. 
To see this freezing mechanism in action, refer the blue curve in Fig.~\ref{fig:EP_GM}.
Here, the sparsity upper bounds (decreasing exponential functions) are visualized as dotted lines. 
Note that for the case of $\lambda = 3\cdot 10^{-6}$, the sparsity of the network does not decay as fast as desired, so it touches the sparsity upper bound around epoch 220. 
The iterative freezing scheme kicks in here, and the sparsity decay is controlled by the upper bound thereafter, achieving the specified target sparsity at the end.

The full pseudocode of \algo{} is provided in Algorithm~\ref{Algo:GM}.
There are two minor implementation details which differ from the explanation above: (i) we impose the iterative freezing every $T$ epochs, not every epoch and (ii) iterative freezing is imposed even when the sparsity bound is not violated.

\begin{figure*}[ht] 
\centering
\includegraphics[width=0.99\textwidth]{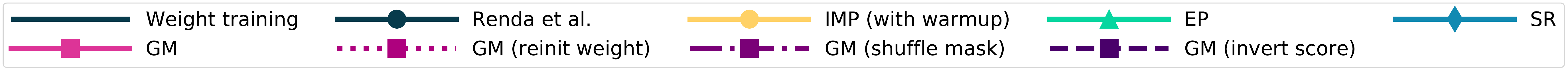}%
\\
\vspace{-2mm}\hspace{-1mm}
\subfloat[ResNet-20]{\includegraphics[height=19mm]{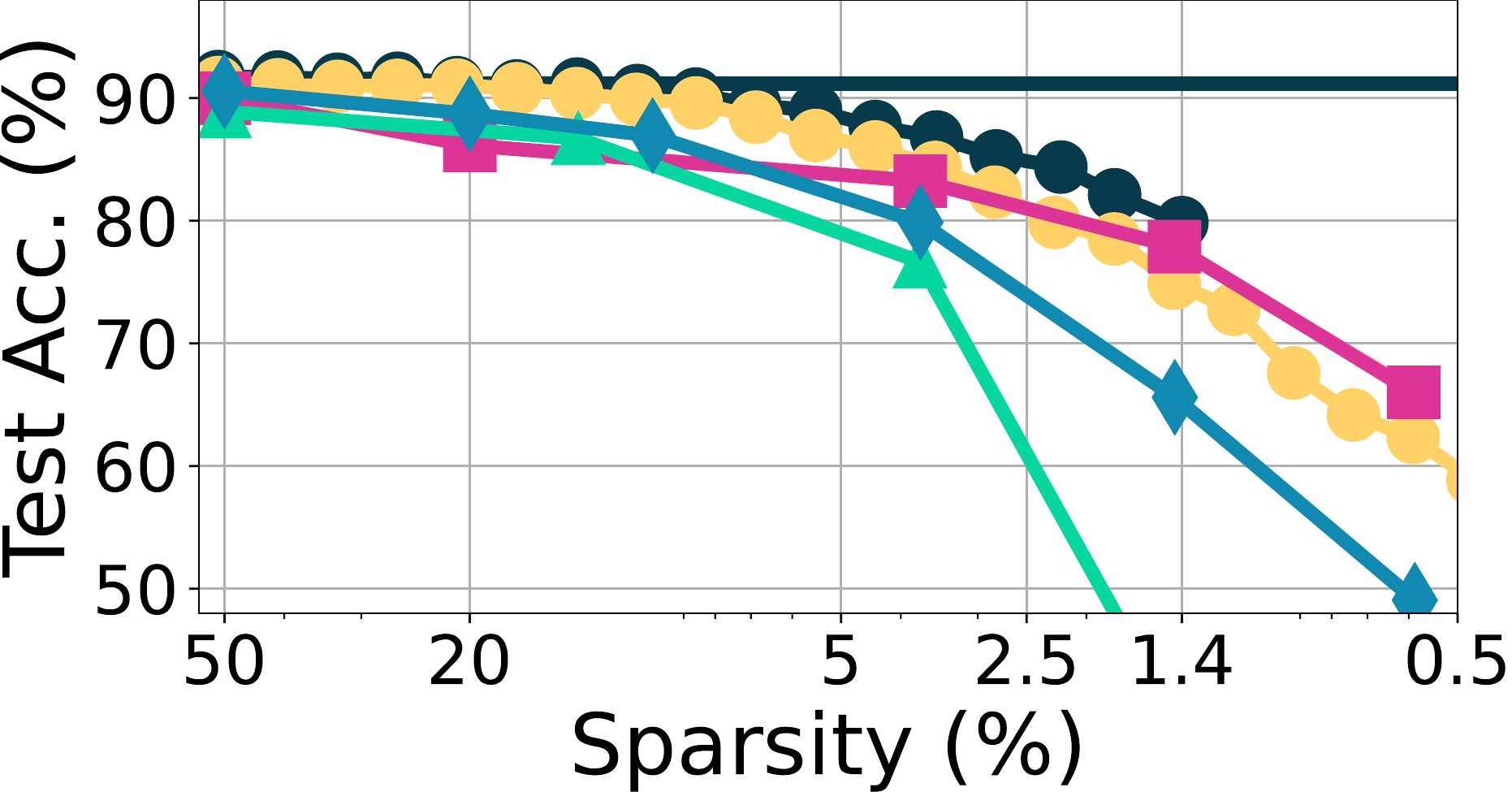}}
\hspace{0.1mm}
\subfloat[MobileNet-V2]{\includegraphics[height=19mm]{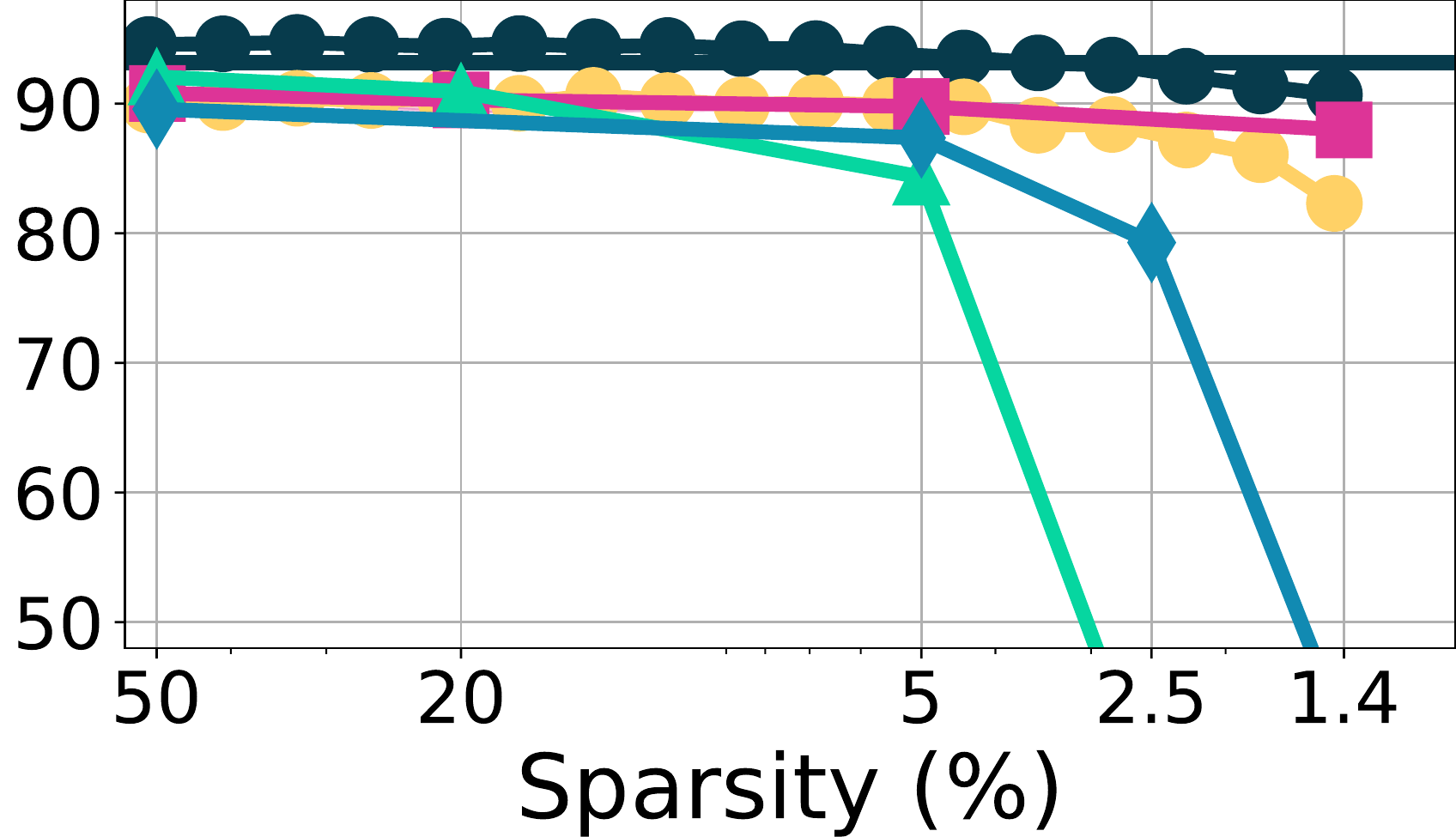}}
\hspace{0.1mm}
\subfloat[VGG-16]{\includegraphics[height=19mm]{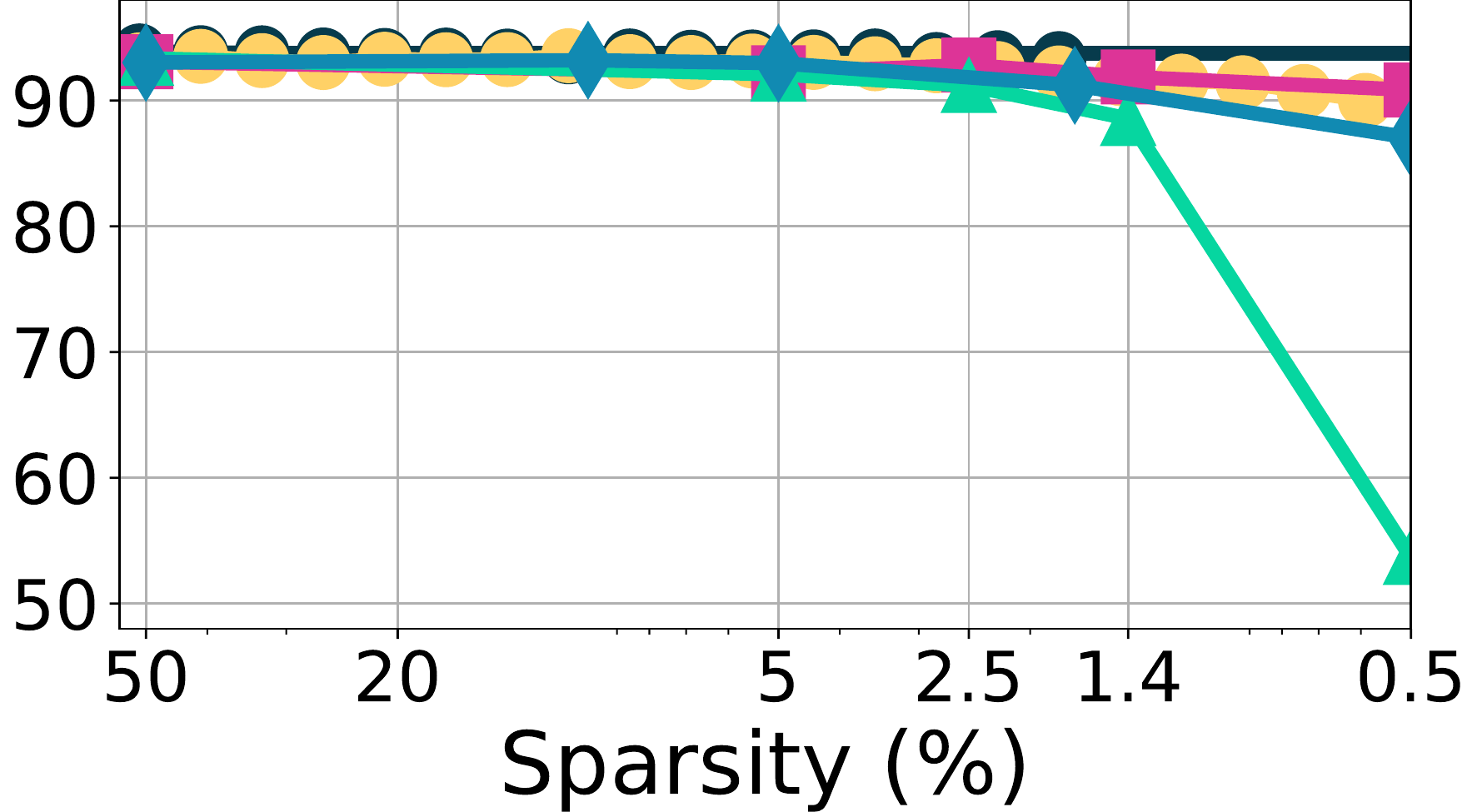}}
\hspace{0.1mm}
\subfloat[WideResNet-28-2]{\includegraphics[height=19mm]{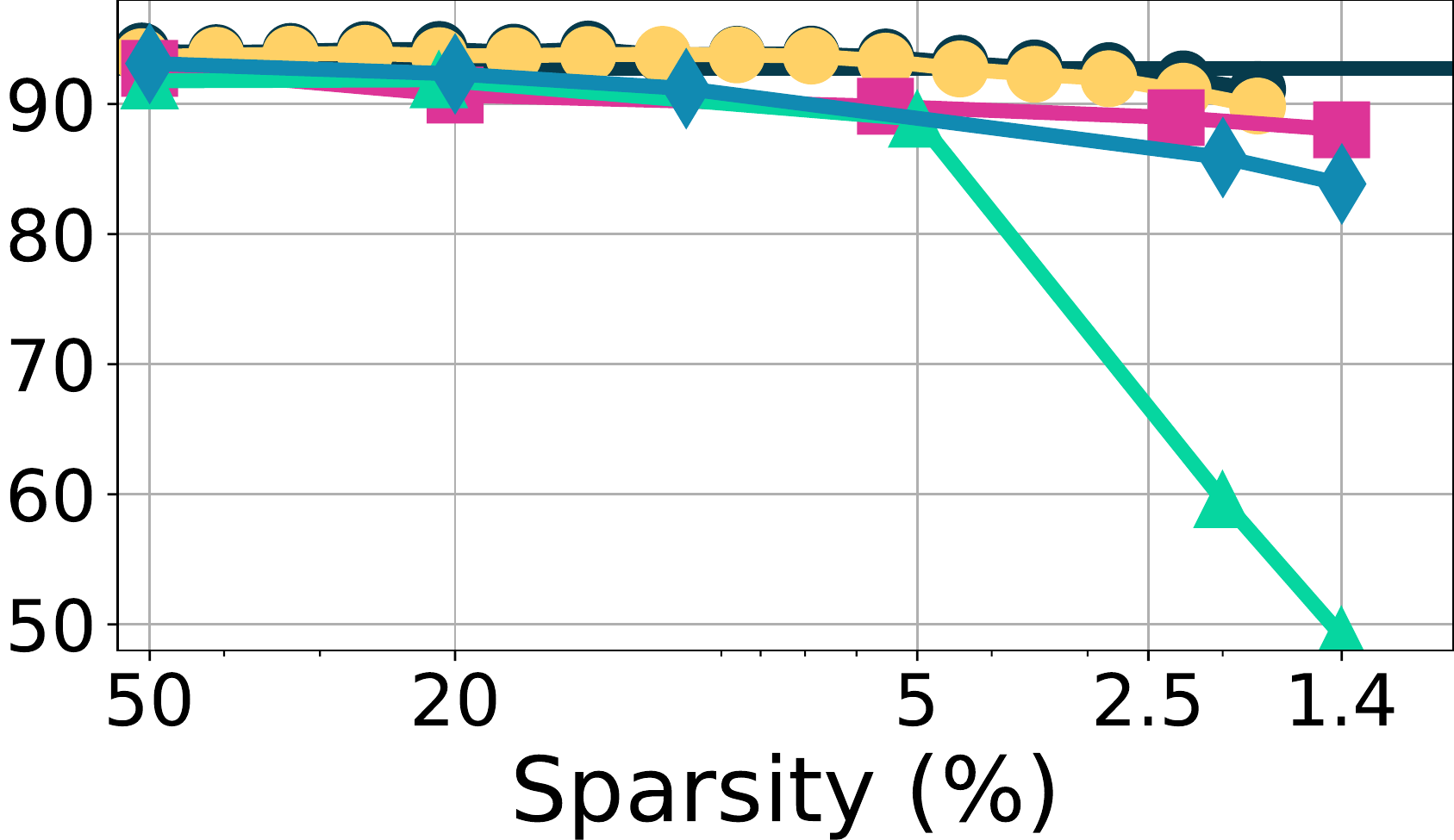}}
\\
\vspace{2mm}
\includegraphics[height=19mm]{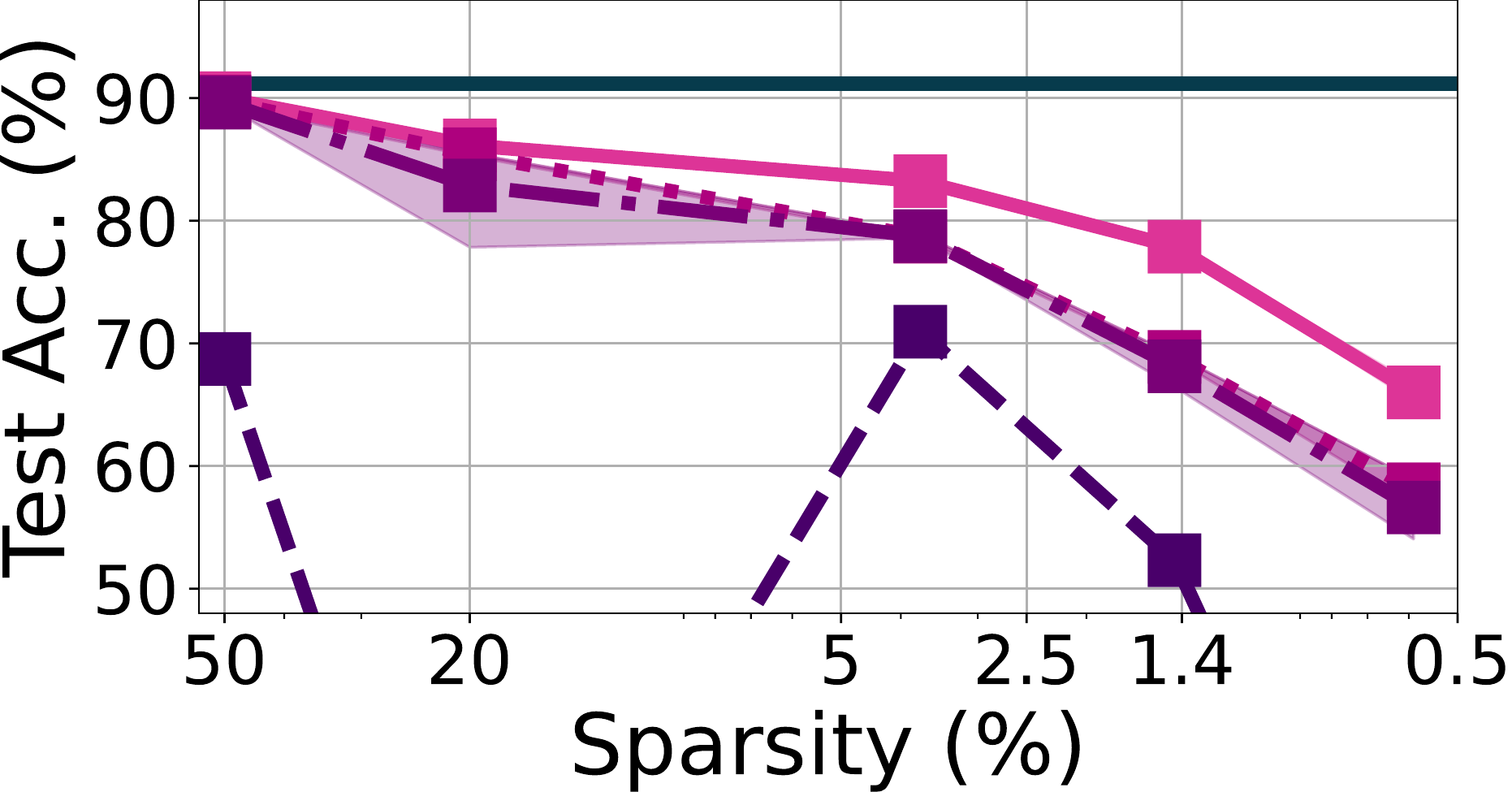}
\includegraphics[height=19mm]{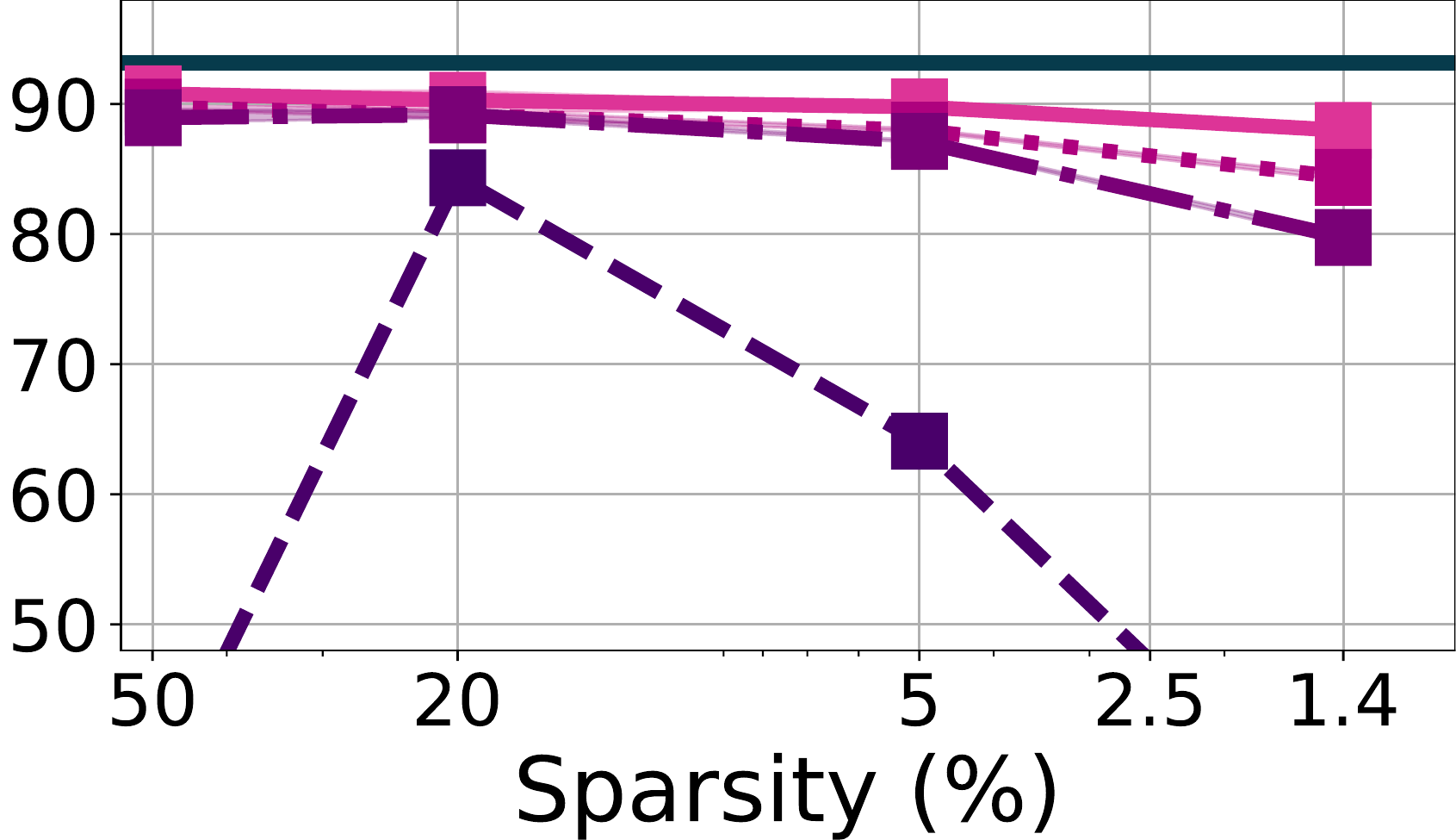}
\includegraphics[height=19mm]{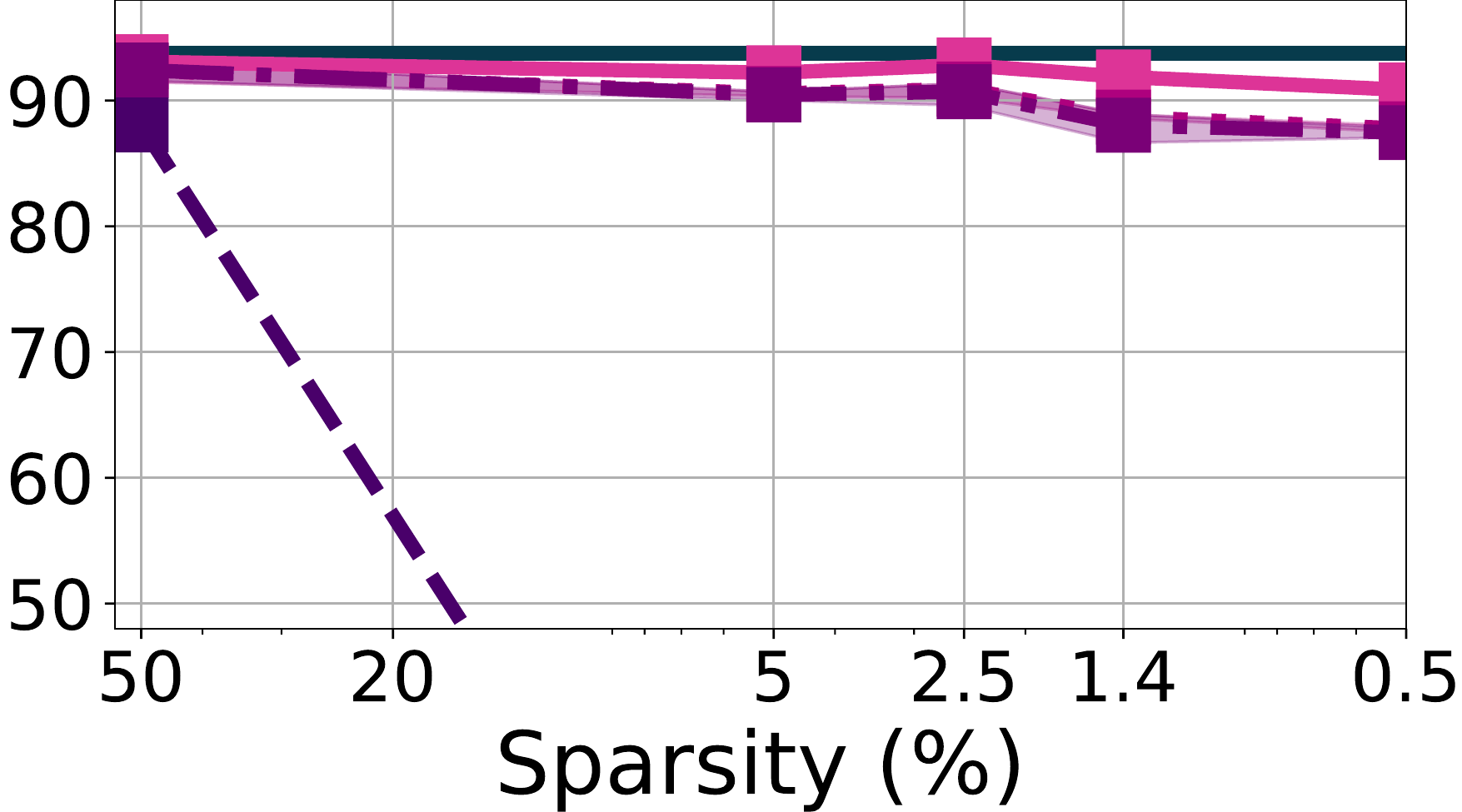}
\includegraphics[height=19mm]{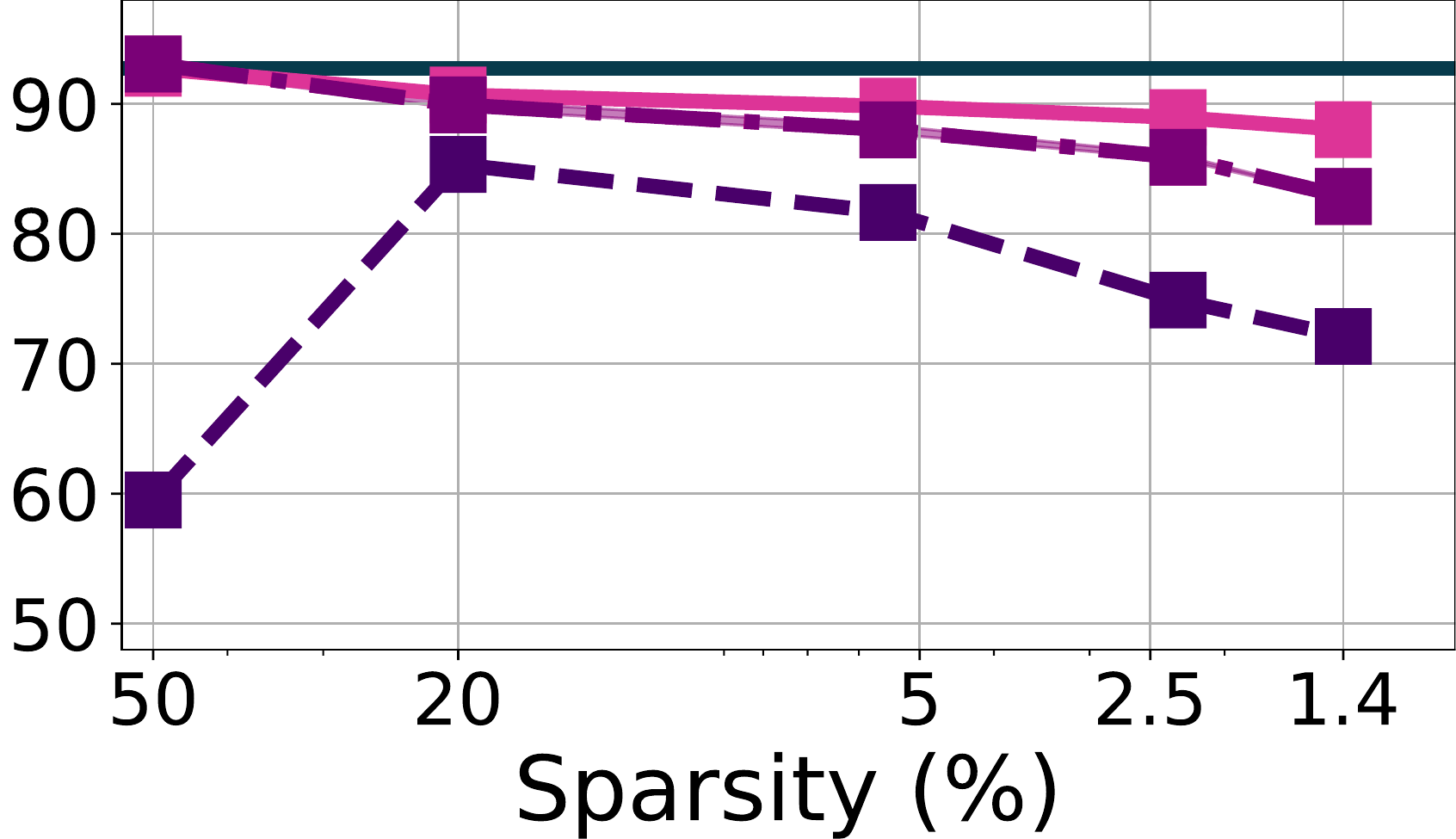}
\caption{
Performance of different pruning algorithms on CIFAR-10 for benchmark networks. Top: post-finetune accuracy; Bottom: sanity check methods suggested in~\citet{frankle2020pruning} applied on \algo{} (GM). Note that GM achieves similar post-finetune accuracy as IMP, and typically outperforms it in the sparse regime. GM has higher post-finetune accuracy than EP and Smart Ratio (SR). GM also passes the sanity checks suggested in~\citet{frankle2020pruning}.
Finally, GM (which prunes \emph{at} init) nearly achieves the performance of Renda et al. (which is a pruning after training method) in the sparse regime, e.g., 1.4\% sparsity in ResNet-20.
}
\label{fig:cifar10}
\end{figure*}

\vspace{-0.3cm}
\section{Experiments}
\label{sec:experiments}
In this section, we present the experimental results\footnote{Our codebase can be found at \url{https://anonymous.4open.science/r/pruning_is_enough-F0B0}.} for the performance of \gm{} across various tasks.

\vspace{-0.3cm}
\paragraph{Tasks.}
We evaluate our algorithm on \taskcifar{} CIFAR-10 classification, on various networks including ResNet-20, MobileNet-V2, VGG-16, and WideResNet-28-2,
\tasktimg{} TinyImageNet classification on ResNet-18 and ResNet-50, 
and \taskcal{} Finetuning on the Caltech-101~\citep{fei2004learning} dataset using a ResNet-50 pretrained on ImageNet.
The detailed description of the datasets, networks and hyperparameters can be found in Section~\ref{sec:exp_setup} of the Appendix. 

\vspace{-2mm}
\paragraph{Proposed scheme.}
We run \algo{} with an $L_2$ regularizer.
If a network reaches its best accuracy after $E$ epochs of weight training, then we run \algo{} for $E$ epochs to get a sparse subnetwork, and then run weight training on the sparse subnetwork for another $E$ epochs. 

\vspace{-2mm}
\paragraph{Comparisons.}
We tested our method against five baselines: dense weight training and four pruning algorithms: (i) IMP~\citep{frankle2020linear}, (ii) Learning rate rewinding~\citep{renda2020comparing}, denoted by Renda et al.,
(iii) \ep{} (EP)~\citep{ramanujan2020s}, and (iv) Smart-Ratio (SR) which is the random pruning method proposed by \citet{su2020sanity}.

We also ran the following sanity checks,  proposed by \citet{frankle2020pruning}:
(i) (Random shuffling): To test if the algorithm prunes specific connections, we randomly shuffle the mask at every layer. (ii) (Weight reinitialization): To test if the final mask is specific to the weight initialization, we reinitialize the weights from the original distribution. (iii) (Score inversion): Since most pruning algorithms use a heuristic/score function as a proxy to measure the importance of different weights, we invert the scoring function to check whether it is a valid proxy. More precisely, this test involves pruning the weights which have the \emph{smallest} scores rather than the largest. In all of the above tests, if the accuracy after finetuning the new subnetwork does not deteriorate significantly, then the algorithm is merely identifying optimal layerwise sparsities.

\subsection{Rare gems obtained by \algo{}}

\paragraph{Task 1.}
Fig.~\ref{fig:cifar10} shows the sparsity-accuracy tradeoff for various pruning methods trained on CIFAR-10 using ResNet-20, MobileNet-V2, VGG-16 and WideResNet-28-2. 
For each column (network), we compare IMP, IMP with learning rate rewinding (Renda et al.), \algo{}, EP, and SR in two performance metrics: the top row shows the accuracy of the subnetwork after weight training and  bottom row shows the result of the sanity checks on \algo{}. 

As shown in the top row of Fig.~\ref{fig:cifar10}, \algo{} finds a lottery ticket \emph{at} initialization. It reaches accuracy similar to IMP after weight training. Moreover, for in the sparse regime (e.g., below 1.4\% for ResNet-20 and MobileNet-V2), \algo{} outperforms IMP in terms of post-finetune accuracy. The bottom row of Fig.~\ref{fig:cifar10} shows that \algo{} passes the sanity check methods. For all networks, the performance in the sparse regime (1.4\% sparsity or below) shows that the suggested \algo{} algorithm enjoys 3--10\% accuracy gap with the best performance among variants. 
The results in the top row show that \algo{} far outperforms the random network with smart ratio (SR).

\begin{figure}[t!] 
\centering
\includegraphics[width=0.99\textwidth]{figures/Legend_with_Renda_Latest_withborder.pdf}\\
\subfloat[\centering{\scriptsize{CIFAR-100, ResNet-32}}]{\includegraphics[height=18mm]{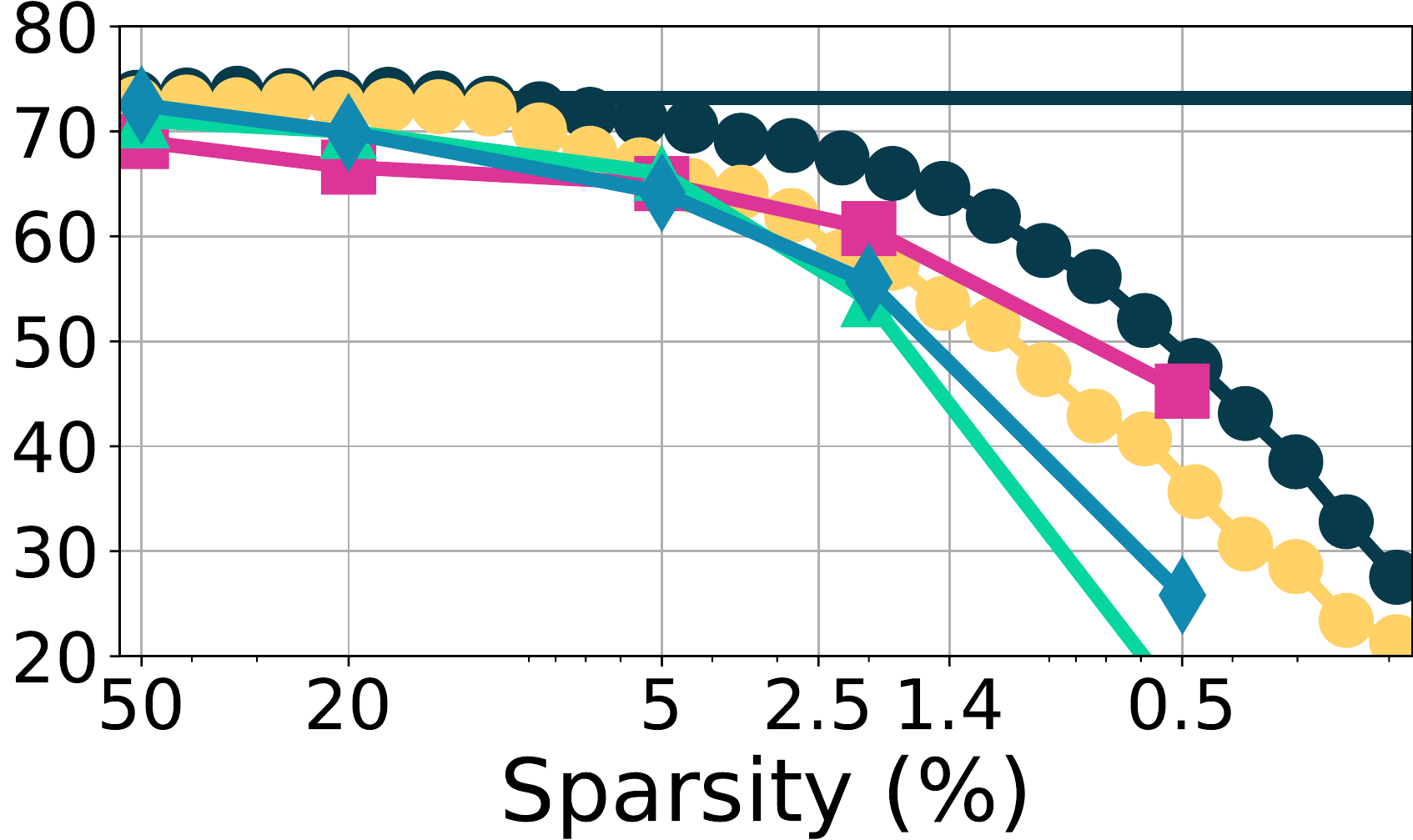}}
\subfloat[\centering{\scriptsize{TinyImageNet, ResNet-18}}]{\includegraphics[height=18mm
]{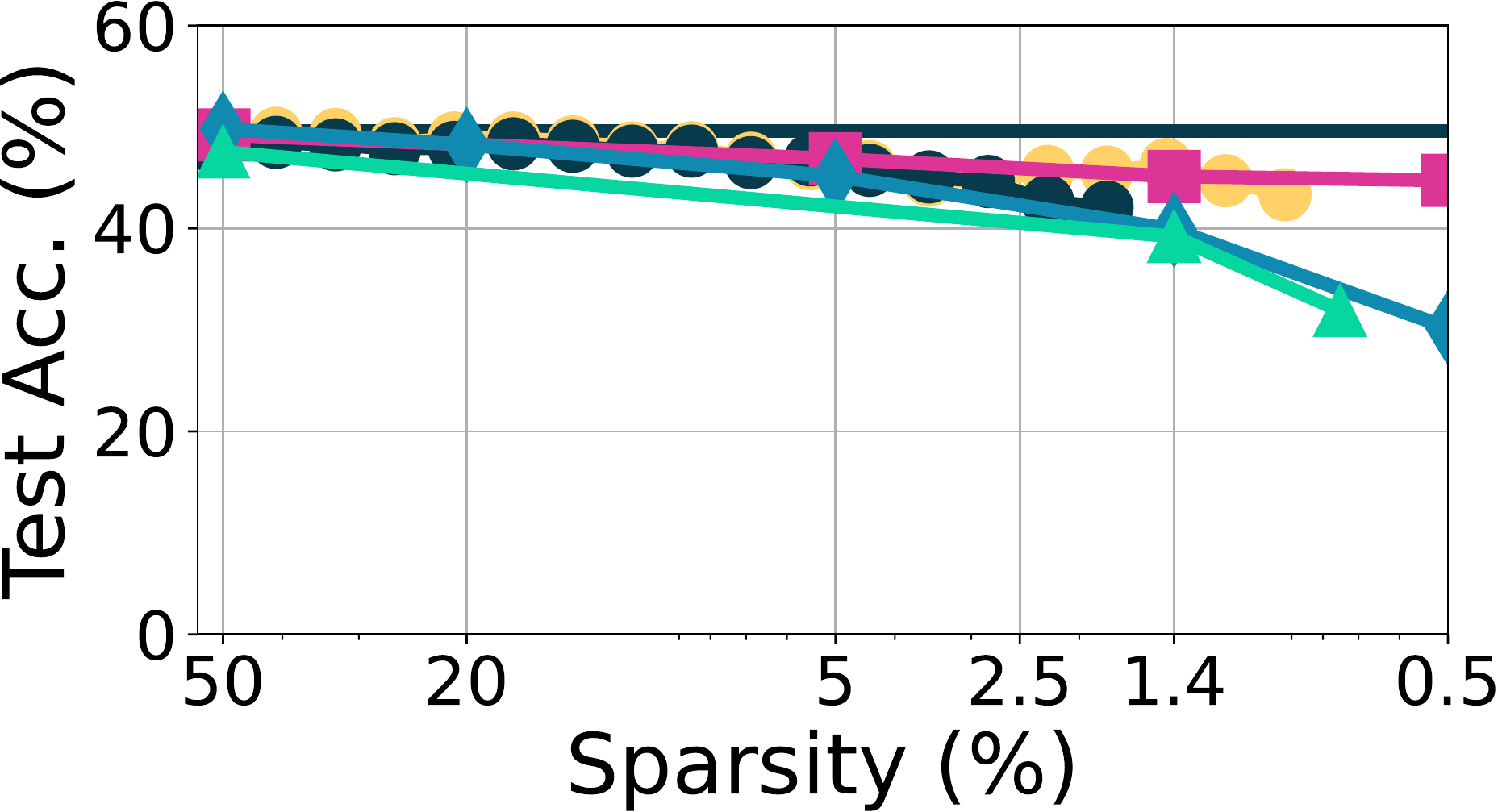}}
\subfloat[\centering{\scriptsize{TinyImageNet, ResNet-50}}]{\includegraphics[height=18mm
]{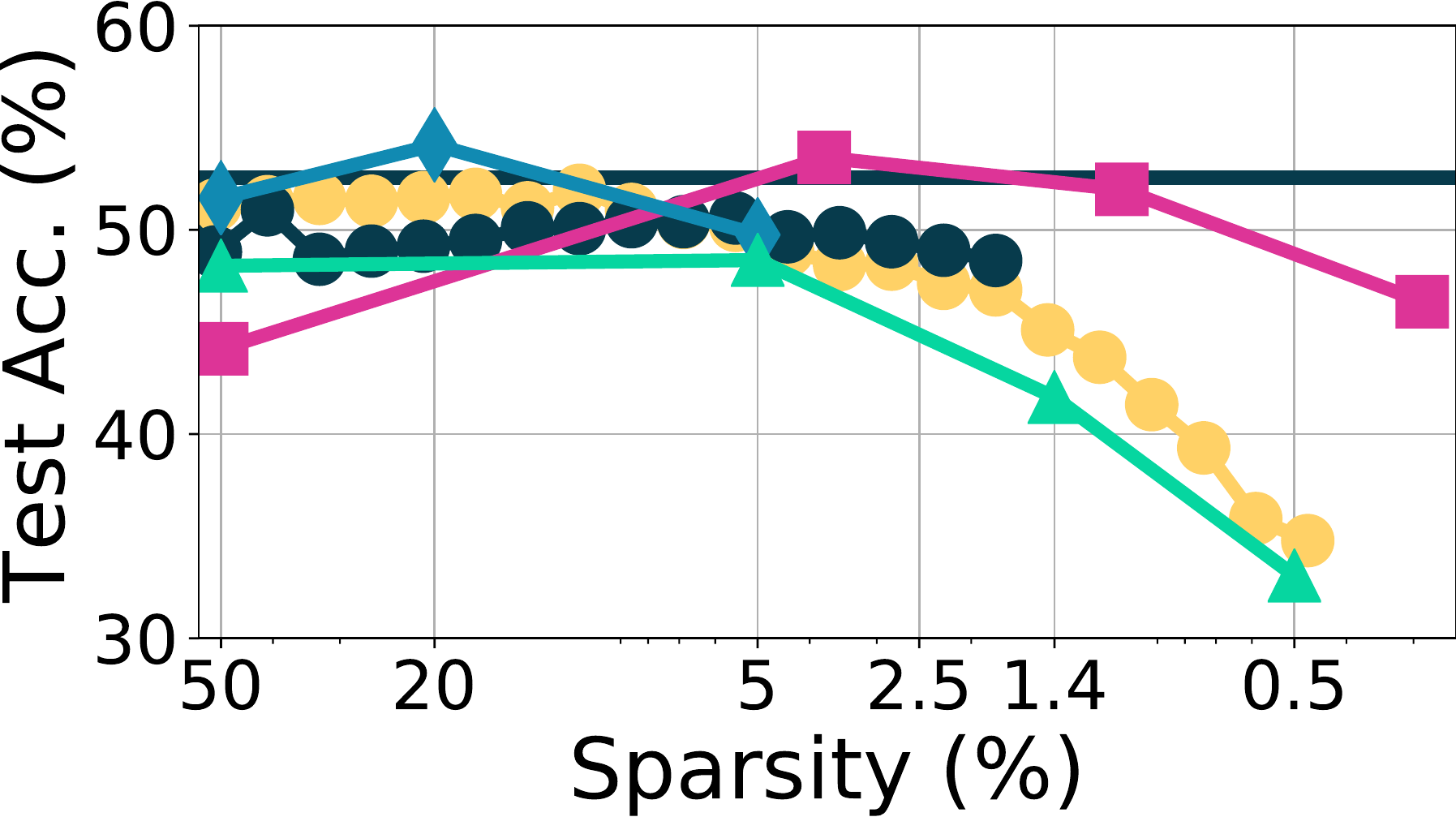}}
\subfloat[\centering{\scriptsize{Caltech-101, ResNet-50}}]{\includegraphics[height=18mm
]{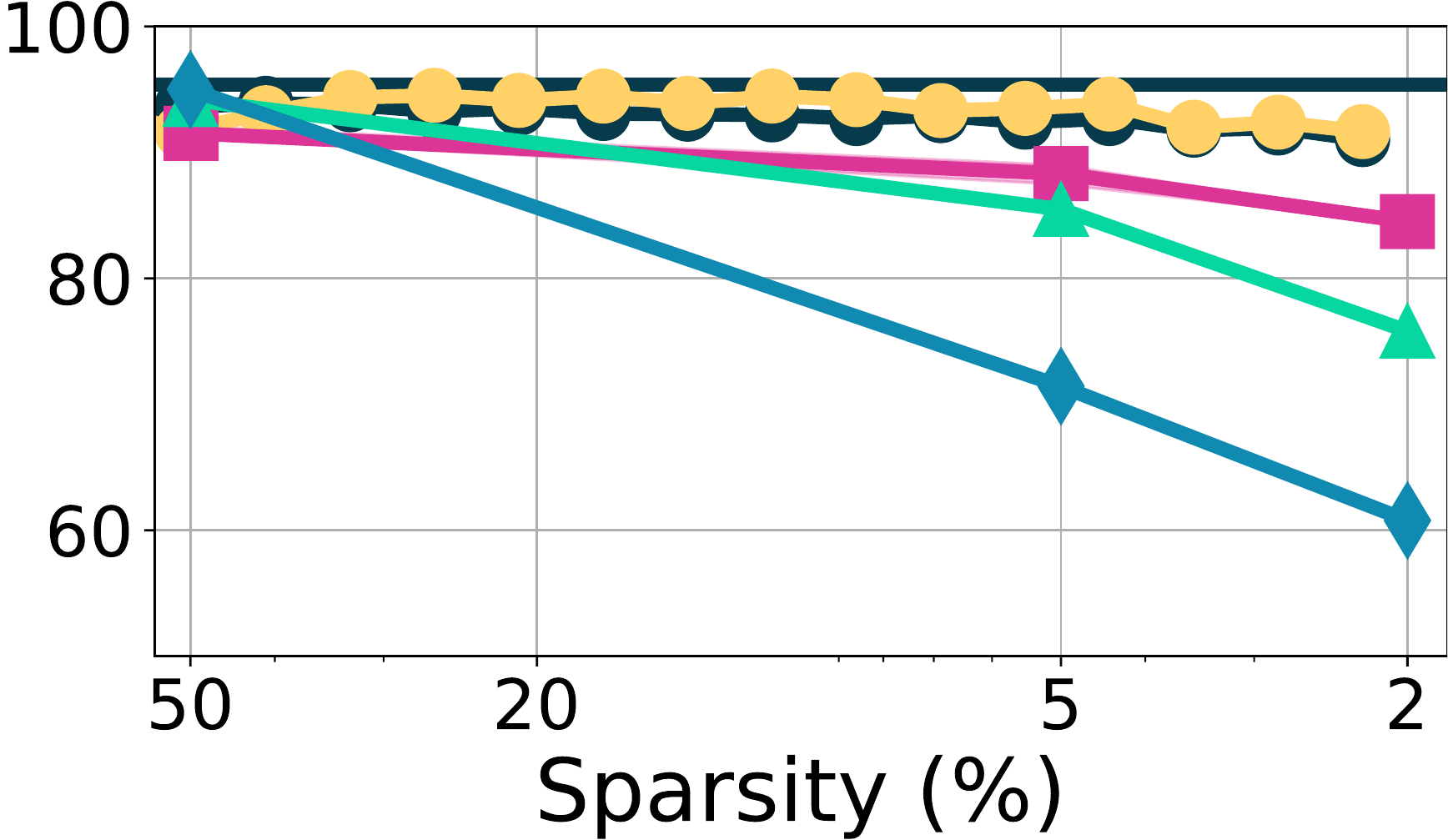}}
\\
\vspace{-7mm}
\subfloat[]{\includegraphics[height=18mm]{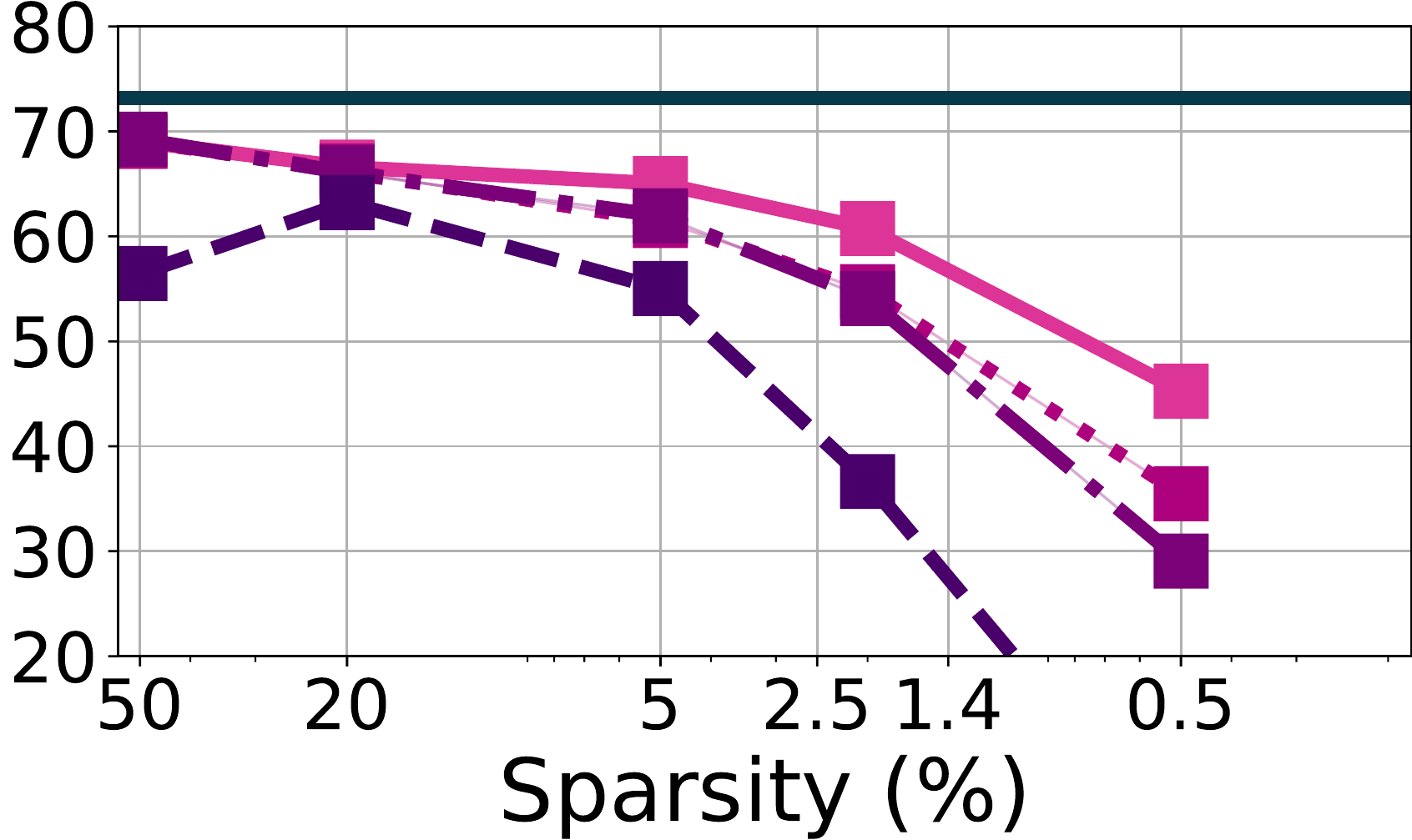}}
\subfloat[]{\includegraphics[height=18mm
]{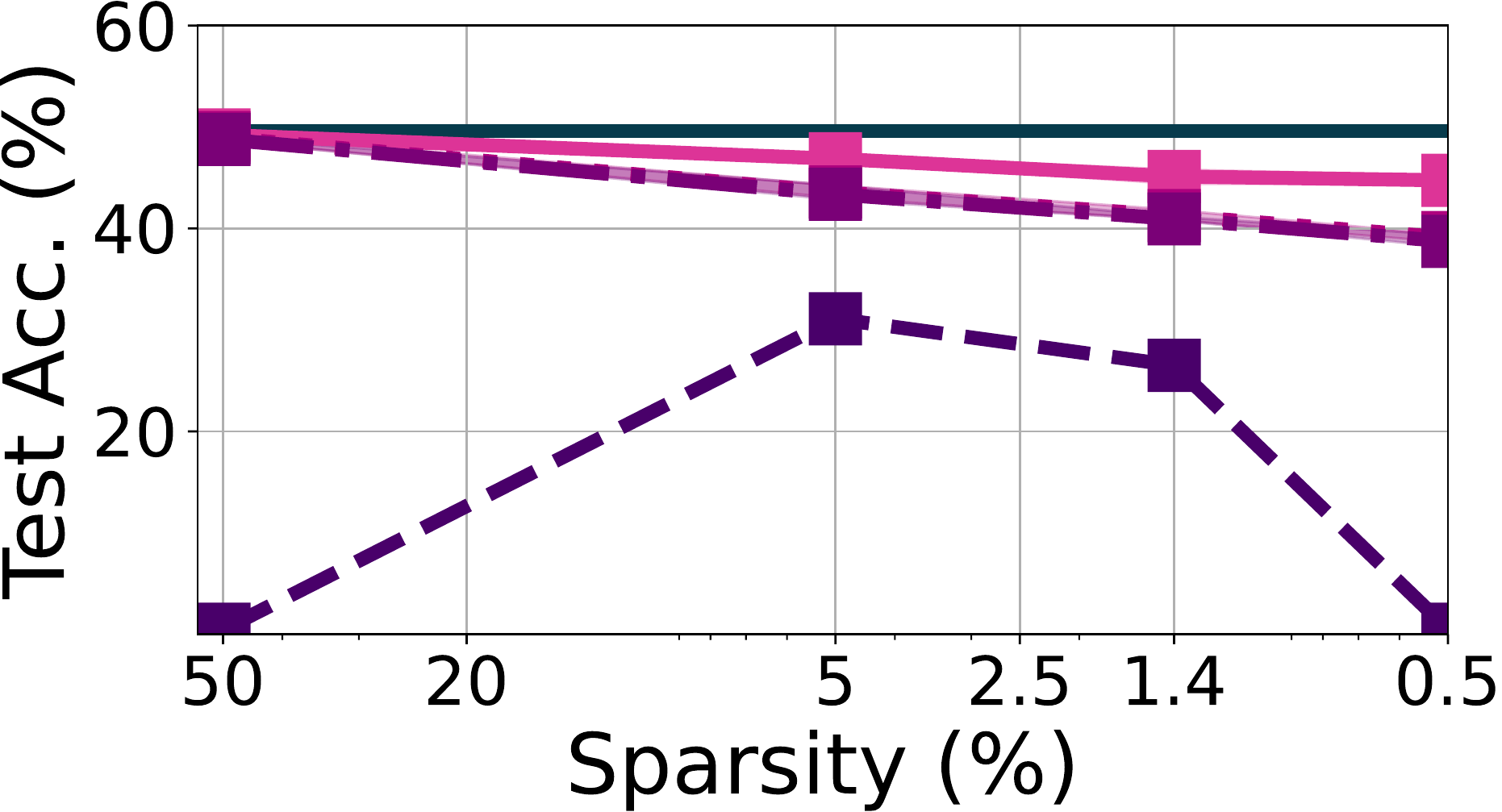}}
\subfloat[]{\includegraphics[height=18mm]{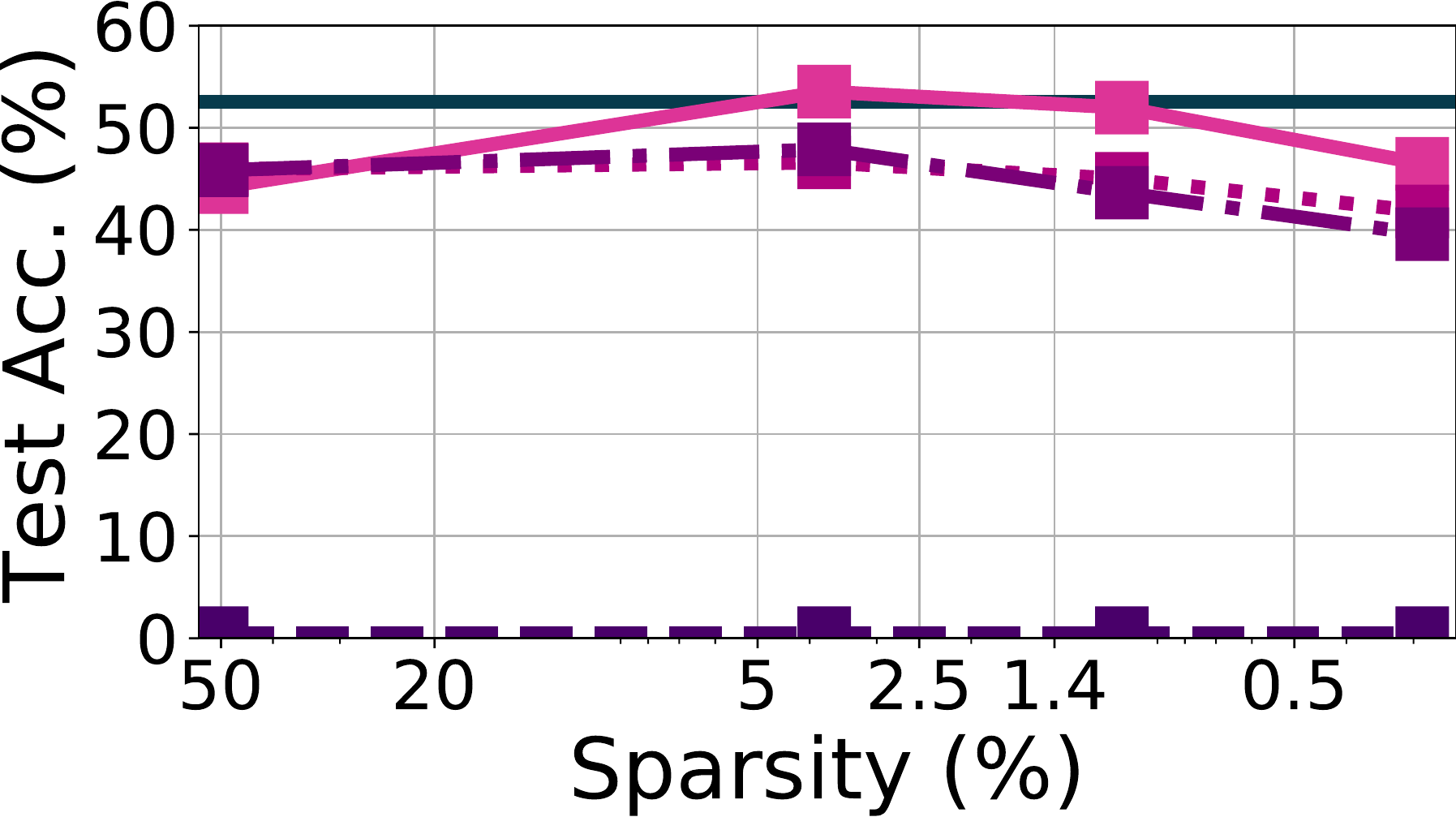}}
\subfloat[]{\includegraphics[height=18mm
]{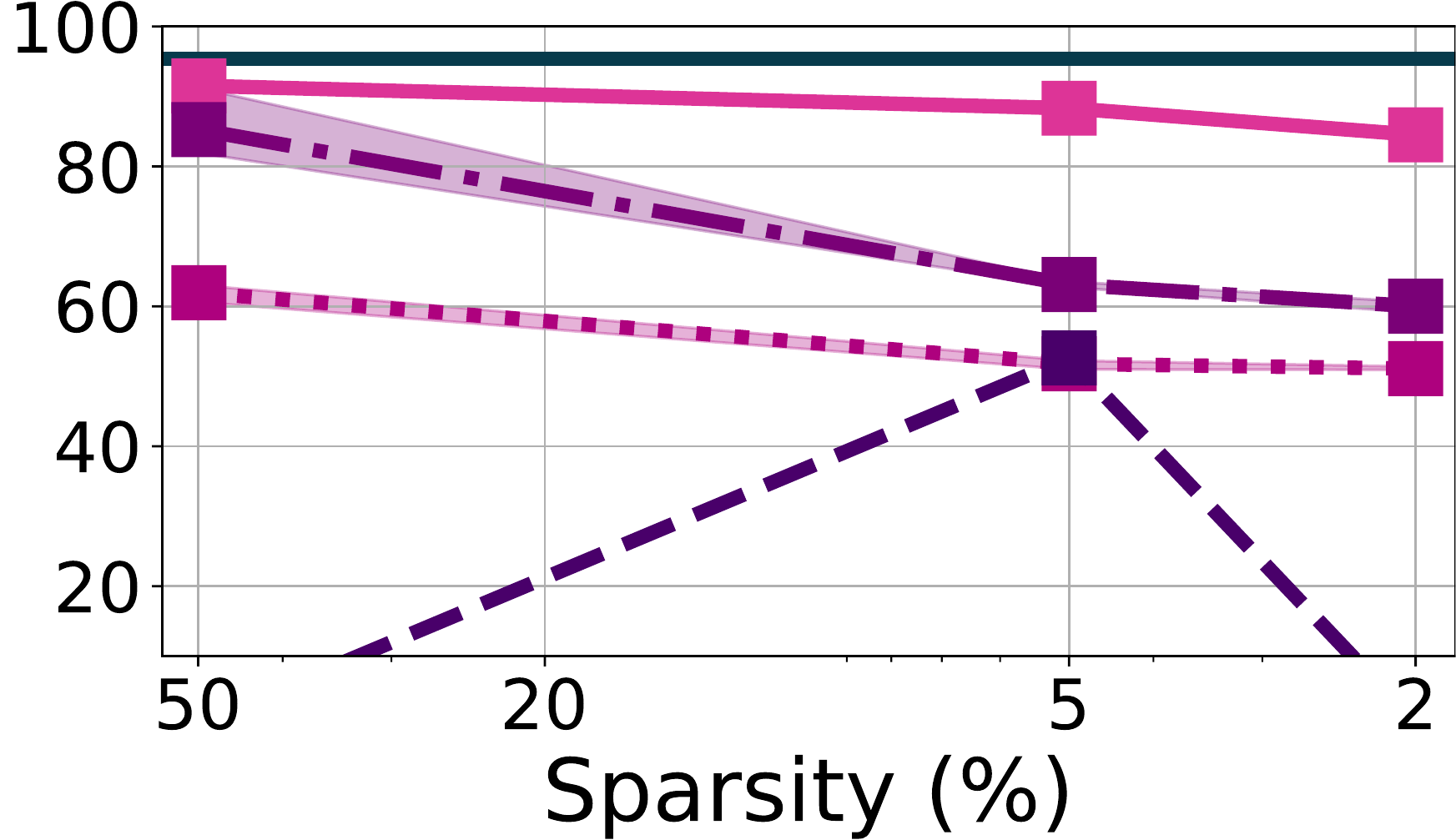}}
\\
\caption{Accuracy on image classification tasks on TinyImageNet, ImageNet and Caltech-101. For Caltech-101, we pruned a pre-trained ImageNet model (ResNet-50). 
Top: post-finetune accuracy, bottom: sanity check methods suggested in~\citet{frankle2020pruning} applied on \algo{}. 
}
\label{fig:img_lang}
\vspace{-0.2cm}
\end{figure}

\begin{wrapfigure}{r}{0.4\columnwidth}
\vspace{-0.5cm}
	\centering
	\includegraphics[width=0.39\textwidth]{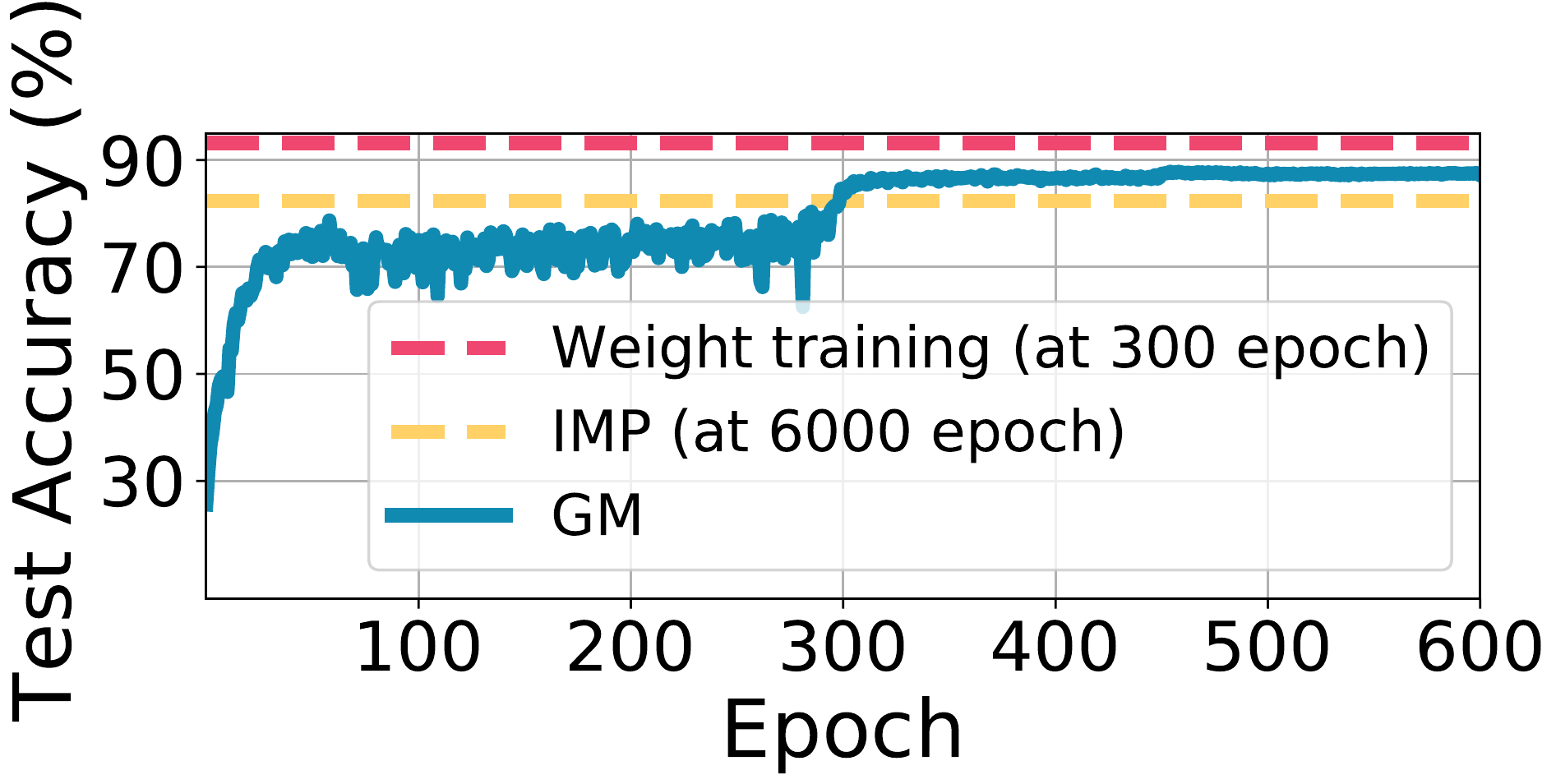}
	\caption{Convergence plot for CIFAR-10, MobileNet-V2 experiments, where we apply \algo{} for 300 epochs and then finetune the sparse model for another 300 epochs, to reach 1.16\% sparse model. We added the accuracy of weight training (dense model) and IMP (1.4\% sparse model) as a reference. Note that we compared with 300 epochs of weight training, and compared with IMP using 20 rounds of iterative pruning, i.e., 300 $\times$ 20 = 6000 epochs, to reach 1.4\% sparsity.
\algo{} achieves a higher accuracy than IMP despite its 19$\times$ shorter runtime to find a sparse subnetwork.}\label{fig:epoch_vs_acc}
\vspace{-0.8cm}
\end{wrapfigure}

\vspace{-3mm}
\paragraph{Tasks 2--4.}

Fig.~\ref{fig:img_lang} shows the sparsity-accuracy tradeoff for Tasks 2--4. Similar to Fig.~\ref{fig:cifar10}, the top row reports the accuracy \emph{after} weight training, and the bottom row contains the results of the sanity checks.

As shown in Fig.~\ref{fig:img_lang}a and Fig.~\ref{fig:img_lang}b, the results for \textbf{Task 2} show that (i) \algo{} achieves accuracy comparable to IMP as well as Renda et al. (IMP with learning rate rewinding) even in the sparse regime, (ii) \algo{} has non-trivial accuracy before finetuning (iii) \algo{} passes all the sanity checks, and (iv) \algo{} outperforms EP and SR.
These results show that \algo{} successfully finds rare gems even in the sparse regime for \textbf{Task 2}.

Fig.~\ref{fig:img_lang}c shows the result for \textbf{Task 3}. Unlike other tasks, \algo{} does not reach the post-finetune accuracy of IMP, but \algo{} enjoys over an 8\% accuracy gap compared with EP and SR. Moreover, the bottom row shows that \algo{} has over 20\% higher accuracy than the sanity checks below 5\% sparsity showing that the subnetwork found by \algo{} is unique in this sparse regime.

\subsection{Comparison to ProsPr}
\label{subsec:prospr}

\citet{alizadeh2021prospect} recently proposed a pruning at init method called ProsPr which utilizes meta-gradients through the first few steps of optimization to determine which weights to prune, thereby accounting for the ``trainability'' of the resulting subnetwork. In Table~\ref{table:prospr_comparison} we compare it against \algo{} on ResNet-20, CIFAR-10 and also run the (i) Random shuffling and (ii) Weight reinitialization sanity checks from \citet{frankle2020pruning}. We were unable to get ProsPr using their publicly available codebase to generate subnetworks at sparsity below $5\%$ and therefore chose that sparsity. Note that \algo{} produces a subnetwork that is higher accuracy despite being more sparse. After finetuning for 150 epochs, our subnetwork reaches $83.4\%$ accuracy while the subnetwork found by ProsPr only reaches $82.67\%$ after training for 200 epochs. More importantly, ProsPr does not show significant decay in performance after the random reshuffling or weight reinitialization sanity checks. Therefore, as \citet{frankle2020pruning} remark, it is likely that it is identifying good layerwise sparsity ratios, rather than a mask specific to the initialized weights.

\begin{table}[t]
    \caption{
    	We compare ProsPr~\cite{alizadeh2021prospect} vs \algo{} on ResNet-20, CIFAR-10 and run the random shuffling as well as the weight reinit sanity checks. Note that \algo{} produces a subnetwork that is higher accuracy despite being more sparse. Moreover, ProsPr does not show significant decay in performance after the sanity checks while \algo{} does. Therefore, it is likely that ProsPr is merely identifying good layerwise sparsity ratios.
	}
	\vspace{1mm}
    \centering
    \scriptsize
    \setlength{\tabcolsep}{4pt} %
    \renewcommand{\arraystretch}{0.5}
		 {
			\begin{tabular}{cccccc}
				\toprule {\shortstack{Algorithm}}
				& Sparsity & Accuracy after finetune  & Accuracy after Random shuffling & Accuracy after Weight reinitialization
				\bigstrut\\
				\midrule
				ProsPr
				& 5\%
				& 82.67\%
				& 82.15\%
				& 81.64\%
				\bigstrut\\
				\textbf{\algo{}} 
				& \textbf{3.72\%}
				& \textbf{83.4\%}
				& \textbf{78.73\%}
				& \textbf{78.6\%}
				\bigstrut\\
				\bottomrule
			\end{tabular}}%
    \vspace{2mm}	
	\label{table:prospr_comparison}
	\vspace{-0.5cm}
\end{table}

\subsection{Observations on \algo{}}
\label{subsec:observations}

\vspace{-0.1cm}
\paragraph{Convergence of accuracy and sparsity.}
Fig.~\ref{fig:epoch_vs_acc} shows how the accuracy of \algo{} improves as training progresses, for MobileNet-V2 on CIFAR-10 at sparsity $1.4\%$.
This shows that \algo{}, reaches high accuracy even early in training, and can be finetuned to accuracy higher than that of IMP (which requires $19\times$ the runtime than our algorithm).

\begin{figure*}[ht] 
\centering
\includegraphics[width=0.99\textwidth]{figures/Legend_with_Renda_Latest_withborder.pdf}%
\\
\vspace{-0.4cm}\hspace{-1mm}
\subfloat[ResNet-20]{
\includegraphics[height=19mm]{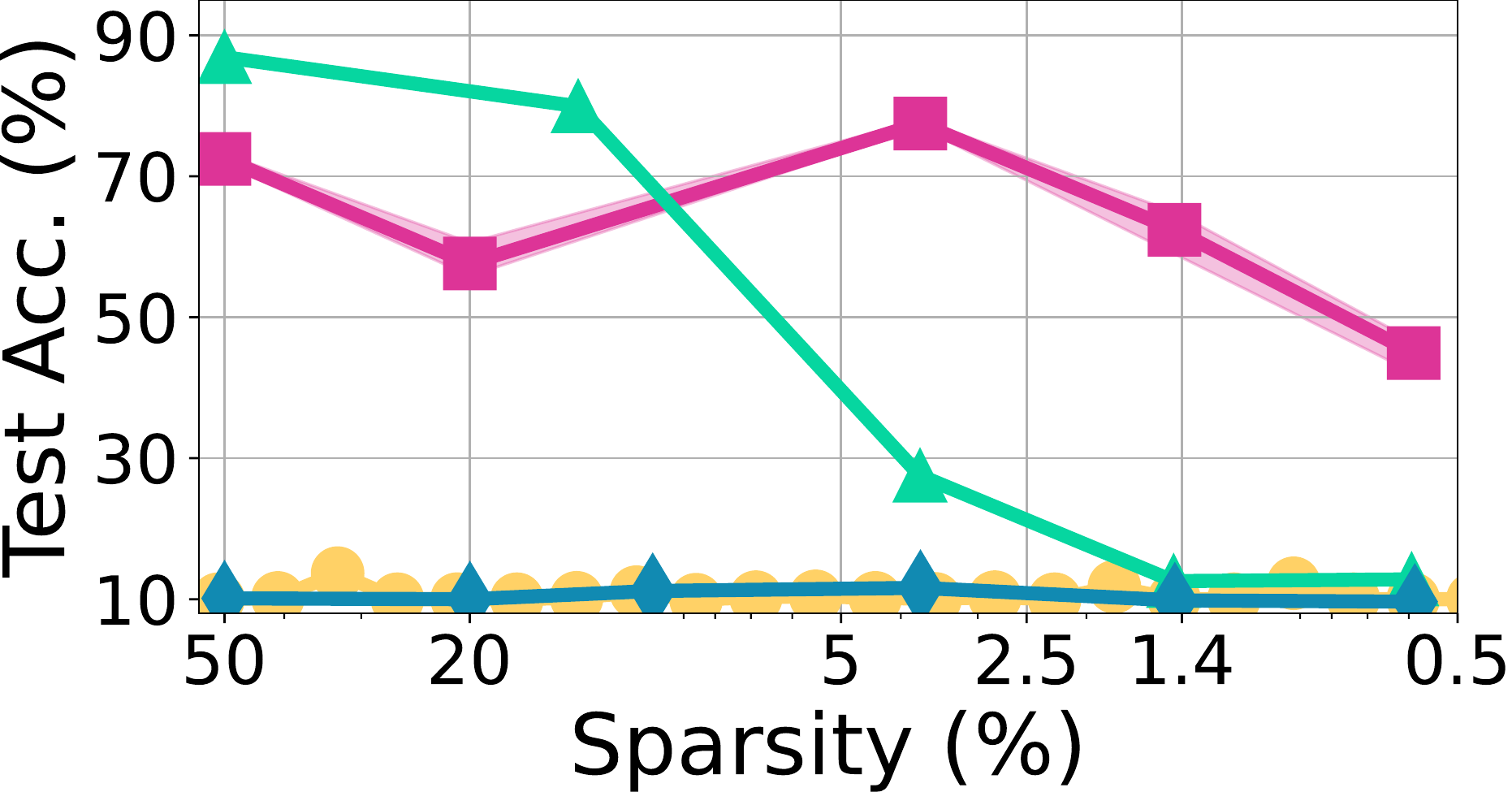}
}
\subfloat[MobileNet-V2]{
\includegraphics[height=19mm]{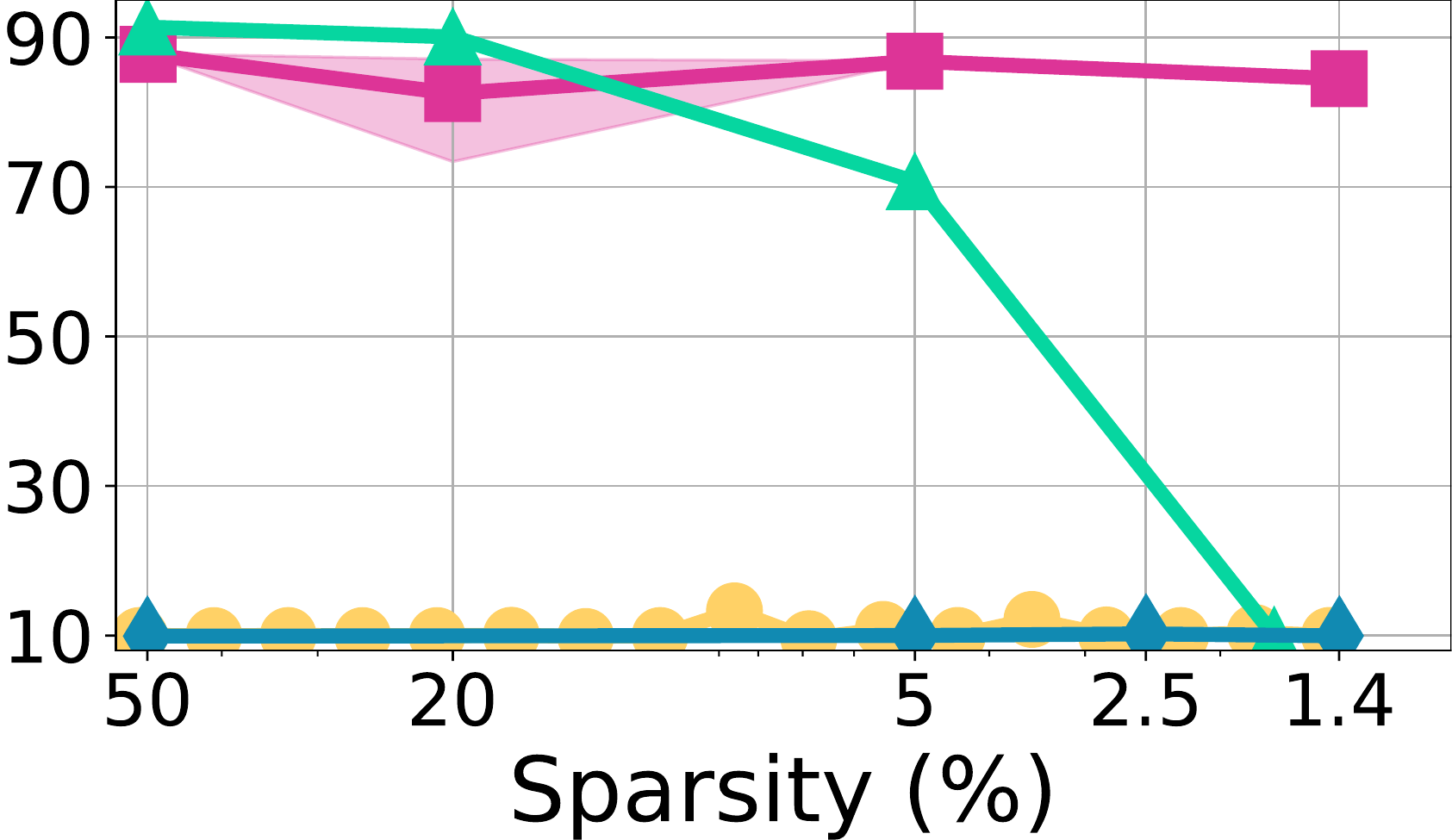}
}
\subfloat[VGG-16]{
\includegraphics[height=19mm]{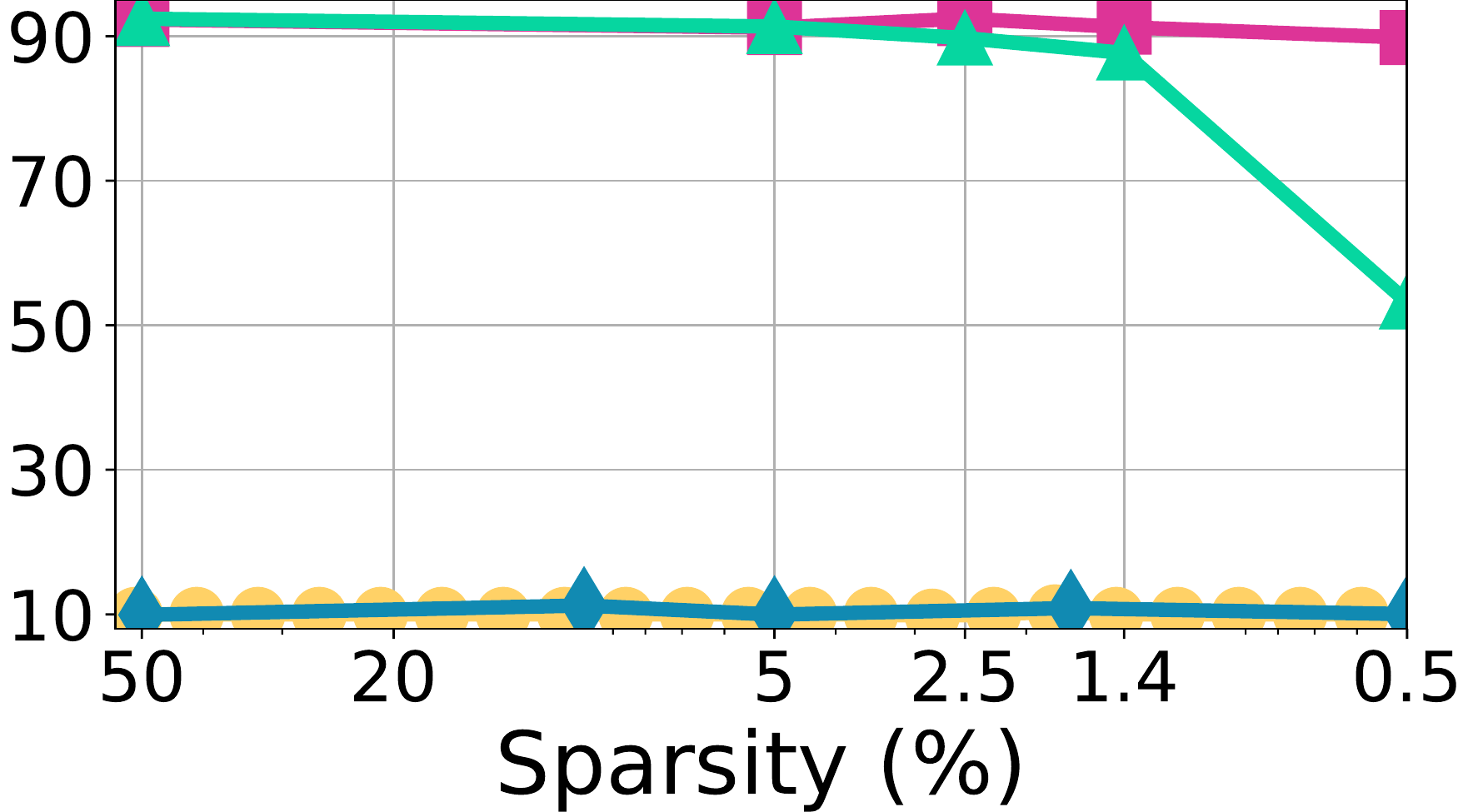}
}
\subfloat[WideResNet-28-2]{
\includegraphics[height=19mm]{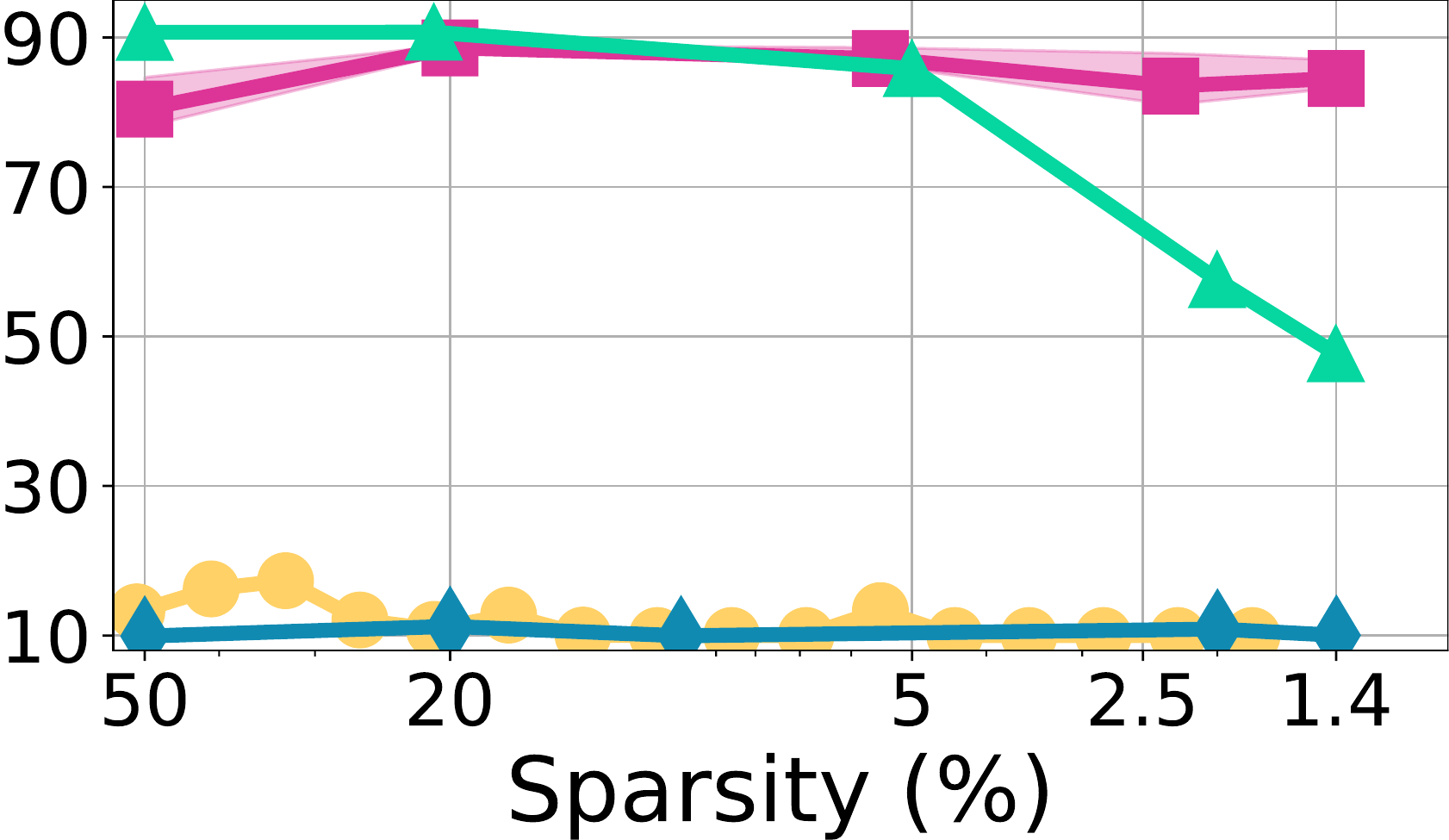}
}
\caption{
Performance of different pruning algorithms before finetuning on CIFAR-10 for benchmark networks. \algo{} finds subnetworks that already have reasonably high accuracy even before weight training. Note that, while IMP and SR have scarcely better than random guessing at initialization, subnetworks found by \algo{} typically perform even better than EP, especially in the sparse regime.
}
\label{fig:cifar10_before_ft}
\vspace{-0.2cm}
\end{figure*}

\begin{wrapfigure}{r}{0.5\columnwidth}
\vspace{-0.4cm}
	\centering
	\includegraphics[width=0.45\textwidth]{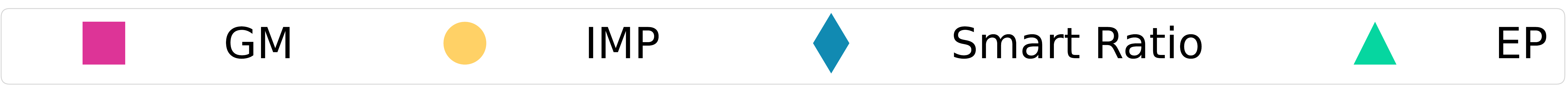}\\
	\subfloat[$50\%$ Sparsity]{\includegraphics[width=0.25\textwidth]{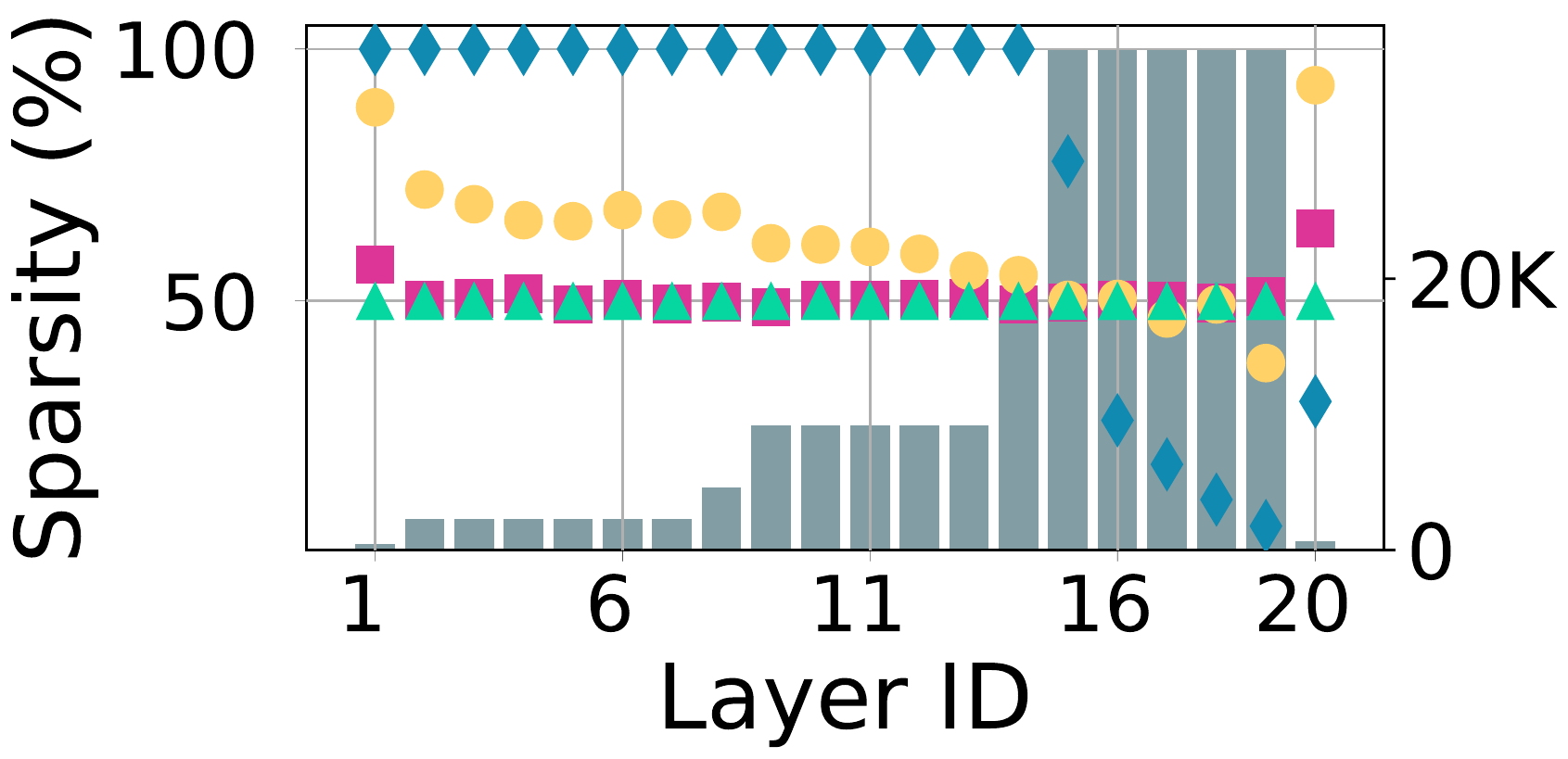}}
\subfloat[$3.72\%$ Sparsity]{\includegraphics[width=0.25\textwidth]{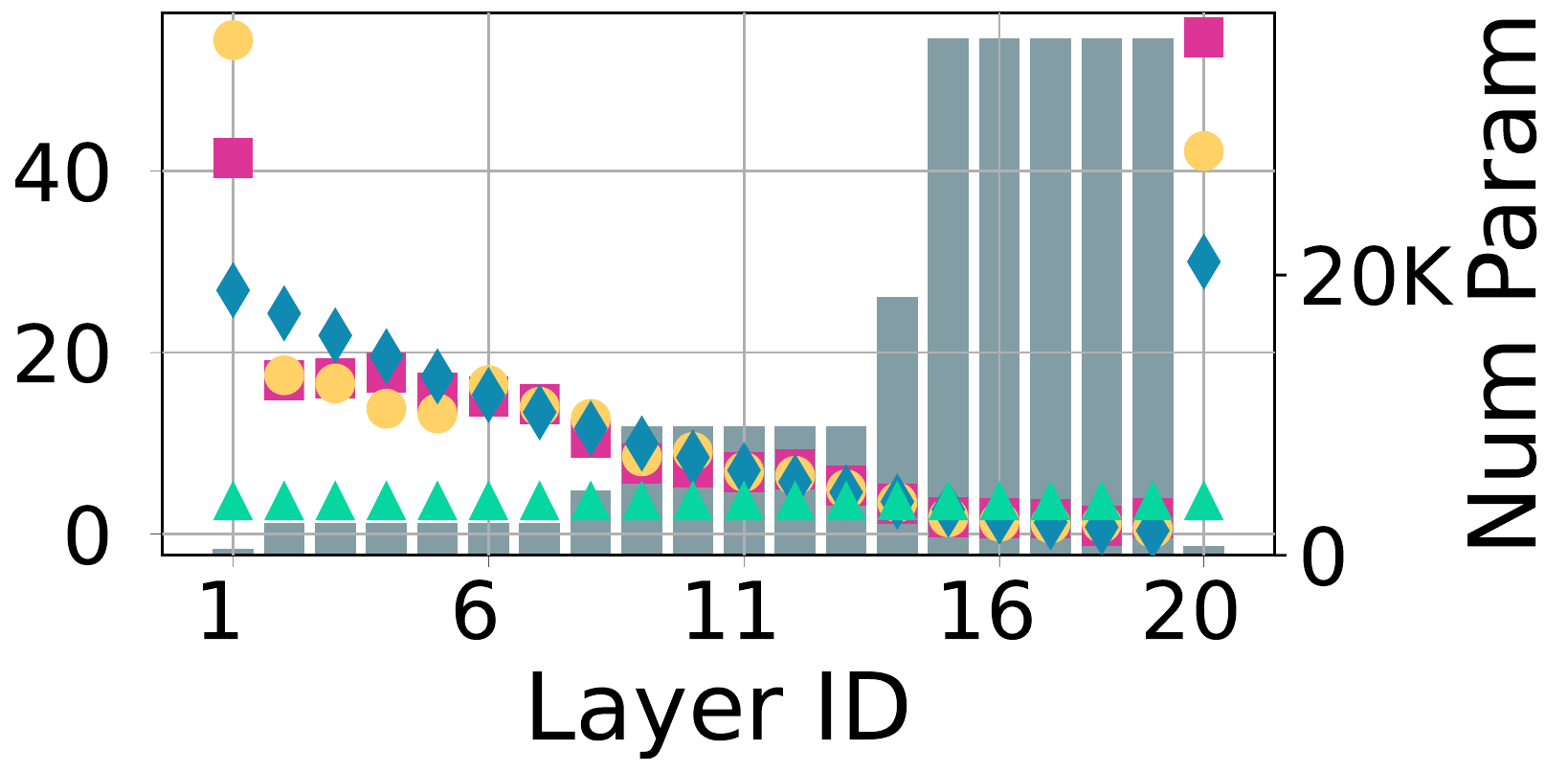}}
\\
\vspace{-2mm}
\subfloat[$1.4\%$ Sparsity]{\includegraphics[width=0.25\textwidth]{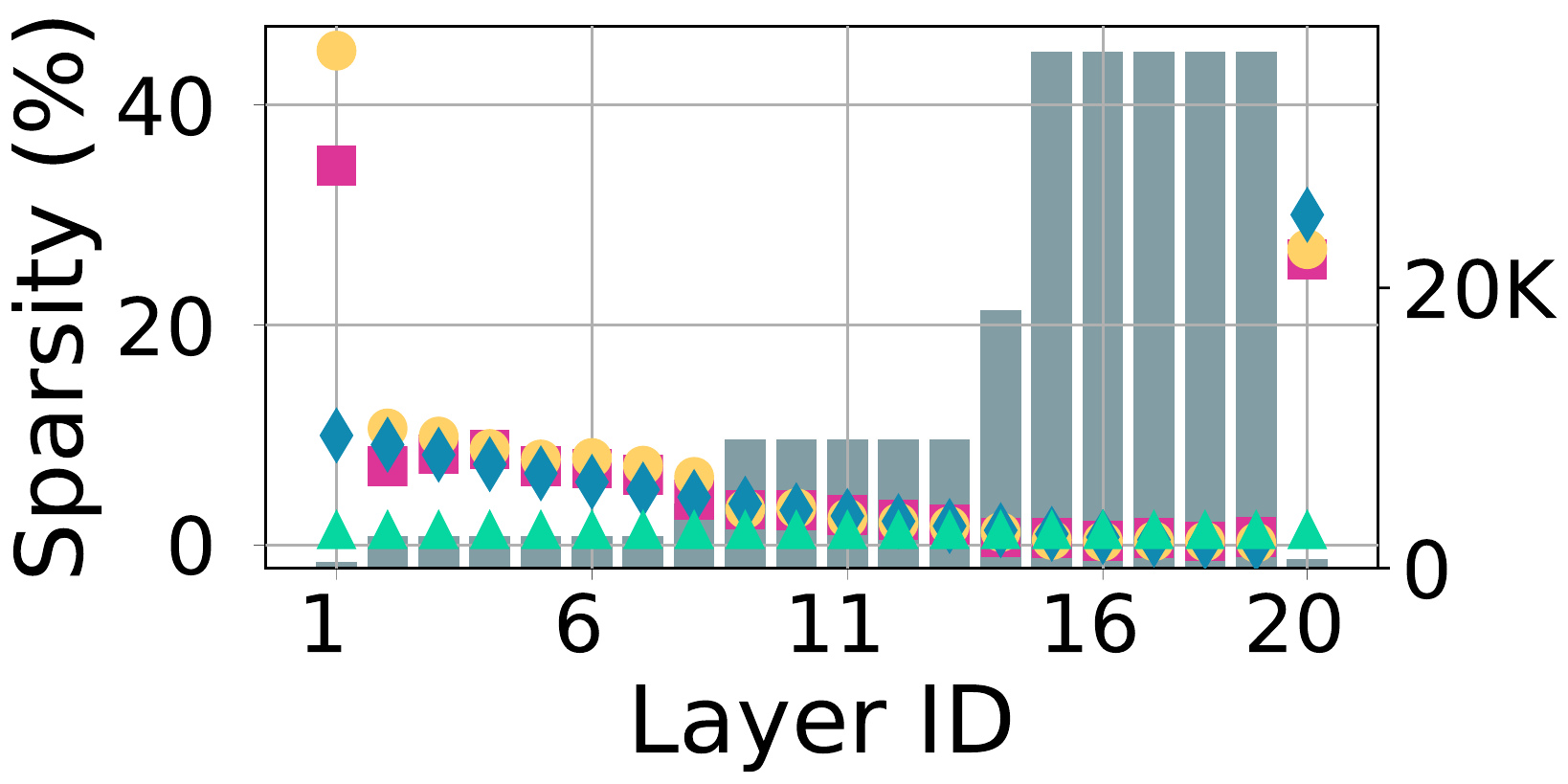}}
\subfloat[$0.59\%$ Sparsity]{\includegraphics[width=0.25\textwidth]{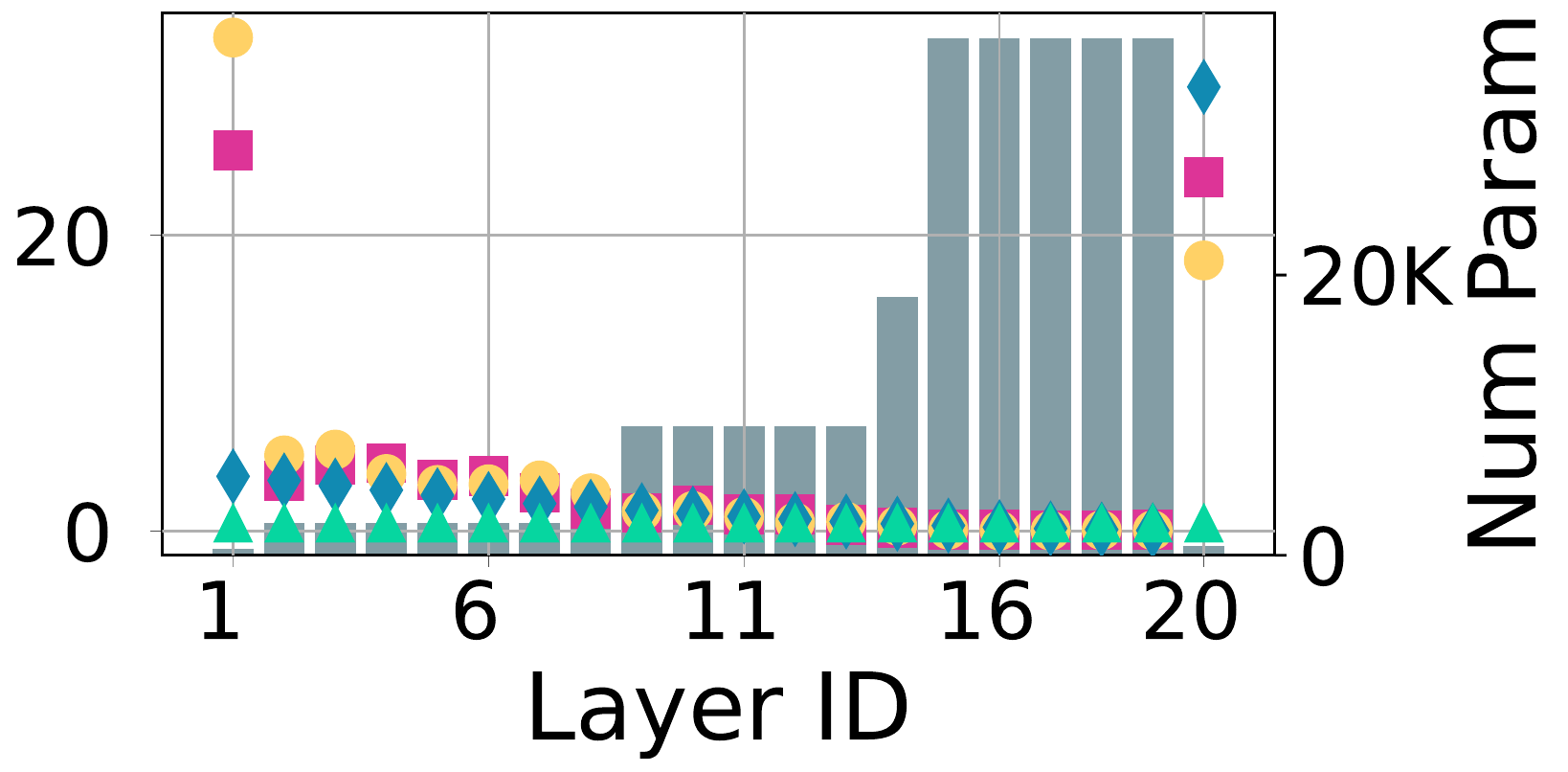}}
\\
	\caption{The layerwise sparsity for ResNet-20 pruned by \algo{}, IMP, Smart Ratio, and EP. The dark bar is the layerwise number of parameters. Both \algo{} and IMP save the most portion of parameter in the first layer and the last layer.}\label{fig:cifar11}
\vspace{-0.6cm}
\end{wrapfigure}

\paragraph{High pre-finetune accuracy.}
As shown in Fig.~\ref{fig:cifar10_before_ft}, \algo{} finds subnetworks at initialization that have a reasonably high accuracy even before the weight training, e.g., above 90\% accuracy for 1.4\% sparsity in VGG-16, and 85\% accuracy for 1.4\% sparsity in MobileNet-V2. Note that, in contrast, IMP and SR have accuracy scarcely better than random guessing at initialization. Clearly, \algo{} fulfills its objective in maximizing accuracy before finetuning and therefore finds rare gems -- lottery tickets at initialization which already have high accuracy.

\vspace{-0.1cm}
\paragraph{Limitations of \algo{}.}
We observed that in the dense regime (50\% sparsity, 20\% sparsity), \algo{} sometimes performs worse than IMP. While we believe that this can be resolved by appropriately tuning the hyperparameters, we chose to focus our attention on the sparse regime. We would also like to remark that \algo{} is fairly sensitive to the choice of hyperparameters and for some models, we had to choose different hyperparameters for each sparsity to ensure optimal performance. Though this occurs rarely, we also find that an extremely aggressive choice of $\lambda$ can lead to \emph{layer-collapse} where one or more layers gets pruned completely. This happens when all the scores $\vp$ of that layer drop below $0.5$.

\vspace{-0.2cm}
\paragraph{Layer-wise sparsity.}
We compare the layer-wise sparsity pattern of different algorithms for ResNet-20 trained on CIFAR-10 in Fig.~\ref{fig:cifar11}. Both \algo{} and IMP spend most of their sparsity budget on the first and last layers. By design, SR assigns $30\%$ sparsity to the last layer and the budget decays smoothly across the others. EP maintains the target sparsity ratio at each layer and therefore is always a horizontal line.

\vspace{-0.2cm}
\paragraph{How does \algo{} resolve EP's failings?}

\begin{table}[t]
    \caption{
    	We construct different variants of EP and compare their performance with \algo{}, for ResNet-20, CIFAR-10, 0.59\% sparsity. We establish that having a global score metric and gradually pruning is key to improved performance.
	}
	\vspace{1mm}
    \centering
    \scriptsize
    \setlength{\tabcolsep}{4pt} %
    \renewcommand{\arraystretch}{0.5}
		 {
			\begin{tabular}{cccccc}
				\toprule {\shortstack{Pruning\\Method}}
				& EP & Global EP  & \shortstack{Global EP with\\ Gradual Pruning} & \shortstack{Global EP with\\ Gradual Pruning and Regularization} & \algo{}
				\bigstrut\\
				\midrule
				\textbf{Pre-finetune acc (\%)} 
				& 19.57
				& 22.22
				& 31.56
				& 19.67
				& \textbf{45.30}
				\bigstrut\\
				\textbf{Post-finetune acc (\%)} 
				& 24.47
				& 34.42
				& 63.54
				& 63.72
				& \textbf{66.15}
				\bigstrut\\
				\bottomrule
			\end{tabular}}%
	\label{table:EP_to_GM}
	\vspace{-0.4cm}
\end{table}

An open problem from \citet{ramanujan2020s} is why the subnetworks found by EP are not fine-tunable. While \gm{} is significantly different from EP, it is reasonable to ask which modification allowed it to find lottery tickets without forgoing high accuracy at initialization.
Table~\ref{table:EP_to_GM} shows how we can modify EP to improve the pre/post-finetune performance, for ResNet-20, CIFAR-10 at 0.59\% sparsity. Here, we compare EP, \algo{}, as well as three EP variants that we construct. (i) \emph{(Global EP)} is a modification where the bottom-$k$ scores are pruned globally, not layer-wise. This allows the algorithm to trade-off sparsity in one layer for another. (ii) \emph{(Gradual pruning)} reduces the parameter $k$ gradually as opposed to setting it to the target sparsity from the beginning. (iii) \emph{(Regularization)}: we add an $L_2$ term on the score $p$ of the weights to encourage sparsity.
The results indicate that \emph{global pruning} and \emph{gradual pruning} significantly improve both the pre and post-finetune accuracies of EP. Adding regularization does not improve the performance significantly. Finally, adding all three features to EP allows it to achieve 63.72\% accuracy, while \algo{} reaches 66.15\% accuracy. It is important to note that even with all three features, EP is inherently different from \gm{} in how it computes the \emph{supermask} based on the scores. But we conjecture that aggressive, layerwise pruning is the key reason for EP's failings.

\paragraph{Applying \algo{} for longer periods.}

\begin{table}[H] 
\vspace{-0.6cm}
    \caption{
	Comparison of \algo{} and its longer version, for ResNet-20, CIFAR-10, 1.4\% sparsity. \textsc{Long} \algo{}, when given the same number of epochs improves post-finetune accuracy by 1.5\%, rivaling the performance of \citet{renda2020comparing}
	}
    \vspace{1mm}
    \centering
    \scriptsize
    \setlength{\tabcolsep}{4pt} %
    \renewcommand{\arraystretch}{0.5}
		 {
			\begin{tabular}{ccccc}
				\toprule \textbf{Method}
				& GM (cold) & Long GM (cold) & IMP (warm) & Renda et al. (pruning after training)
				\bigstrut\\
				\midrule
				\textbf{Number of Epochs} 
				& 300
				& 3000
				& 3000
				& 3000
				\bigstrut\\
				\textbf{Accuracy (\%)} 
				& 77.89
				& 79.50
				& 74.52
				& 80.21
				\bigstrut\\
				\bottomrule
			\end{tabular}}%
	\label{table:long_GM}
	\vspace{-0.5cm}
\end{table}

Recall that \algo{} uses 19 $\times$ fewer training epochs than iterative train-prune-retrain methods like IMP~\citep{frankle2020linear} and Learning rate rewinding (\citet{renda2020comparing}), to find a subnetwork at $1.4\%$ sparsity which can then be trained to high accuracy. Here, we consider a long version of \algo{} to see if it can benefit if it is allowed to run for longer. 
Table~\ref{table:long_GM} shows the comparison of post-finetune accuracy for \algo{}, \textsc{Long} \algo{}, IMP and \citet{renda2020comparing} tested on ResNet-20, CIFAR-10, sparsity=1.4\% setting.
Conventional \algo{}, applies iterative freezing every $5$ epochs to arrive at the target sparsity in $150$ epochs. \textsc{Long} \algo{} instead prunes every $150$ epochs and therefore reaches the target sparsity in $3000$ epochs.

We find that applying \algo{} for longer periods improves the post-finetune accuracy in this regime by 1.5\%.
This shows that given equal number of epochs, \algo{}, which prunes at initialization, can close the gap to Learning rate rewinding~\cite{renda2020comparing} which is a prune-after-training method.
\vspace{-2mm}\section{Conclusion}
In this work, we resolve the open problem of pruning at initialization by proposing \algo{} that finds \emph{rare gems} -- lottery tickets \emph{at initialization} that have non-trivial accuracy even before finetuning and accuracy rivaling prune-after-train methods after finetuning. Unlike other methods, subnetworks found by \algo{} pass all known sanity checks and baselines. Moreover, we show that \algo{} is competitive with IMP despite not using warmup and up to 19$\times$ faster.

\bibliography{refine_gems}
\bibliographystyle{icml2022}

\appendix

{\hypersetup{linkcolor=black}
\parskip=0em
\renewcommand{\contentsname}{Contents of the Appendix}
\tableofcontents
\addtocontents{toc}{\protect\setcounter{tocdepth}{3}}
}

\newcommand{\blocka}[2]{\multirow{3}{*}{\(\left[\begin{array}{c}\text{3$\times$3, #1}\\[-.1em] \text{3$\times$3, #1} \end{array}\right]\)$\times$#2}
}
\newcommand{\blockb}[3]{\multirow{3}{*}{\(\left[\begin{array}{c}\text{1$\times$1, #2}\\[-.1em] \text{$3\times$3, #2}\\[-.1em] \text{1$\times$1, #1}\end{array}\right]\)$\times$#3}
}

\section{Experimental Setup}\label{sec:exp_setup}

In this section, we introduce the datasets (\ref{appdx:dataset}) and models (\ref{appdx:model}) that we used in the experiments. We also report the detailed hyperparameter choices (\ref{appdx:config}) of \algo{} each network and sparsity level. For competing methods, we used hyperparameters used by the original authors whenever possible. In other cases, we tried SGD (with momentum) and Adam optimizers, initial learning rate (LR) of $\eta$, 0.1$\eta$, 10$\eta$, and cosine/multi- step LR decay, where $\eta$ is the best LR for weight training. All of our experiments are run using PyTorch 1.10 on Nvidia 2080 TIs and Nvidia V100s.
 
\subsection{Dataset}\label{appdx:dataset}
\vspace{-3mm}
In the experiments, we demonstrate the performance of \algo{} across various datasets. For each dataset, we optimize the training loss, and tune hyperparameters based on the validation accuracy. The test accuracy is reported for the best model chosen based on the validation accuracy. 

\vspace{-3mm}
\paragraph{CIFAR-10.} CIFAR-10 consists of $60,000$ images from $10$ classes, each with size $32 \times 32$, of which $50000$ images are used for training, and $10,000$ images are for testing~\citep{krizhevsky2009learning}. For data processing, we follow the standard augmentation: normalize channel-wise, randomly horizontally flip, and random cropping. For hyperparameter tuning we randomly split the train set into $45000$ train images and retain $5000$ images as the validation set. Once the hyperparameters are chosen, we retrain on the full train set and report test accuracy.

\vspace{-3mm}
\paragraph{TinyImageNet.} TinyImageNet contains $100000$ images of $200$ classes ($500$ each class), which are downsized to $64 \times 64$ colored images. Each class has $500$ training images, $50$ validation images and $50$ test images. Augmentation includes normalizing, random rotation and random flip. Train set, validation set, and test set are provided.

\vspace{-3mm}
\paragraph{Caltech-101.} Caltech-101 contains figures of objects from $101$ categories. There are around 40 to 800 images per category, and most categories have about 50 images~\citep{fei2004learning}. The size of each image is roughly $300 \times 200$ pixels. When processing the image, we resize each figure to $224 \times 224$, and normalize it across channels. We split $20\%$ of the data to be test set, and in the remaining training set, we retain $25\%$ as the validation set, giving us train/val/test = $60\%/20\%/20\%$ split.

\subsection{Model}\label{appdx:model}
\vspace{-3mm}
Unless otherwise specified, in all of our experiments experiments, we use Non-Affine BatchNorm, and disable bias for all the convolution and linear layers. We find that most implementations of pruning algorithms instead use them and merely ignore them while pruning and while computing sparsity. While they do not alter sparsity by much (since there are few parameters when compared to weights), we still find this to be inaccurate. Moreover, it is not obvious how to prune biases -- should they be treated as weight? Should they be treated as a different set of parameters? In order to make sure we compare all the baselines on the same platform, we decided that eliminating them was the fair choice (As \cite{fischer2021towards} describe, pruning with biases is an interesting problem but needs to be handled slightly carefully). We use uniform initialization for scores, signed constant initialization~\citep{ramanujan2020s} for weight parameters for \algo{} while dense training initializes the weights using the standard Kaiming normal initializaiton~\citep{he2015delving}.

The networks that we used in our experiments are summarized as follows.

\vspace{-3mm}
\paragraph{ResNet-18, ResNet-20 and ResNet-50~\citep{he2016deep}.} We follow the standard ResNet architecture. ResNet-20 is designed for CIFAR-10 while ResNet-18 and ResNet-50 are for ImageNet. In Table~\ref{arch:resnet-typed}, we use the convention [kernel $\times$ kernel, output]$\times$(times repeated) for the convolution layers found in ResNet blocks. We base our implementation on the following GitHub repository\footnote{\url{https://github.com/akamaster/pytorch_resnet_cifar10/blob/master/resnet.py}}.

\paragraph{WideResNet-28-2~\citep{zagoruyko2016wide}.} We base our implementation on the following GitHub repository\footnote{\url{https://github.com/xternalz/WideResNet-pytorch/blob/master/wideresnet.py}}. Architecture details can be found in Table~\ref{arch:resnet-typed}.

\begin{table}[ht]
\caption{ResNet architecture used in our experiments. The output layer of the network is changed according to the dataset. For example, ResNet50 is used for both TinyImageNet and ImageNet so we changed the output dimension to 200 and 1000 respectively.}
\footnotesize
\vspace{1mm}
\centering
\begin{tabular}{ccccc} %
\toprule
\textbf{Layer} & \textbf{ResNet-20} & \textbf{ResNet-18} & \textbf{ResNet-50} & \textbf{WideResNet-28-2}
\bigstrut\\\midrule
\multirow{3}{*}{Conv 1} & 3$\times$3, 16 & 3$\times$3, 64 & 7$\times $7, 64 & 3$\times$3, 16\\
& padding 1 & padding 3 & padding 3 & padding 1\\
& stride 1 & stride 2 & stride 2 & stride 1 \\
& & \multicolumn{2}{c}{Max Pool, kernel size 3, stride 2, padding 1} &\\
\midrule
\multirow{3}{*}{\shortstack{Layer\\stack 1}} & \blocka{16}{3} & \blocka{64}{2} & \blockb{256}{64}{3} & \blocka{32}{$4$} \\
& & & &\\
& & & &\\
\midrule
\multirow{3}{*}{\shortstack{Layer\\stack 2}} & \blocka{32}{3} & \blocka{128}{2} & \blockb{512}{128}{4} & \blocka{64}{4} \\
& & & &\\
& & & &\\
\midrule
\multirow{3}{*}{\shortstack{Layer\\stack 3}} & \blocka{64}{3} & \blocka{256}{2} & \blockb{1024}{256}{6} & \blocka{128}{4} \\
& & & &\\
& & & &\\
\midrule
\multirow{3}{*}{\shortstack{Layer\\stack 4}} &  & \blocka{512}{2} & \blockb{2048}{512}{3} & \\
& - & & & -\\
& & & &\\
\midrule
\multirow{2}{*}{FC} & Avg Pool, kernel size 8 & \multicolumn{2}{c}{Adaptive Avg Pool, output size $(1, 1)$} & Avg Pool, kernel size 8 \\
& $64 \times \textsc{n\_classes}$ & $512 \times \textsc{n\_classes}$ & $2048 \times \textsc{n\_classes}$ & $128 \times \textsc{n\_classes}$\\
\bottomrule
\end{tabular}
\label{arch:resnet-typed}
\end{table}

\paragraph{VGG-16~\citep{simonyan2014very}.} In the original VGG-16 network, there are 13 convolution layers and 3 FC layers (including the last linear classification layer). We follow the VGG-16 architectures used in~\cite{frankle2020pruning, frankle2020linear} to remove the first two FC layers while keeping the last linear classification layer. This finally leads to a 14-layer architecture, but we still call it VGG-16 as it is modified from the original VGG-16 architectural design. Detailed architecture is shown in Table~\ref{arch:vgg}. We base our implementation on the GitHub repository\footnote{\url{https://github.com/kuangliu/pytorch-cifar/blob/master/models/vgg.py}}. 

\begin{table}[ht] %
	\caption{Detailed architecture of the VGG-16 architecture used in our experiments. We have a non-affine batchnnorm layer followed by a ReLU activation after each convolutional layer (omitted in the table). The shapes for convolution layers follow $(c_{in}, c_{out}, k, k)$.}
    \vspace{1mm}
    \centering
		 \scriptsize{
		 \resizebox{0.6\linewidth}{!}{
			\begin{tabular}{ccc}
				\toprule \textbf{Parameter}
				& Shape &  Layer hyper-parameter \bigstrut\\
				\midrule
				\textbf{layer1.conv1.weight} & $3 \times 64 \times 3 \times 3$ & stride:$1$;padding:$1$ \bigstrut\\
				\textbf{layer2.conv2.weight} & $64 \times 64 \times 3 \times 3$ & stride:$1$;padding:$1$  \bigstrut\\
				\textbf{pooling.max} & N/A & kernel size:$2$;stride:$2$  \bigstrut\\
				\textbf{layer3.conv3.weight} & $64\times 128 \times 3 \times 3$ & stride:$1$;padding:$1$ \bigstrut\\
				\textbf{layer4.conv4.weight} & $128\times 128 \times 3 \times 3$ & stride:$1$;padding:$1$ \bigstrut\\
                \textbf{pooling.max} & N/A & kernel size:$2$;stride:$2$  \bigstrut\\
				\textbf{layer5.conv5.weight} & $128 \times 256 \times 3 \times 3$ & stride:$1$;padding:$1$  \bigstrut\\
				\textbf{layer6.conv6.weight} & $256\times 256 \times 3 \times 3$ & stride:$1$;padding:$1$  \bigstrut\\
				\textbf{layer7.conv7.weight} & $256 \times 256 \times 3 \times 3$ & stride:$1$;padding:$1$  \bigstrut\\
                \textbf{pooling.max} & N/A & kernel size:$2$;stride:$2$  \bigstrut\\
				\textbf{layer8.conv9.weight} & $256 \times 512 \times 3 \times 3$ & stride:$1$;padding:$1$  \bigstrut\\
				\textbf{layer9.conv10.weight} & $512 \times 512 \times 3 \times 3$ & stride:$1$;padding:$1$  \bigstrut\\
				\textbf{layer10.conv11.weight} & $512 \times 512 \times 3 \times 3$ & stride:$1$;padding:$1$  \bigstrut\\
				\textbf{pooling.max} & N/A & kernel size:$2$;stride:$2$  \bigstrut\\
				\textbf{layer11.conv11.weight} & $512 \times 512 \times 3 \times 3$ & stride:$1$;padding:$1$  \bigstrut\\
				\textbf{layer12.conv12.weight} & $512 \times 512 \times 3 \times 3$ & stride:$1$;padding:$1$  \bigstrut\\
				\textbf{layer13.conv13.weight} & $512 \times 512 \times 3 \times 3$ & stride:$1$;padding:$1$  \bigstrut\\
				\textbf{pooling.max} & N/A & kernel size:$2$;stride:$2$  \bigstrut\\
				\textbf{pooling.avg} & N/A & kernel size:$1$;stride:$1$  \bigstrut\\
				\textbf{layer14.conv14.weight} & $512 \times 10 \times 1 \times 1$ & stride:$1$;padding:$1$  \bigstrut\\
				\bottomrule
			\end{tabular}}%
			}
	\label{arch:vgg}
\end{table}

\vspace{-3mm}
\paragraph{MobileNet-V2~\citep{sandler2018mobilenetv2}.} %
We base our implementation on the GitHub repository.\footnote{\url{https://github.com/kuangliu/pytorch-cifar/blob/master/models/mobilenetv2.py}}  Details of the architecture is shown in Table~\ref{arch:mobilenet}.

\begin{table}[ht]
\caption{The MobileNet-V2 structure that we use. Each layer consists of 3 total SubnetConv layer which correspond to the respective matrix. Inside the matrix is [kernal$\times$kernal, C$_{out}$]$\times$(number of times to repeat).}
\footnotesize
\vspace{1mm}
\centering
\begin{tabular}{cc} %
	\toprule \textbf{Layer Name}
	& MobileNet-V2 \\
	\midrule
	Conv1 & 3$\times$3, 32, stride 1, padding 1 \\
	\midrule
 	\multirow{3}{*}{Conv2} & \blockb{16}{32}{1}  \\
     &   \\
     &   \\
     \midrule
	\multirow{3}{*}{Conv3} & \blockb{24}{96}{1} \\
	&   \\
	&   \\
    \midrule
    \multirow{3}{*}{Conv4} & \blockb{24}{144}{1} \\
    &   \\
    &   \\
    \midrule
    \multirow{3}{*}{Conv5} & \blockb{32}{144}{1} \\
    &   \\
    &   \\
    \midrule
    \multirow{3}{*}{Conv6} & \blockb{32}{192}{2} \\
    &   \\
    &   \\
    \midrule
    \multirow{3}{*}{Conv7} & \blockb{64}{192}{1} \\
    &   \\
    &   \\
    \midrule
    \multirow{3}{*}{Conv8} & \blockb{64}{384}{3} \\
    &   \\
    &   \\
    \midrule
    \multirow{3}{*}{Conv9} & \blockb{96}{384}{1} \\
    &   \\
    &   \\
    \midrule
    \multirow{3}{*}{Conv10} & \blockb{96}{576}{2} \\
    &   \\
    &   \\
    \midrule
    \multirow{3}{*}{Conv11} & \blockb{160}{576}{1} \\
    &   \\
    &   \\
    \midrule
    \multirow{3}{*}{Conv12} & \blockb{160}{960}{2} \\
    &   \\
    &   \\
    \midrule
    \multirow{3}{*}{Conv13} & \blockb{320}{960}{1} \\
    &   \\
    &   \\
    \midrule
    \multirow{1}{*}{FC1} & 320$\times$1280 \\
    \multirow{1}{*}{FC2} & 1280$\times$10 \\
	\bottomrule
\end{tabular}
\label{arch:mobilenet}
\end{table}

\subsection{Hyper-Parameter Configuration}\label{appdx:config}

In this section, we will state the hyperparameter configuration for \algo{} and finetuning lottery tickets. For each dataset, model and different target sparsity, we tuned our hyperparameters for \algo{} by trying out different values of learning rate and L2 regularization weight $\lambda$. We also test different pruning periods of 5, 8, and 10 epochs. Finally, we also tried ADAM~\citep{kingma2014adam} and SGD. While SGD usually comes out on top, there were some settings where ADAM performed better.

\subsubsection{\algo{} Training}

We tested the CIFAR-10 dataset on the following architectures:
\begin{inparaenum}[i)]
\item ResNet-20
\item MobileNet-V2
\item VGG-16
\item WideResNet-28-2.
\end{inparaenum}
For TinyImageNet, we test on the architectures:
\begin{inparaenum}[i)]
\item ResNet-18
\item ResNet-50.
\end{inparaenum}
We tested the transfer learning on pretrained ImageNet model, where the target task is classification on Caltech-101 dataset with 101 classes. 
We first loaded the ResNet-50 model pretrained for ImageNet\footnote{\url{https://pytorch.org/vision/stable/models.html}} and changed the last layer to a single fully-connected network having size $2048 \times 101$. To match the performance of the pretrained model, we used Affine BatchNorm. The hyperparameter choices for each network, dataset and their corresponding sparsities are listed in Tables~(\ref{table:hyperparam_resnet20_cifar10}, \ref{table:hyperparam_mobilenet_cifar10},  \ref{table:hyperparam_vgg16_cifar10}, \ref{table:hyperparam_wideresnet_cifar10}, \ref{table:hyperparam_resnet18_tinyimagenet}, \ref{table:hyperparam_resnet50_tinyimagenet}, \ref{table:hyperparam_resnet50_caltech})

\begin{table}[ht!]
\caption{Hyper Parameters used for different sparsities for \algo{} on ResNet-20 on CIFAR-10.}
\footnotesize
\vspace{1mm}
\centering
\begin{tabular}{cccccc} %
\toprule
Network/Dataset & Sparsity & Pruning Period & Optimizer & LR & Lambda \\
\midrule 
\multirow{5}{*}{\shortstack{ResNet-20\\CIFAR-10}} &  50\% &  8 &  SGD &  0.05 &  $10^{-8}$ \\
& 13.74\% & 5 & SGD & 0.1 & $10^{-5}$ \\ 
&  3.73\% &  5 &  SGD &  0.1 &  $3\times10^{-5}$  \\
& 1.44\% & 5 & SGD & 0.1 & $10^{-4}$ \\
&  0.59\% &  5 &  SGD &  0.1 &  $10^{-4}$ \\
\bottomrule
\end{tabular}
\label{table:hyperparam_resnet20_cifar10}
\end{table}

\begin{table}[ht!]
\caption{Hyper Parameters used for different sparsities for \algo{} on MobileNet-V2 on CIFAR-10.}
\footnotesize
\vspace{1mm}
\centering
\begin{tabular}{cccccc}
\toprule
Network/Dataset & Sparsity & Pruning Period & Optimizer & LR & Lambda \\
\midrule 
\multirow{4}{*}{\shortstack{MobileNet-V2\\CIFAR-10}} & 50\% & 5 & SGD & 0.05 & 0 \\
&  20\% &  5 &  SGD &  0.1 &  $3\times10^{-6}$ \\
& 5\% &  5 & SGD & 0.1 & $7\times10^{-6}$ \\
&  1.44\% & 5 &  SGD &  0.1 &  $10^{-4}$ \\
\bottomrule
\end{tabular}
\label{table:hyperparam_mobilenet_cifar10}
\end{table}

\begin{table}[ht!]
\caption{Hyper Parameters used for different sparsities for \algo{} on VGG-16 on CIFAR-10. }
\footnotesize
\vspace{1mm}
\centering
\begin{tabular}{cccccc}
\toprule
Network/Dataset & Sparsity & Pruning Period & Optimizer & LR & Lambda \\
\midrule 
\multirow{5}{*}{\shortstack{VGG-16\\CIFAR-10}} & 50\%  & 5 & SGD & 0.01 & 0 \\ 
&  5\%  &  5 &  SGD &  0.01 &  $10^{-6}$ \\
& 2.5\% & 5 & SGD & 0.01 & $10^{-6}$ \\
&  1.4\% &  5 &  SGD &  0.01 &  $10^{-6}$ \\
& 0.5\% & 5 & SGD & 0.01 & $10^{-6}$ \\
\bottomrule
\end{tabular}
\label{table:hyperparam_vgg16_cifar10}
\end{table}

\begin{table}[ht!]
\caption{Hyper Parameters used for different sparsities for \algo{} on WideResNet-28-2 on CIFAR-10.}
\footnotesize
\vspace{1mm}
\centering
\begin{tabular}{cccccc}
\toprule
Network/Dataset & Sparsity & Pruning Period & Optimizer & LR & Lambda \\
\midrule 
\multirow{5}{*}{\shortstack{WideResNet-28-2\\CIFAR-10}} &  50\% &  5 &  SGD &  0.1 &  0 \\
& 20\%& 10 & SGD & 0.1 & $10^{-5}$ \\
&  5\%&  10 &  SGD &  0.1 &  $10^{-5}$ \\
& 1.44\%& 10 & SGD & 0.1 & $10^{-5}$ \\
&  0.5\% &  10 &  SGD &  0.1 &  $10^{-5}$ \\
\bottomrule
\end{tabular}
\label{table:hyperparam_wideresnet_cifar10}
\end{table}

\begin{table}[ht!]
\caption{Hyper Parameters used for different sparsities for \algo{} on ResNet-18 on TinyImageNet.}
\footnotesize
\vspace{1mm}
\centering
\begin{tabular}{cccccc}
\toprule
Network/Dataset & Sparsity & Pruning Period & Optimizer & LR & Lambda \\
\midrule 
\multirow{4}{*}{\shortstack{ResNet-18\\TinyImageNet}} &  50\% &  10 &  SGD &  0.1 &  $0$ \\
& 5\% & 5 & SGD & 0.001 & $8\times 10^{-6}$ \\
&  1.4\% &  5 &  SGD &   0.001 &  $5 \times 10^{-6}$ \\
& 0.5\% & 5 & SGD & 0.001 & $10^{-5}$ \\
\bottomrule
\end{tabular}
\label{table:hyperparam_resnet18_tinyimagenet}
\end{table}

\begin{table}[ht!]
\caption{Hyper Parameters used for different sparsities for \algo{} on ResNet-50 on TinyImageNet.}
\footnotesize
\vspace{1mm}
\centering
\begin{tabular}{cccccc}
\toprule
Network/Dataset & Sparsity & Pruning Period & Optimizer & LR & Lambda \\
\midrule 
\multirow{4}{*}{\shortstack{ResNet-50\\TinyImageNet}} &  50\% &  5 &  ADAM &  0.01 &  0 \\
& 5\% & 5 & ADAM & 0.01 & $10^{-6}$ \\
&  1.4\% &  5 &  ADAM &   0.01 &  $10^{-6}$ \\
& 0.5\% & 5 & ADAM & 0.01 & $10^{-6}$ \\
\bottomrule
\end{tabular}
\label{table:hyperparam_resnet50_tinyimagenet}
\end{table}

\begin{table}[ht!]
\caption{Hyper Parameters used for different sparsities for \algo{} on ResNet-50 on Caltech101.}
\footnotesize
\vspace{1mm}
\centering
\begin{tabular}{cccccc}
\toprule
Network/Dataset & Sparsity & Pruning Period & Optimizer & LR & Lambda \\
\midrule 
\multirow{4}{*}{\shortstack{ResNet-50\\Caltech101}} &  50\% &  5 &  ADAM &  0.01 &  0 \\
& 5\% & 5 & ADAM & 0.01 & $10^{-6}$ \\
&  1.4\% &  5 &  ADAM &   0.01 &  $10^{-6}$ \\
& 0.5\% & 5 & ADAM & 0.01 & $10^{-6}$ \\
\bottomrule
\end{tabular}
\label{table:hyperparam_resnet50_caltech}
\end{table}

\subsubsection{Finetuning the Rare Gems}
The details of the hyperparameter we used in finetuning the rare gems we find is shown in Table~\ref{tab:hyperparameter}.

\begin{table}[ht!]
    \caption{Hyperparameters used for finetuning. We use the same number of epochs for \algo{} and finetuning.}
    \label{tab:hyperparameter}
    \centering
    \resizebox{\columnwidth}{!}{%
    \begin{tabular}{ccccccc}
    \toprule
    Model & Dataset & Sparsity & Epochs & Batch Size & LR & Multi-Step Milestone \\ \midrule
    
    \multirow{2}{*}{ResNet-20} & CIFAR-10 &  50\% &  150 &  128 & 0.1 & [80, 120] \bigstrut\\
    
    & CIFAR-10 & others & 150 & 128 & 0.01 & [80, 120] \bigstrut\\
    
    \hline
    \multirow{1}{*}{MobileNet-V2} & CIFAR-10 & all &  300 &  128 &  0.1 & [150, 250] \bigstrut\\
    
    \hline
    \multirow{1}{*}{VGG-16} &  CIFAR-10 & all & 200 & 128 & 0.05 & [100, 150] \bigstrut\\
    
    \hline
    \multirow{2}{*}{WideResNet-28-2} &  CIFAR-10 &  50\% &  150 & 128 & 0.1 &  [80, 120] \bigstrut\\
    
    & CIFAR-10 & others & 150 & 128 & 0.01 & [80, 120] \bigstrut\\
    
    \hline
    \multirow{1}{*}{ResNet-18} &  TinyImageNet &  all &  200 &  256 &  0.1 &  [100, 150] \bigstrut\\
    
    \hline
    \multirow{1}{*}{ResNet-50} & TinyImageNet & all & 150 & 256 & 0.1 & [80, 120] \bigstrut\\
    
    \hline
    \multirow{1}{*}{ResNet-50 (Pretrained)} & Caltech-101 &  all &  50 &  16 &  0.0001 & [20, 40]  \bigstrut\\ \bottomrule
    \end{tabular}
    }
\end{table}

\begin{figure}[ht] 
\centering
\includegraphics[width=0.99\textwidth]{figures/Legend_with_Renda_Latest_withborder.pdf}

\subfloat[\centering{CIFAR-10, ResNet-18}]{\includegraphics[height=25mm]{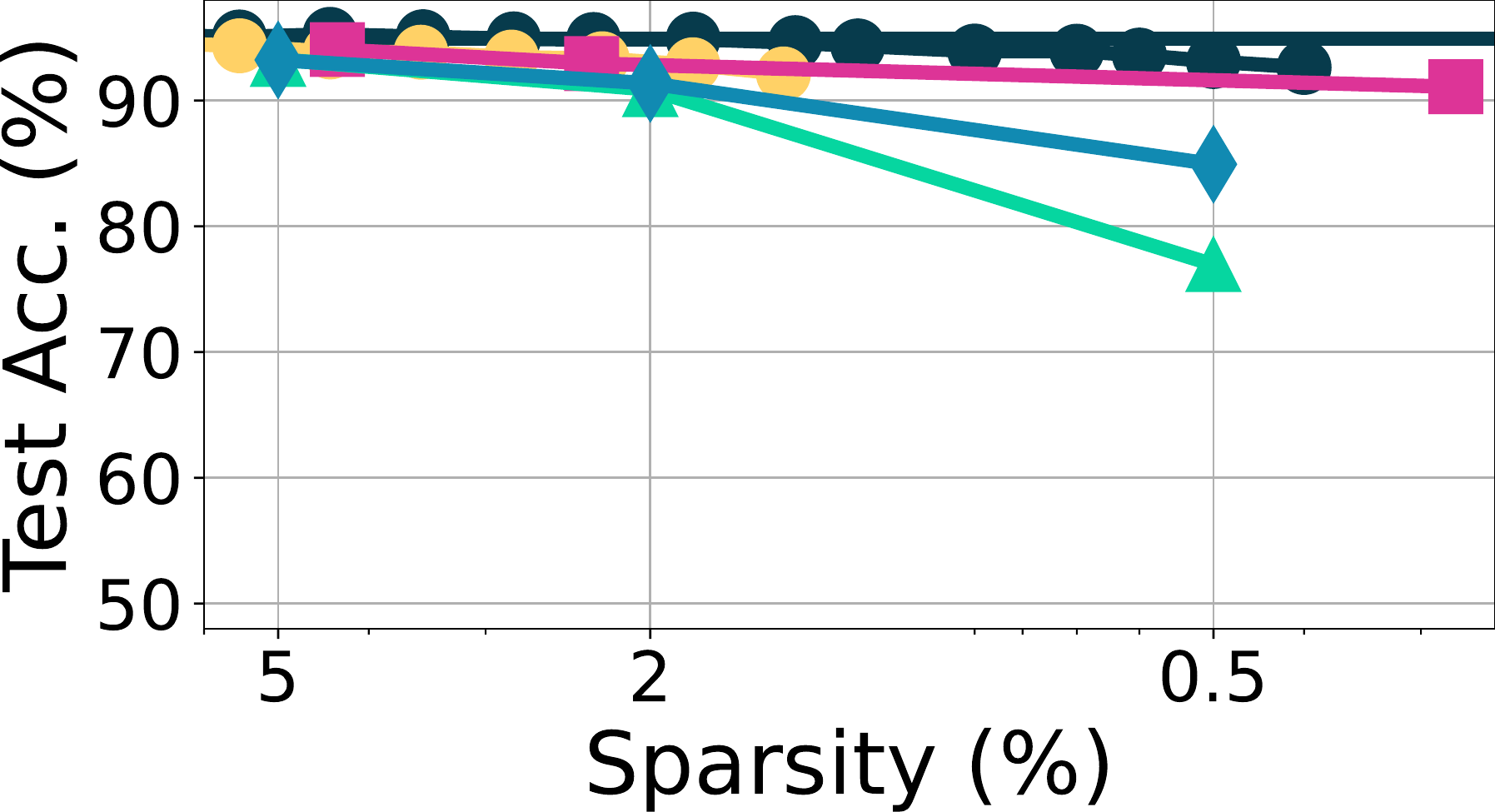}}
\hspace{0.5cm}
\subfloat[\centering{TinyImageNet, MobileNet-V2}]{\includegraphics[height=25mm]{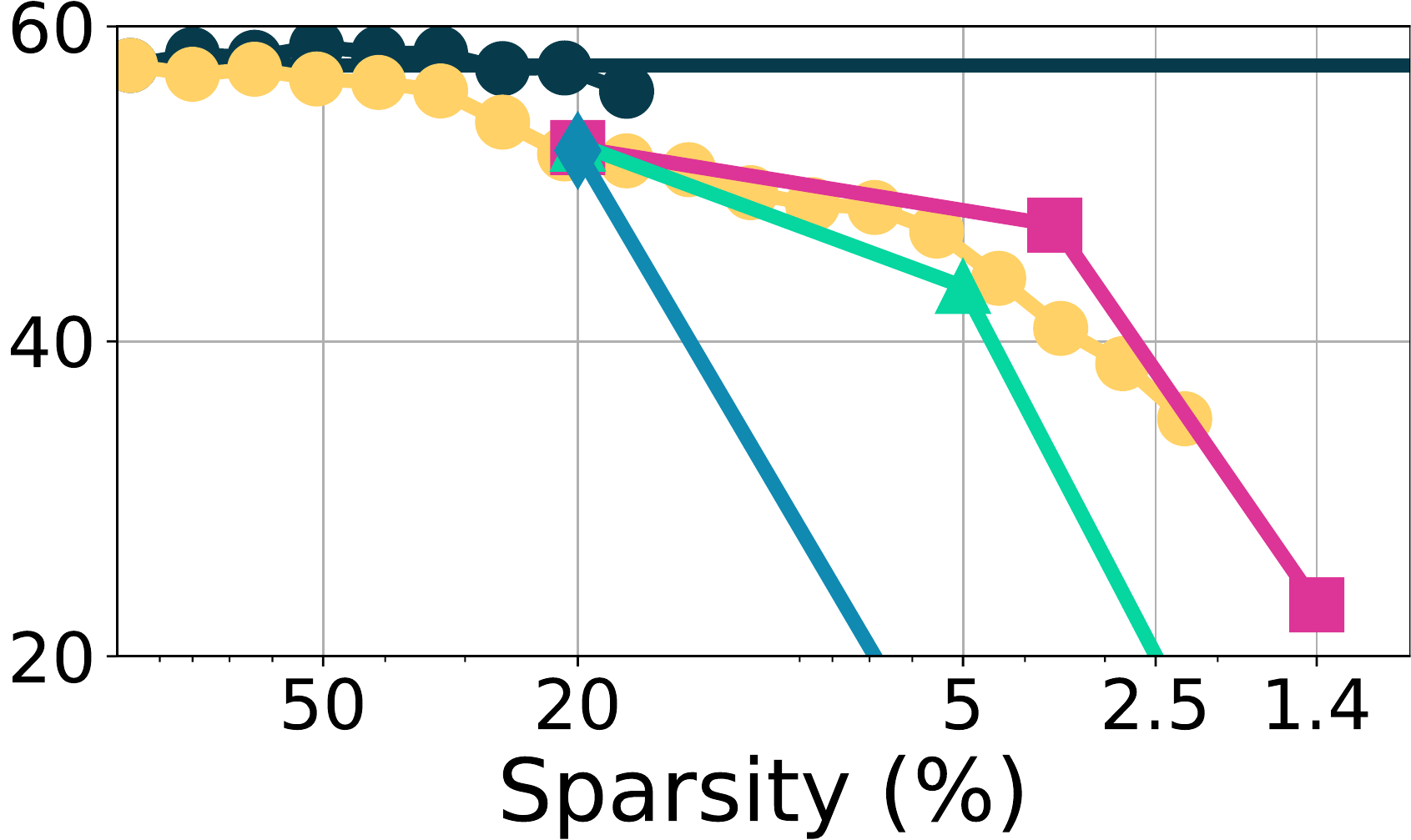}}\\

\vspace{-0.5cm}
\subfloat[]{\includegraphics[height=25mm]{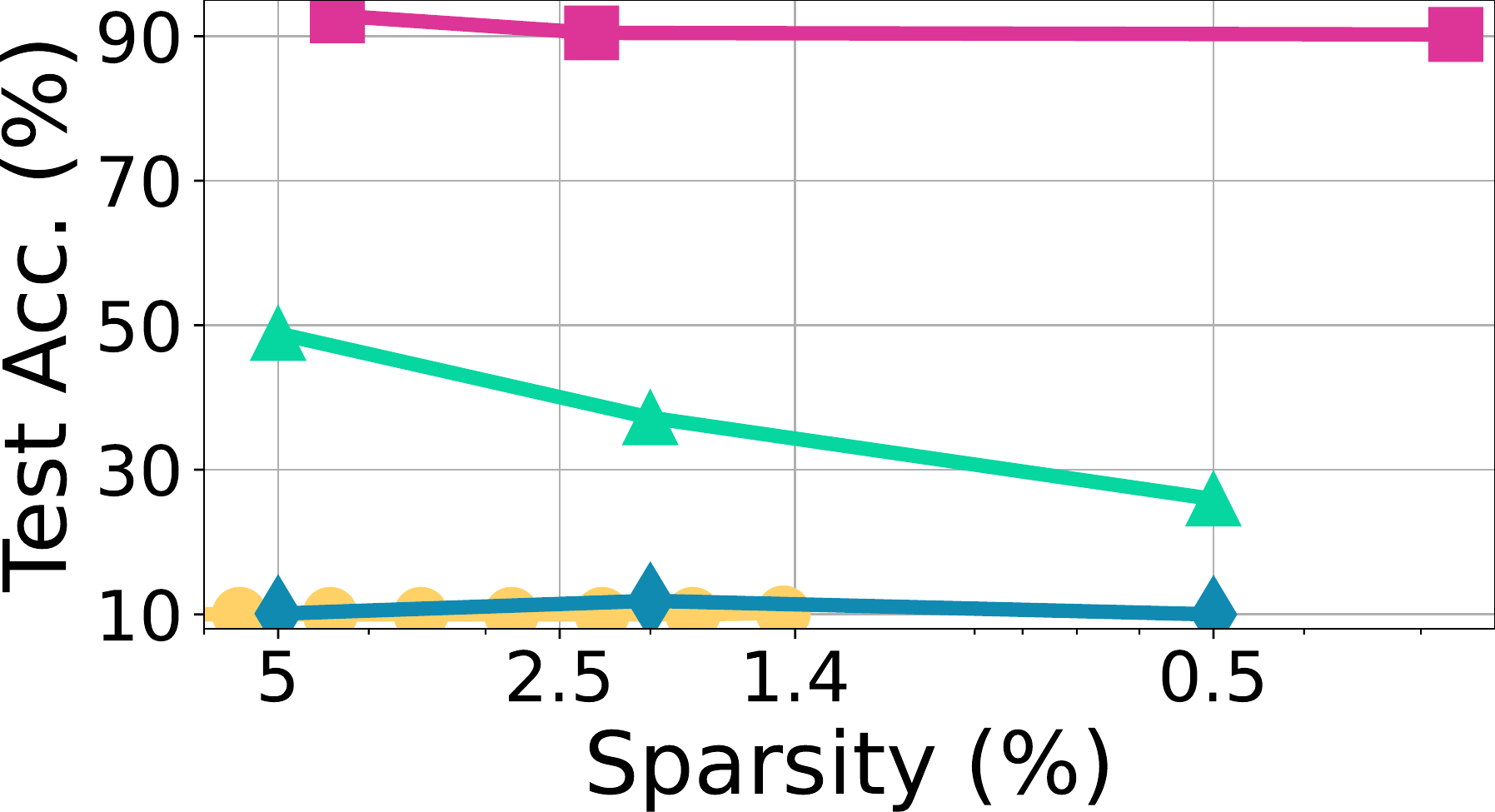}}
\hspace{0.5cm}
\subfloat[]{\includegraphics[height=25mm]{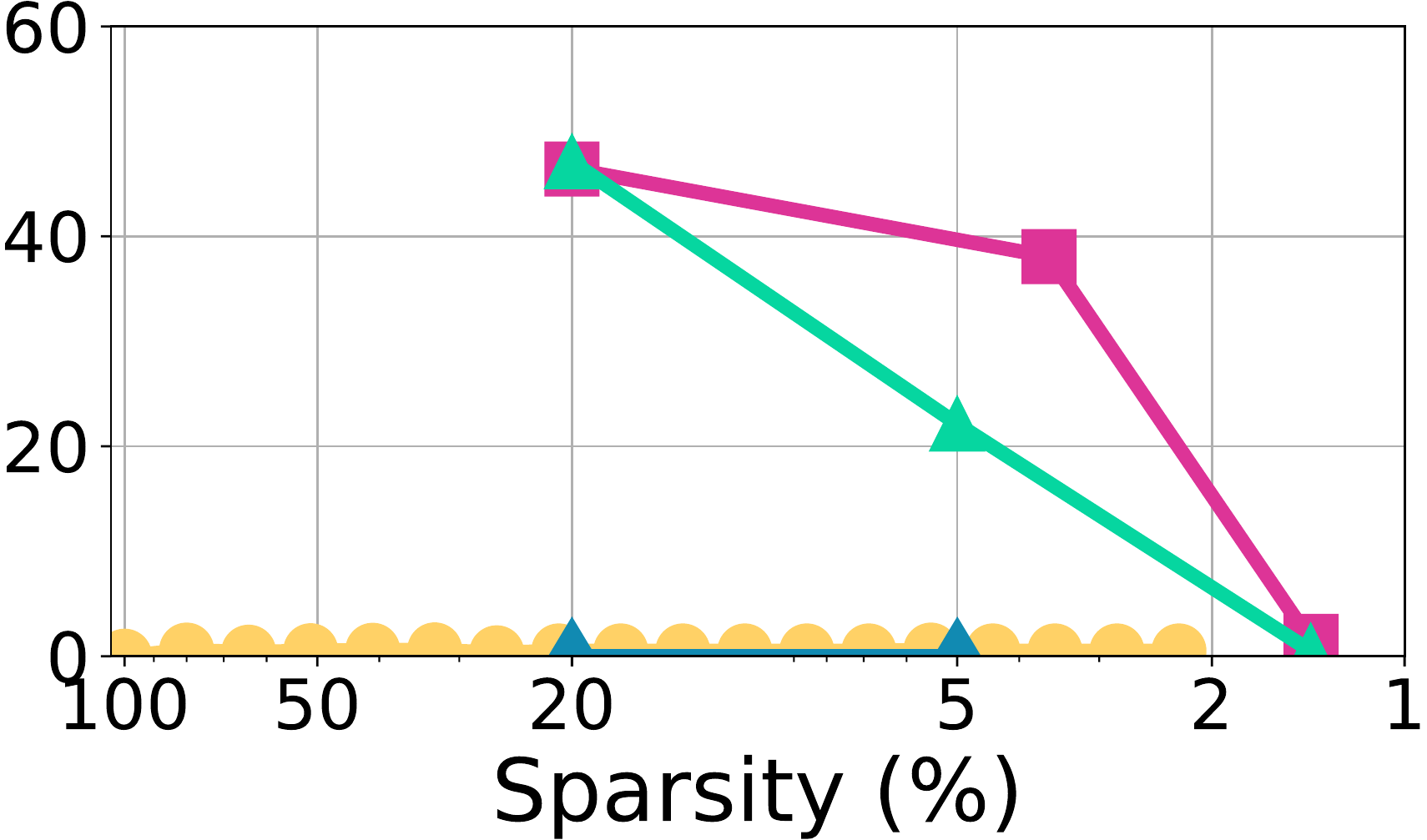}}\\

\vspace{-0.5cm}
\subfloat[]{\includegraphics[height=25mm]{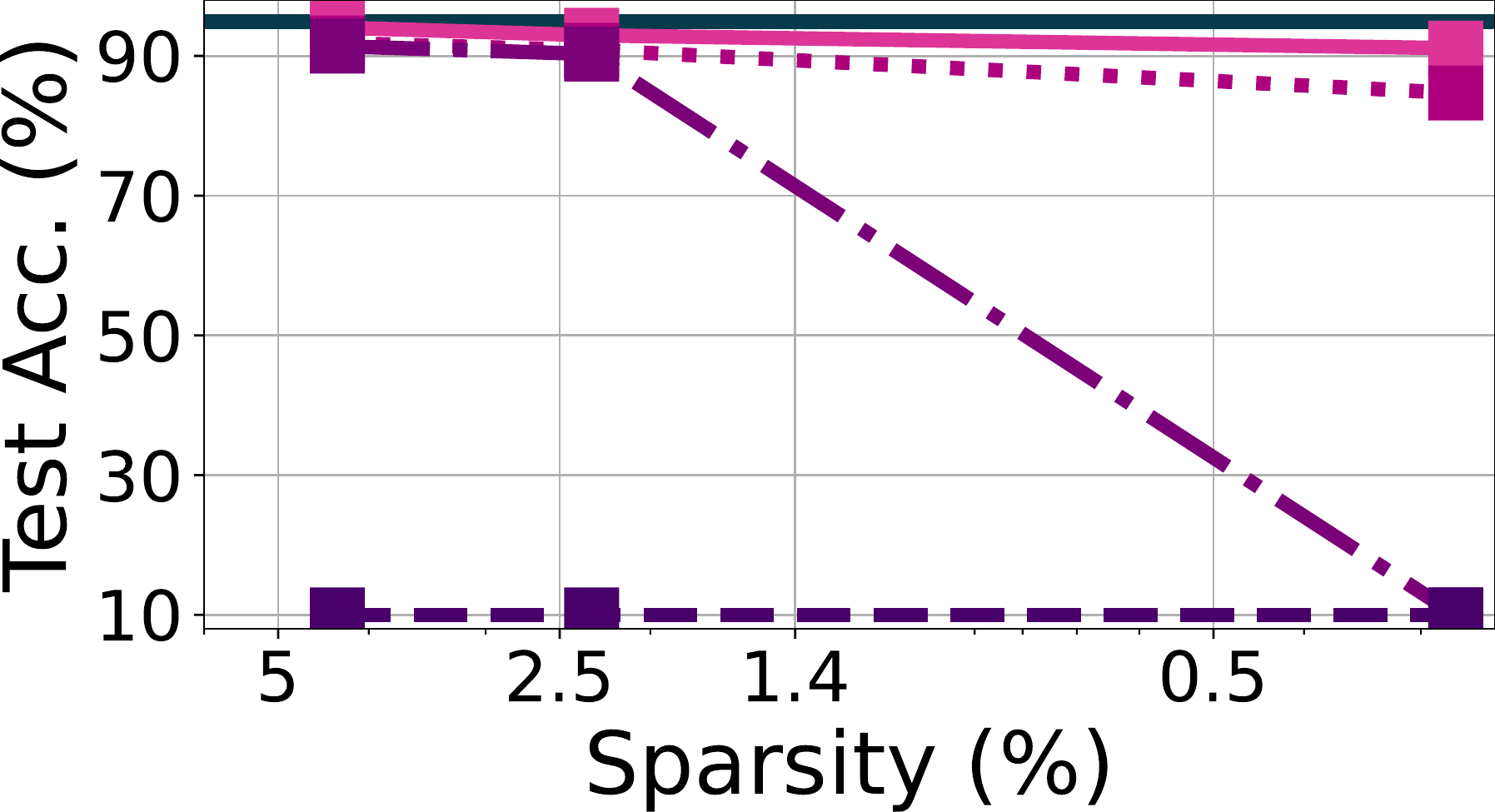}}
\hspace{0.5cm}
\subfloat[]{\includegraphics[height=25mm]{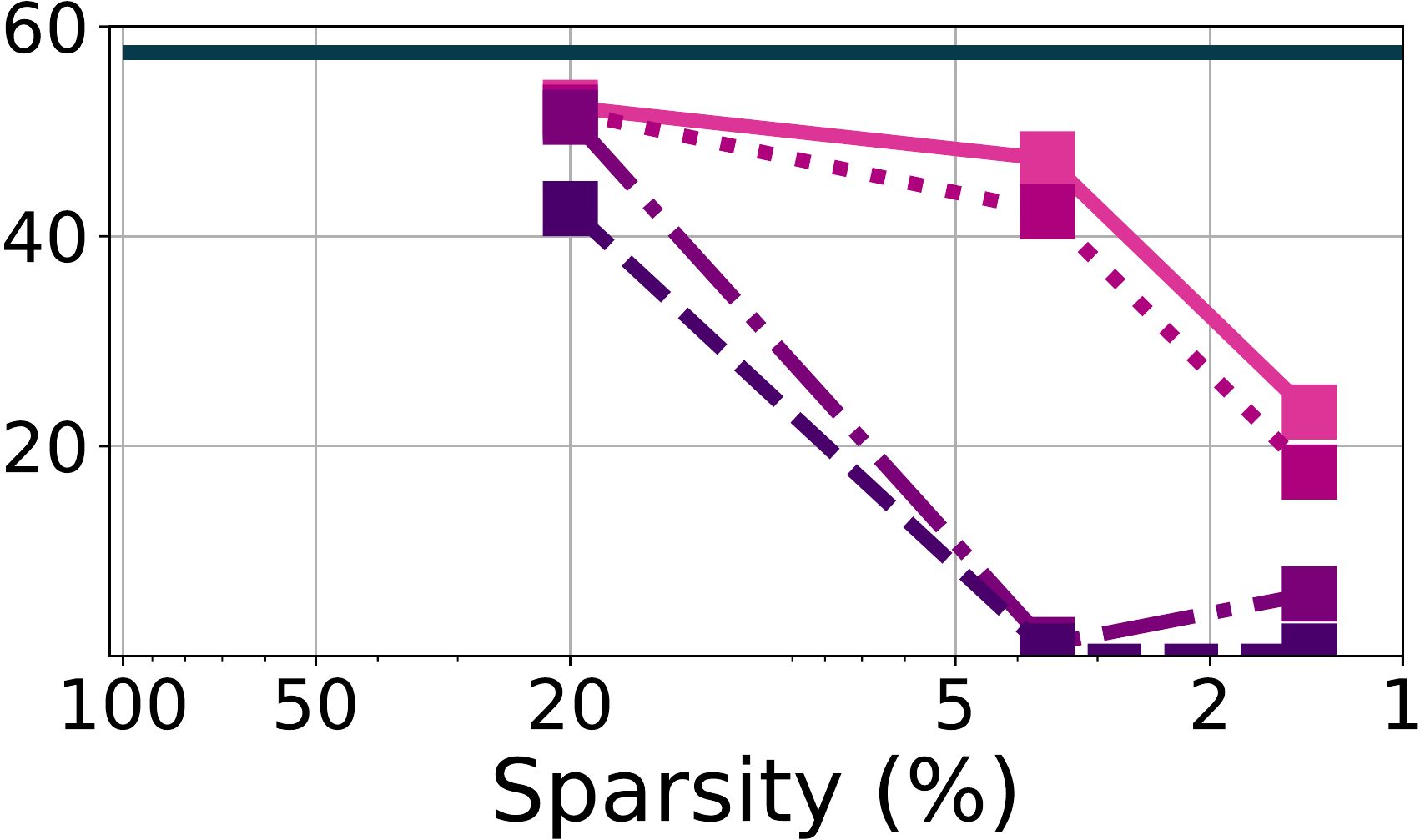}}

\caption{
Additional experimental results on comparing different pruning algorithms.
Top: post-finetune accuracy, Middle: pre-finetune accuracy, Bottom: sanity check methods suggested in~\citet{frankle2020pruning} applied on \algo{}. Similar to earlier experiments, we find that \algo{} outperforms IMP (with warmup) in the sparse regime in terms of both post finetune as well as pre-finetune accuracy. We also note that \algo{} passes all of the sanity checks.
}
\label{fig:additional_exp}
\end{figure}

\section{Additional Experiments}

We repeated the comparison of \algo{} with the baselines on MobileNet-V2 on TinyImagenet as well as ResNet-18 on CIFAR-10. We show the results in Fig.~\ref{fig:additional_exp}. Similar to our earlier experiments, we have the following observations. Note that \algo{} outperforms IMP (with warmup) in the sparse regime. Also, as is expected, \algo{} has non-trivial accuracy before finetune, which is higher than both EP and significantly higher than IMP. \algo{} shows a significant deterioration in performance when subjected to the sanity checks suggested in \citet{frankle2020pruning}. Therefore, \algo{} is considered to pass the sanity checks.
Finally \algo{} (\emph{at initialization}) nearly achieves the performance of \textbf{Renda et al.} (Learning rate rewinding) in the sparse regime. This is particularly impressive given that \textbf{Renda et al.} is a pruning-after-training method \ie it prunes and trains iteratively, never finding a subnetwork at or even near initiailization.

\paragraph{How far can we take random pruning?}
\label{par:sr_versions}

Recall that subnetworks identified by \algo{} outperform random subnetworks found by sampling based on smart ratio (SR)~\citep{su2020sanity}. However, SR performs surprisingly well given that it is still random pruning. Therefore, we tried improving SR to see how far we could take random pruning. For simplicity, we call these different versions of SR starting with \textbf{v1} which is the original algorithm suggested by \citet{su2020sanity}.
Table~\ref{table:SRvsGM} compares the post-finetune accuracy of subnetworks found by training randomly initialized subnetworks at initialization, each using ``smart ratios'' found by increasingly sophisticated means. Given $L$-layer network, each version of SR uses its own sparsity pattern $\vp = [p_1, \cdots, p_L]$ to generate random subnetwork, where each layer randomly picks $0 < p_l \leq 1$ fraction of weights to use. The sparsity patterns are identified as follows:

\begin{itemize}
    \item \emph{SR-v1}: vanilla SR suggested in~\citep{su2020sanity}
    \item \emph{SR-v2}: set $p_l = p_l^{\op{SR}}$ for $2 \leq l \leq L-1$ and $p_l = p_l^{\op{GM}}$ for $l \in \{1, L\}$
    \item \emph{SR-v3}: set $p_l = p_l^{\op{SR}}$ for $2 \leq l \leq L-1$ and $p_l = 1$ for $l \in \{1, L\}$
    \item \emph{SR-v4}: start with $\vp^{\op{IMP}}$ and search sparsity patterns in a small ball around it.
    \item \emph{SR-v5}: start with the sparsity pattern of v2, and tune $p_l$ using Eq~(\ref{eqn:tune_p})
    \item \emph{SR-v6}: start with the sparsity pattern of v4, and tune $p_l$ using Eq~(\ref{eqn:tune_p})
\end{itemize}

The vanilla SR proposed by \citet{su2020sanity} is denoted by \emph{SR-v1}. Motivated by the fact that (i) \algo{} outperforms SR and (ii) \algo{} and SR primarily differ in the layerwise sparsity $p_l$ for the first and the last layers (refer Figure~\ref{fig:cifar11}), we construct \emph{SR-v2}. It uses the layerwise sparsity of \emph{SR-v1} for intermediate layers ($2 \leq l \leq L$) and layerwise sparsity of \algo{} for $l \in \{1, L\}$. As a simple variant of \emph{SR-v2}, we also considered using full dense layer ($p_l = 1$) for the first and the last layer, which is denoted by \emph{SR-v3}. This was because we observed that the first and last layers found by \algo{} were relatively dense compared to \emph{SR-v1}.
\emph{SR-v4} searches a few points around the sparsity pattern of IMP and chooses the best option. We chose only a few options based on intuition to make the search computationally tractable. 

Finally, we tried to write finding the optimal random pruning method by writing it as an optimization problem in $p_l$ as follows:

\begin{align}\label{eqn:tune_p}
\min_{p_l} \sum_{(\vx, y) \in S} \ell ( f(\vw_l \odot \op{Bern}(p_l); \vx), y)
\end{align}

Intuitively, this is equivalent to choosing $p_l$ such that the loss of the random subnetwork generated by sampling the mask of layer $l$ with probability $p_l$ is minimized.

In order to make the problem more tractable, we set the output of previous versions as the initial value of $p_l$. Choosing $p_l^{(0)}$ as the $p_l^{\text{SR-v2}}$ and then applying SGD on Eq~\ref{eqn:tune_p} gives us \emph{SR-v5}. Repeating this with $p_l^{(0)} = p_l^{\text{SR-v4}}$ results in \emph{SR-v6}.

The results of Table~\ref{table:SRvsGM} show that different strategies in random pruning can improve the performance of random pruning, but \algo{} still has an 8\% accuracy gap with the best random network we found for ResNet-20, CIFAR-10 classification, when the target sparsity is 1.44\%.

\begin{table}[t]
    \caption{
	Performance comparison of \algo{} and variants of Smart Ratio (SR), for ResNet-20 trained for CIFAR-10 classification task, for target sparsity $1.44\%$.
	We denoted the vanilla SR in~\citep{su2020sanity} as \emph{SR-v1} and tested five additional variants, from \emph{SR-v2} to \emph{SR-v6}. The detailed description of the variants are given in Paragraph~\ref{par:sr_versions}. 
	}
    \centering
    \vspace{1mm}
    \scriptsize
    \setlength{\tabcolsep}{8pt} %
    \renewcommand{\arraystretch}{0.5}
		 {
			\begin{tabular}{c|c|c|c|c|c|c|c}
				\toprule \textbf{Schemes}
				& Gem-Miner &  SR (v1)~\citep{su2020sanity} & SR (v2) & SR (v3) & SR (v4) & SR (v5) & SR (v6) 
				\bigstrut\\
				\midrule
				
				\textbf{Sparsity (\%)} 
				& 1.44
				& 1.44
				& 1.47
				& 1.75
				& 1.44
				& 1.53
				& 1.47
				\bigstrut\\
				\textbf{Accuracy (\%)} 
				& \textbf{77.89}
				& 65.61
				& 68.59
				& 69.78
				& 69.92
				& 69.01
				& 69.08
				\bigstrut\\
				\bottomrule
			\end{tabular}}
	\label{table:SRvsGM}
\end{table}

\paragraph{Relationship between pre-finetune accuracy and post-finetune accuracy.}

Recall that rare gems need to have not only high \emph{post-finetune} accuracy but also non-trivial \emph{pre-finetune} accuracy.

\begin{wrapfigure}{hr}{0.5\columnwidth}
	\centering
	\includegraphics[width=0.45\textwidth]{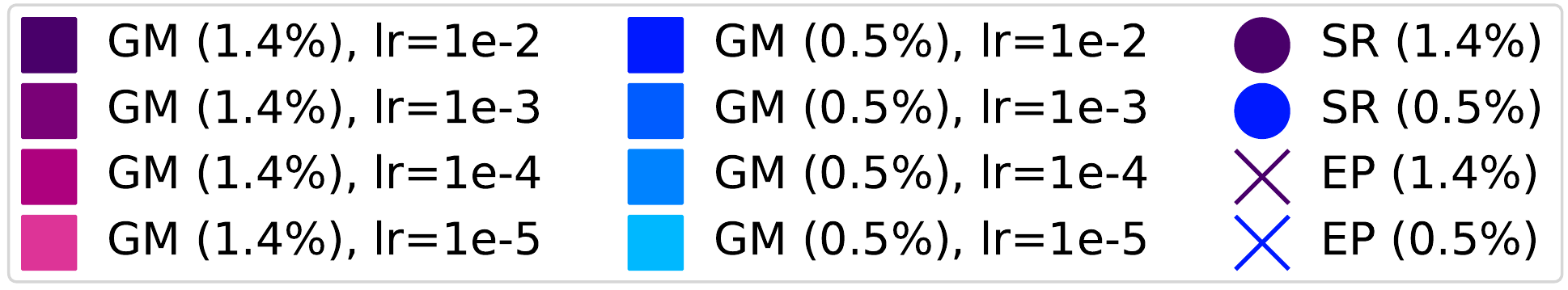}
	\includegraphics[width=0.45\textwidth]{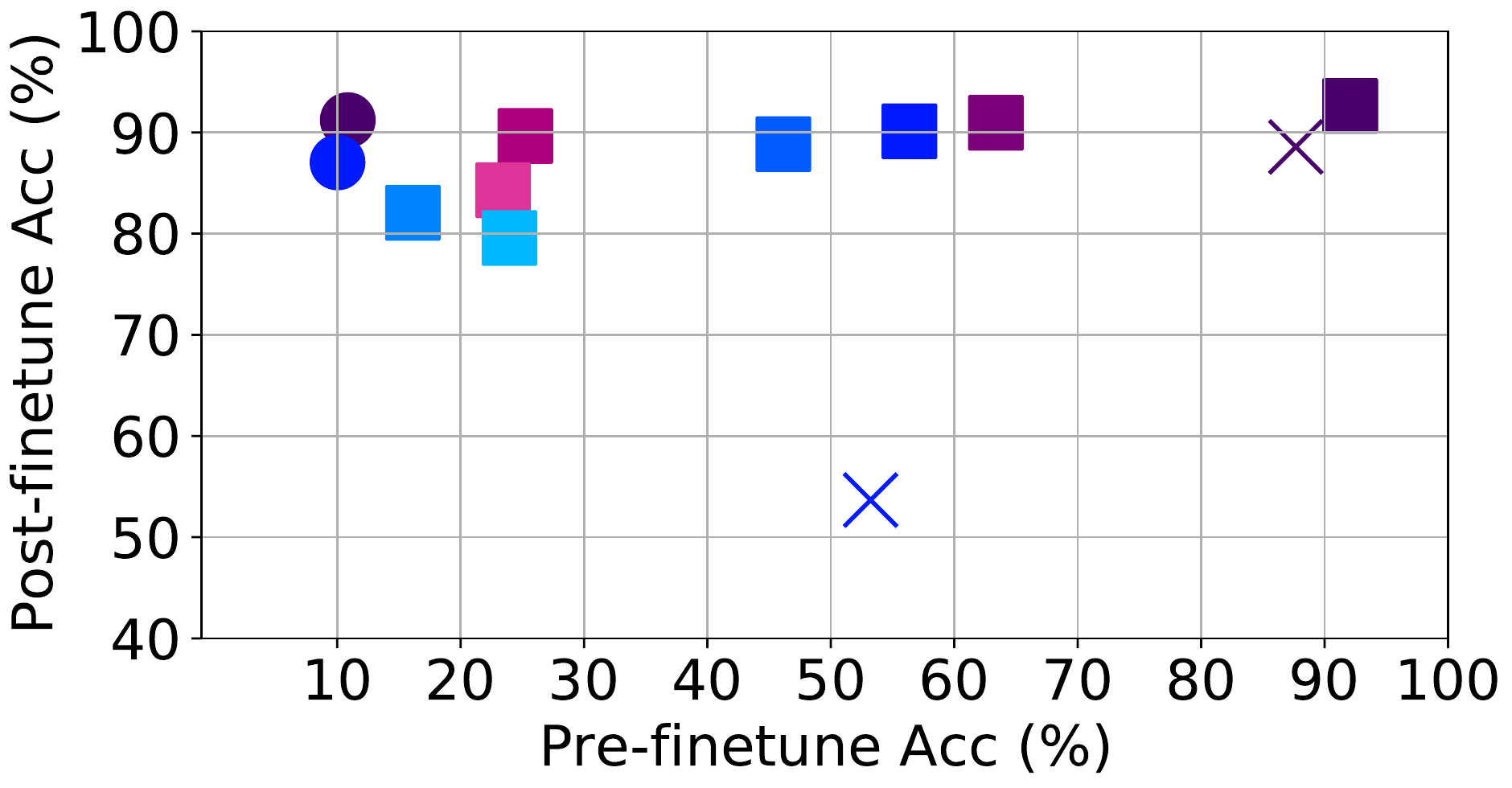}
    \caption{Relationship between pre-finetune accuracy and post-finetune accuracy, on CIFAR-10, VGG-16, for \algo{} (GM), edge-popup (EP), and smart ratio (SR). For both 1.4\% and 0.5\% sparsity, \algo{} typically shows higher post-finetune accuracy when it has higher pre-finetune acuracy. As noted by \citet{ramanujan2020s}, subnetworks found by EP are not finetunable.}
\label{fig:pre_vs_post}
\vspace{-0.2cm}
\end{wrapfigure}

Since the latter is a lower bound on \emph{post-finetune} accuracy, we design \algo{} to just maximize the accuracy at initialization.
However, it is not clear that this actually maximizes post-finetune accuracy. In fact, the performance of EP and IMP clearly show that it is neither a necessary nor sufficient condition. Fig.~\ref{fig:pre_vs_post} shows both pre and post finetune accuracies for \algo{} (GM), edge-popup (EP), and smart ratio (SR), at both 1.4\% and 0.5\% sparsity, for CIFAR-10 classification using VGG-16. For \algo{}, we show the results of using four different learning rates $\eta= 0.01, 0.001, 0.0001, 0.00001$ for 200 epochs using lighter colors to indicate lower learning rates. It turns out that $\eta=0.01$ has the best pre and post-finetune accuracies for both sparsities.
Moreover, it shows that there is some correlation between pre-finetune and post-finetune accuracy \ie subnetworks that have higher pre-finetune accuracies typically have higher accuracy after finetuning as well. However, comparing \algo{}'s results for 0.5\% sparsity with $\eta=0.0001$ and $\eta=0.00001$ shows that this pattern does not always hold. With $\eta=0.00001$, \algo{} achieves a higher pre-finetune accuracy but ends up with lower accuracy after finetuning. Therefore, we conclude that while pre-finetune accuracy is a reasonable proxy for accuracy after finetuning, it does not guarantee it in any way. 
Note that both points for EP have ``post-finetune accuracy'' $\simeq$ ``pre-finetune accuracy'', which confirms the observation by \citet{ramanujan2020s} that subnetworks found by EP are not lottery tickets \ie they are not conducive to further training.

\paragraph{Discussions on the need of warmup for IMP.}
\vspace{-2mm}

For completeness, we also tried different variants of IMP. We refer to it as ``cold'' IMP when the weights are rewound to initialization, while ``warm'' IMP rewinds to some early iteration, \ie after training for a few epochs. Further, we classify it depending on the number of epochs per magnitude pruning. We say IMP is ``short'' if it only trains for a few epochs (\eg 8 epochs for ResNet-20 on CIFAR-10) before pruning, and ``long'' if it takes a considerably larger number of epochs before pruning (\eg 160 epochs for ResNet-20 on CIFAR-10). Regardless of ``long'' or ``short'', the number of epochs to finetune the pruned model are the same.

\begin{wrapfigure}{hr}{0.5\columnwidth}
	\centering
	\includegraphics[width=0.49\textwidth]{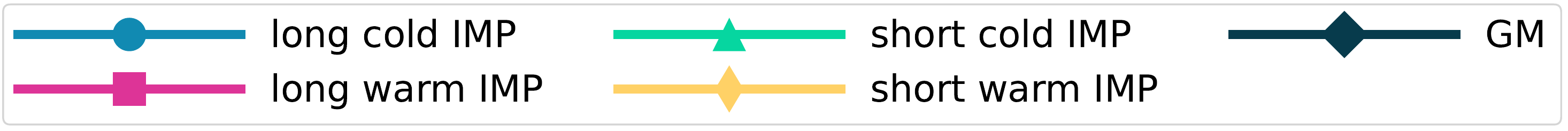}
    \subfloat[\centering{Before Finetuning}]{
    \includegraphics[width=0.24\textwidth]{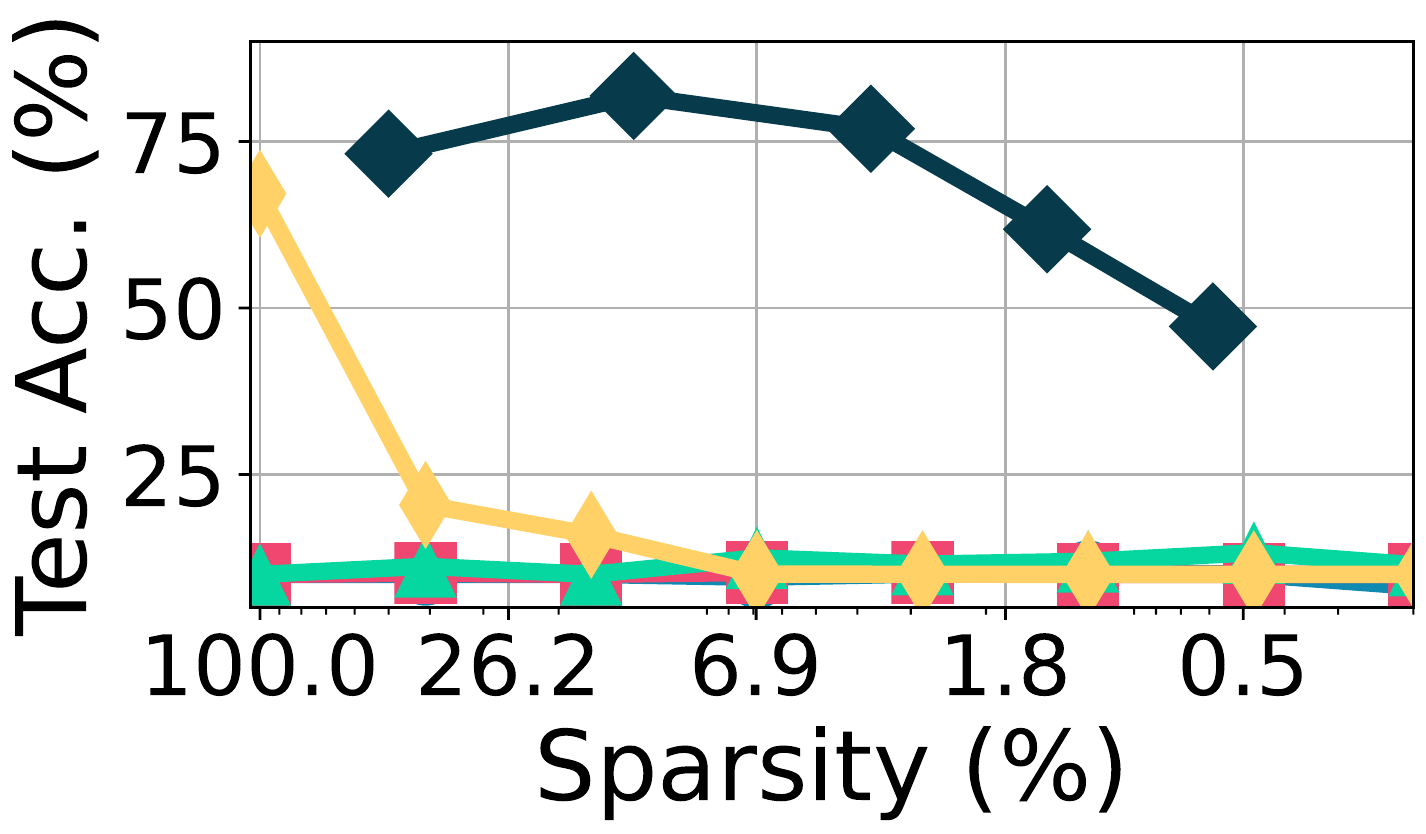}}
    \subfloat[\centering{After Finetuning}]{
    \includegraphics[width=0.24\textwidth]{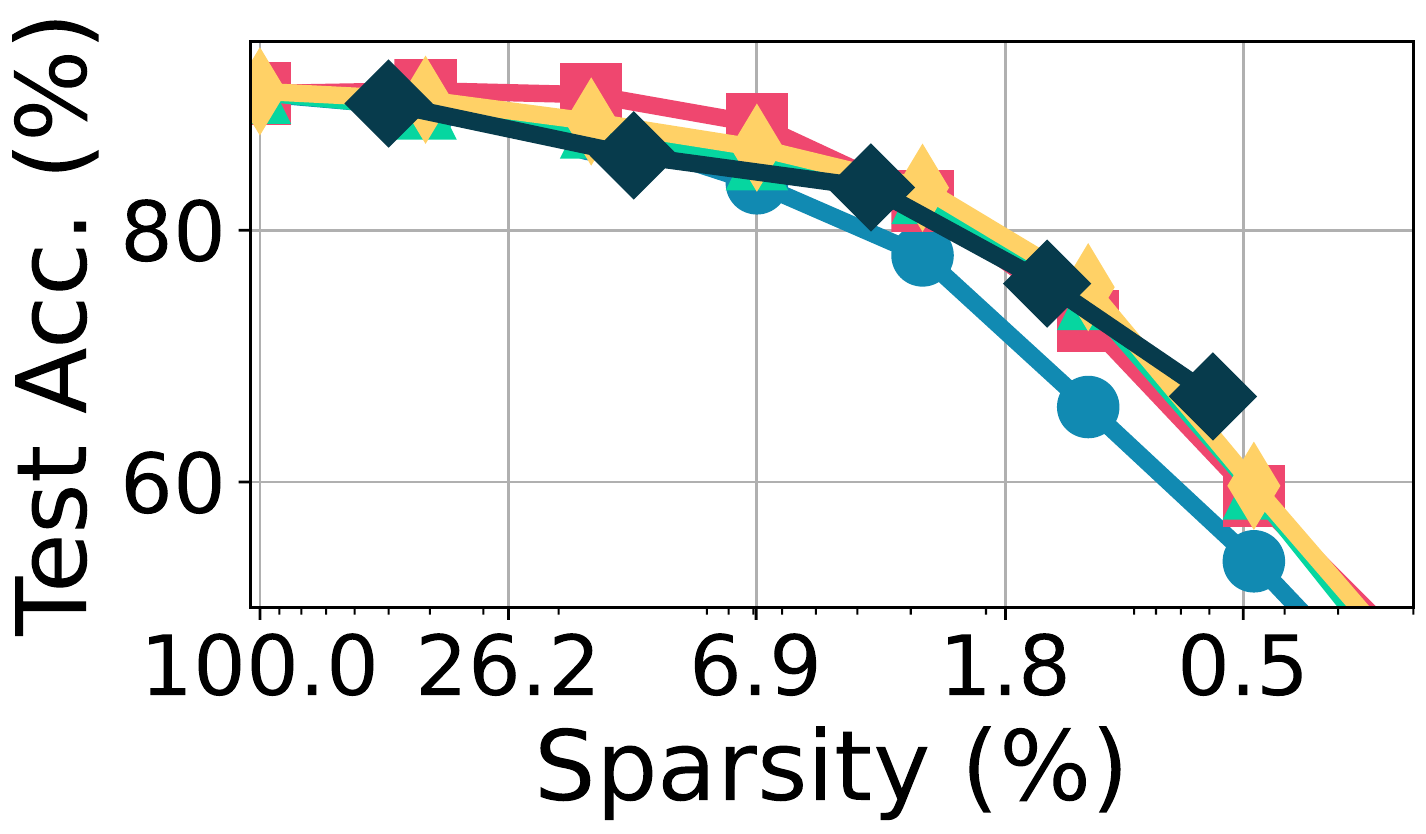}}
    \caption{Test accuracy before (\emph{left}) and after (\emph{right}) finetuning for ResNet-20 on CIFAR-10. The ``short'' version of IMP achieves the same accuracy-sparsity tradeoff as \emph{long-warm} IMP in the sparse regime. However, comparing the before finetuning accuracy (\emph{left}) shows that \algo{} is capable of finding rare gems at initialization, whereas \emph{short-cold} IMP can only find lottery tickets.}
\label{fig:different_IMP}
\vspace{-0.7cm}
\end{wrapfigure}

With these informal definitions, we can categorize IMP into four different versions:
\begin{inparaenum}[a)]
\item \emph{short-cold} IMP,
\item \emph{short-warm} IMP,
\item \emph{long-cold} IMP, and
\item \emph{long-warm} IMP.
\end{inparaenum}

In the literature, only \emph{long} variants have been studied thoroughly. \citet{frankle2020linear} noted that for large networks and difficult tasks, long-cold IMP fails to find lottery tickets, which is why they introduce long-warm IMP. 

Somewhat surprisingly, as shown in Fig. \ref{fig:different_IMP}, 
we find that short versions of IMP can achieve the same or an even better accuracy-sparsity tradeoff especially in the sparse regime.
In particular, \emph{short-cold} IMP matches the performance of \emph{long-warm} one without any warmup, \ie \emph{short-cold} IMP can find lottery-tickets at initialization.
However, note that \emph{short-cold} IMP only finds lottery tickets, \textbf{not} rare gems in that the subnetworks it finds have accuracy close to that of random guessing.

\end{document}